\documentclass[10pt, letter, onecolumn]{arxiv}

\usepackage{kantlipsum, lipsum}
\usepackage{dm-colors}
\usepackage{amsmath}
\usepackage{pstricks, pst-node}
\usepackage{verbatim}
\usepackage{multirow}
\usepackage{array}
\usepackage{longtable}
\usepackage{pdflscape}
\usepackage{scalerel}
\usepackage{booktabs}
\usepackage{enumitem}
\usepackage{xspace}
\usepackage{bm}
\usepackage{bbm}
\usepackage{mathtools}
\usepackage{soul}
\usepackage{epsfig}
\usepackage{graphicx}
\usepackage{amssymb}
\usepackage{colortbl}
\usepackage{csquotes}
\usepackage{setspace}
\usepackage{colortbl}
\usepackage{bbding}
\usepackage{threeparttable}
\usepackage{tabularx,ragged2e}
\usepackage{placeins}
\usepackage{bbding}
\usepackage{float}
\usepackage{afterpage}
\definecolor{darkpastelgreen}{rgb}{0.13, 0.55, 0.13}
\definecolor{darkpastelred}{rgb}{0.55, 0.13, 0.13}
\definecolor{mygray}{rgb}{1, 1, 1}
\usepackage{lmodern}
\usepackage[hang,flushmargin]{footmisc}

\usepackage{amsfonts, amssymb}
\usepackage{nameref}
\usepackage{varioref}
\usepackage{amssymb}
\usepackage{pifont}
\usepackage{rotating}
\usepackage{graphicx}

\usepackage[pagebackref=false,breaklinks=false,%
            colorlinks=true,bookmarks=true,citecolor=ourdarkblue,%
            urlcolor=ourdarkblue,linkcolor=ourdarkblue]{hyperref}
\usepackage[noabbrev,capitalize]{cleveref}
\usepackage{etoc}
\usepackage{lineno}
\usepackage{algorithm}
\usepackage{makecell} 
\usepackage{algpseudocode}
\usepackage{adjustbox}

\usepackage{amsmath}
\usepackage{thmtools}
\usepackage{tcolorbox}
\usepackage{lmodern}
\usepackage{booktabs, multirow, makecell, array, graphicx, adjustbox}

\usepackage{listings}
\usepackage{tcolorbox} 
\usepackage{enumitem}

\declaretheoremstyle[
    spaceabove=6pt, spacebelow=6pt,
    headfont=\bfseries, headpunct={.}, headformat={\NAME\ \NUMBER},
    bodyfont=\normalfont,
    postheadspace=0.5em
]{promptstyle}

\tcolorboxenvironment{prompt}{
    colback=gray!10!,
    colframe=gray!75!,
    fonttitle=\bfseries,
    title=Prompt,
    boxrule=0.5pt,
    sharp corners
}

\graphicspath{{figures/}}
\definecolor{mygray}{rgb}{0.85, 0.85, 0.85}

\title{\Large{GenTac: Generative Modeling and Forecasting of Soccer Tactics}}

\author[1]{Jiayuan Rao} 
\author[1,2]{Tianlin Gui} 
\author[1]{Haoning Wu} 
\author[1]{Yanfeng Wang} 
\author[1$\dag$]{Weidi Xie}

\affil[1]{\normalsize Shanghai Jiao Tong University, Shanghai, China \authorcr \vspace{0.1cm}}
\affil[2]{\normalsize Nanjing University, Nanjing, China \authorcr \vspace{0.1cm}}

\affil[$\dag$]{\normalsize Corresponding author\authorcr Weidi Xie: weidi@sjtu.edu.cn}

\begin{document}

\begin{abstract}
Modeling open-play soccer tactics is a formidable challenge due to the stochastic, multi-agent nature of the game. Existing computational approaches typically produce single, deterministic trajectory forecasts or focus on highly structured set-pieces, fundamentally failing to capture the inherent variance and branching possibilities of real-world match evolution. 
Here, we introduce \textbf{GenTac}, a diffusion-based generative framework that conceptualizes soccer tactics as a stochastic process over continuous multi-player trajectories and discrete semantic events. By learning the underlying distribution of player movements from historical tracking data, GenTac samples diverse, plausible, long-horizon future trajectories. The framework supports rich contextual conditioning—including opponent behavior, specific team or league playing styles, and strategic objectives—while grounding continuous spatial dynamics into a 15-class tactical event space. Extensive evaluations on our proposed benchmark, \textbf{TacBench}, demonstrate four key capabilities: (1) GenTac achieves high geometric accuracy while strictly preserving the collective structural consistency of the team; (2) it accurately simulates stylistic nuances, distinguishing between specific teams ({\em e.g.}, Auckland FC) and leagues ({\em e.g.}, A-League versus German leagues); (3) it enables controllable counterfactual simulations, demonstrably altering spatial control and expected threat metrics based on offensive or defensive guidance; and (4) it reliably anticipates future tactical outcomes directly from generated rollouts. Finally, we demonstrate that GenTac can be successfully trained to generalize to other dynamic team sports, including basketball, American football, and ice hockey. By unifying trajectory forecasting, event anticipation, and counterfactual simulation, GenTac establishes a robust generative foundation for decoding complex multi-agent dynamics in sports AI.

\end{abstract}

\maketitle

\section{Introduction}
Association football~(soccer) is one of the most globally renowned and influential sports, carrying profound cultural and economic significance. While the determinant of victory is straightforward, {\em i.e.}, scoring more goals than the opponent, the game itself constitutes a complex multi-player interaction among 22 players on the pitch, making tactical coordination central to competitive outcomes. Soccer tactics are inherently multifaceted, encompassing player formations, spatial positioning, and cooperative strategies formulated in response to continuously evolving match dynamics. Rapid fluctuations in player decision-making, opponent behavior, and match situations make systematic modeling and prediction of on-pitch tactics highly challenging. Critically, however, the essence of these tactical dynamics can be abstracted through a series of multi-player trajectories and key game events, providing a rich geometric canvas for rigorous computational analysis.

The recent explosion of high-fidelity tracking data, combined with advances in deep learning, has catalyzed a wide array of sports analytics applications. These range from event classification~\cite{SoccerNet, SoccerNetv2, SoccerNetv3, SoccerNetv3-Tracking, rao2025unisoccer}, automated commentary~\cite{rao2024matchtimeautomaticsoccergame, densecap} and visual question aswering~\cite{rao2025socceragent} to pitch reconstruction~\cite{SoccerNet-GSR, SoccerNet-Camera, CameraCalibration, yang2025soccermaster}. Nevertheless, capturing the true tactical flow of open play remains elusive. Research specifically targeting soccer tactics has predominantly focused on the statistical summarization and post-hoc qualitative interpretation of historical matches~\cite{bialkowski2014identifying, garrido2020consistency, buldu2019defining, huang2025impact}. While a few recent AI frameworks have attempted to predict future tactical developments, such as TacticAI~\cite{wang2024tacticai}, TacEleven~\cite{zhao2025taceleven}, and JointDiff~\cite{capellera2025jointdiff}, these approaches are largely constrained to static, highly structured set-pieces ({\em e.g.}, corner kicks) or are designed to produce single, deterministic trajectory forecasts. Crucially, this overlooks a fundamental reality of the sport: open play is inherently non-deterministic. A single pass or off-the-ball run can cause the match to branch into a multitude of plausible future scenarios.

To bridge this gap, we conceptualize soccer tactics as a stochastic generative process over multi-player trajectories and their associated semantic events. We introduce \textbf{GenTac}, a diffusion-based generative framework designed for the comprehensive exploration and simulation of tactical dynamics. Given a segment of historical match data, GenTac learns the underlying distribution of player movements, enabling us to sample a diverse array of plausible, long-horizon future trajectories for all players during open play. By drawing multiple coherent samples rather than yielding a single deterministic prediction, this approach effectively captures the natural variance and the unpredictability of match evolution. Furthermore, GenTac can be conditioned on diverse contextual signals, such as opponent behavior, specific team or league playing styles, and strategic objectives. By linking continuous spatiotemporal trajectory forecasting with discrete semantic event types, GenTac unifies tactical evolution, event anticipation, and counterfactual match simulation within a single framework.

Extensive experiments validate this generative approach on our proposed benchmark, \textbf{TacBench}, yielding four key findings:
(i)~{GenTac produces robust, multi-agent trajectory predictions} that maintain high geometric accuracy and preserve the collective structural consistency of the team.
(ii)~{GenTac successfully captures stylistic nuances across different teams and leagues}; for instance, conditioning the model on specific team identities ({\em e.g.}, Auckland FC) significantly reduce geometrical error and improve collective structure consistency over a 5-second horizon, 
while league-specific conditioning accurately reflects the distinct playing tempos and spatial patterns of different competitions ({\em e.g.}, the Australian A-League versus German leagues).
(iii)~{GenTac supports controllable, counterfactual tactical simulation}: conditioning the generative process on offensive objectives demonstrably increases attacking threat metrics, whereas conditioning on defensive objectives suppresses threat and strengthens spatial control.
(iv)~{GenTac effectively grounds continuous trajectories into discrete tactical events}, achieving highly accurate event recognition across 15 distinct categories and successfully anticipating upcoming match outcomes directly from generated futures.

Collectively, these findings establish that soccer tactics can be rigorously characterized as a generative process over coordinated multi-player behavior and its associated event structure. GenTac provides a unified framework for forecasting match dynamics, simulating tactical responses, grounding event semantics, and steering collective behavior toward specified objectives. By successfully modeling the intricacies of open-play soccer, this approach not only advances the frontier of sports analytics but also demonstrates how generative AI can illuminate the strategic depths of the world's most popular sport.

\section{Evaluation Settings}
We establish the evaluation framework for \textbf{GenTac} as follows. 
Section~\ref{subsec:overview} provides a brief overview. 
Section~\ref{subsec:task} defines the trajectory forecasting and tactical event understanding tasks. 
Section~\ref{subsec:benchmark} describes the benchmarks and metrics used to assess predictive accuracy, tactical structure, and tactical event recognition.

\subsection{GenTac Overview}
\label{subsec:overview}
We introduce \textbf{GenTac}, a trajectory-centric generative framework for forecasting and analyzing open-play soccer tactics. Built upon multi-player trajectories as its core representation, the framework models the evolution of collective movement patterns conditioned on game context and opponent dynamics.
By operating directly on these spatiotemporal trajectories, GenTac synthesizes and analyzes multiple coherent and tactically plausible future scenarios. This facilitates probabilistic forecasting, tactical structure interpretation, and reasoning regarding alternative strategic developments in open play.

\subsection{Task Formulation}
\label{subsec:task}

Our proposed trajectory-centric generative framework, \textbf{GenTac}~($\Phi = \{\Phi_{\mathrm{traj}}, \Phi_{\mathrm{event}}\}$), features a module~($\Phi_{\mathrm{traj}}$) for the task of \textbf{multi-player trajectory forecasting} and a module~($\Phi_{\mathrm{event}}$) for the task of \textbf{tactical event recognition}.
These tasks systematically benchmark the capability to anticipate future trajectories and interpret the semantic outcomes of tactics emerging from trajectories.

\paragraph{Multi-player trajectory forecasting.}
This task involves predicting the future 2D coordinates~(on a standard 105\,m\,$\times$\,68\,m pitch) of \textbf{all players and the ball} based on a fixed window of historical observations. 
Formally, given historical trajectories~($\mathbf{x}_{\text{h}}$) and game context~($\mathbf{c}$), the trajectory prediction module~($\Phi_{\mathrm{traj}}$) of GenTac generates future trajectories~($\mathbf{x}_{\text{f}}$) via:
\begin{equation}
    \mathbf{x}_{\text{f}} = \Phi_{\mathrm{traj}}(\mathbf{x}_{\text{h}}, \mathbf{c})
\end{equation}
Here, the game context~$\mathbf{c} \in \mathcal{C}$ denotes an optional conditioning variable that specifies the tactical context of forecasting, with
$\mathcal{C}=\{\varnothing, \mathbf{c}^{\mathrm{opp}}, \mathbf{c}^{\mathrm{team}}, \mathbf{c}^{\mathrm{league}}, \mathbf{c}^{\mathrm{obj}}\}.$
We first consider two base forecasting settings, depending on whether the model predicts the future trajectories of both teams jointly or those of a single team:

\vspace{-4pt}
\begin{itemize}
    \setlength \itemsep{3pt}
    \item \textbf{Unconditioned ($\mathbf{c}=\varnothing$)}: Jointly predicts the future trajectories of both teams from historical observations, serving as the default formulation for two-team forecasting.
    \item \textbf{Opponent-conditioned ($\mathbf{c}=\mathbf{c}^{\mathrm{opp}}$):} Predicts the future trajectory of a single target team conditioned on historical observations and the full trajectory (both past and future) of the opposing team. This serves as the primary formulation for simulating targeted tactical responses.
\end{itemize}
\vspace{-4pt}

Beyond opponent conditioning, we introduce higher-level contextual conditions to enrich the forecasting process, capturing specific stylistic identities and strategic goals:

\vspace{-4pt}
\begin{itemize}
    \setlength \itemsep{3pt}
    \item \textbf{Team-conditioned ($\mathbf{c}=\mathbf{c}^{\mathrm{team}}$):} Conditions the generation on a specific team identity. This enables the simulation of how a particular team reacts to opponents by leveraging learned, team-specific tactical priors and stylistic tendencies.
    \item \textbf{League-conditioned ($\mathbf{c}=\mathbf{c}^{\mathrm{league}}$):} Conditions the generation on a specific league identity. This facilitates the joint prediction of both teams' trajectories, governed by the broader tactical patterns and stylistic regularities characteristic of that league.
    \item \textbf{Objective-conditioned ($\mathbf{c}=\mathbf{c}^{\mathrm{obj}}$):} Conditions the generation on a specific tactical objective, such as maximizing attacking threat. This guides the trajectory generation toward desired strategic outcomes, providing a powerful mechanism for counterfactual analysis and tactical optimization.
\end{itemize}

\begin{figure}[p]
    \centering
    \includegraphics[width=\linewidth]{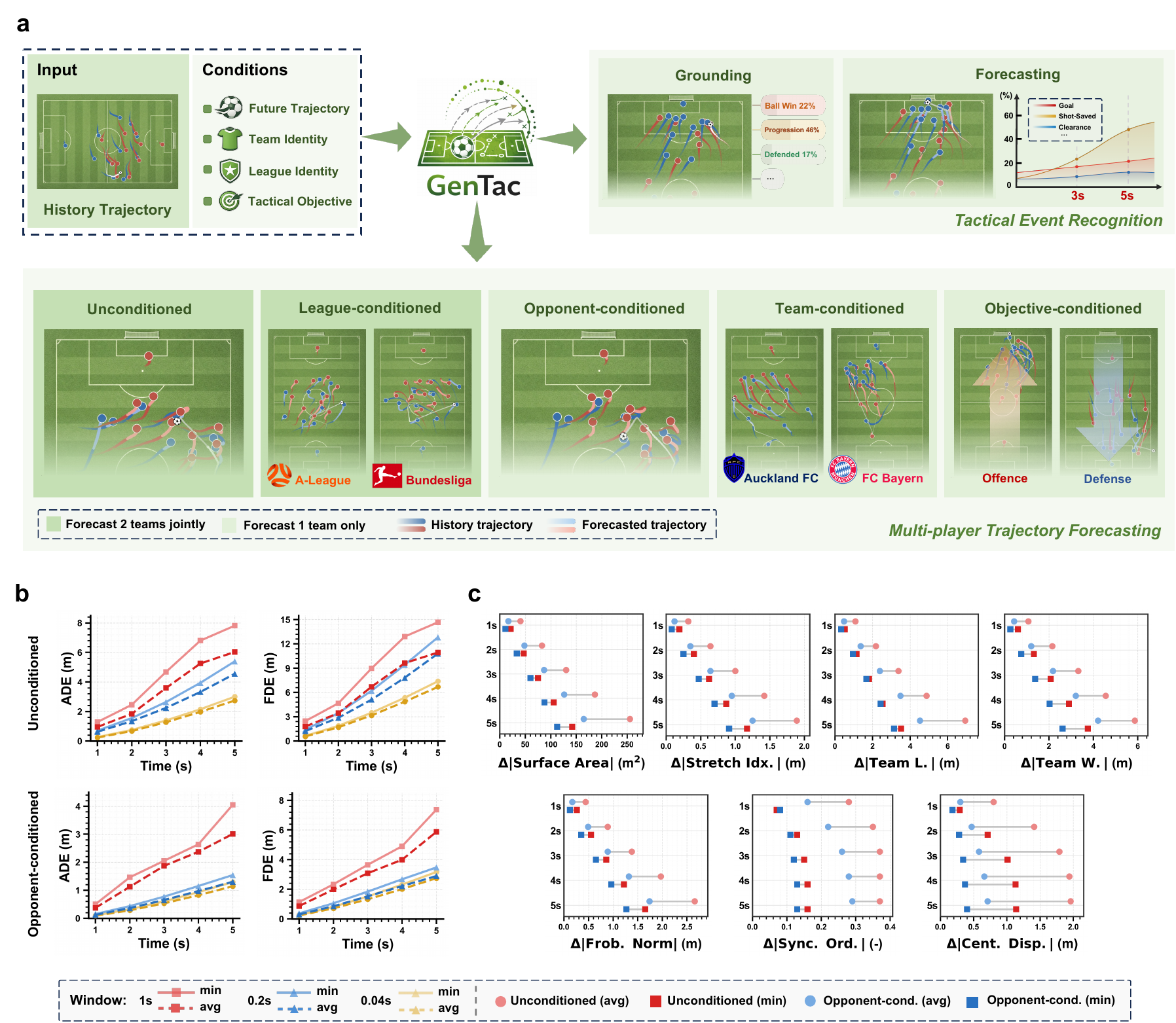}
    \vspace{12pt}
    \caption{\textbf{(a) Overview of GenTac.} We illustrate the two primary capabilities of the framework: multi-player trajectory forecasting and tactical event recognition. Trajectory forecasting encompasses five distinct conditioning configurations based on the number and identity of the teams being predicted. Meanwhile, tactical event recognition is evaluated under two settings: event grounding (inferred from observed data) and event forecasting (anticipated from generated futures).
    \textbf{(b) Geometric accuracy of trajectory prediction.} The x-axis represents the future time horizon (up to 5 seconds), and the y-axis reports the prediction error measured by Average Displacement Error~(ADE) and Final Displacement Error~(FDE) in meters. The top and bottom rows display results for unconditioned and opponent-conditioned forecasting, respectively. Lower values indicate superior geometric accuracy.
    \textbf{(c) Collective structural consistency of forecasted trajectories.} We report the absolute differences between structural metrics computed from the predicted trajectories and the ground truth over a 5-second future horizon. Circles and squares denote the mean and minimum errors, respectively. Red markers correspond to unconditioned forecasting, while blue markers indicate opponent-conditioned forecasting. The evaluated structural metrics include surface area~($\Delta|\mathrm{Surface\ Area}|$), stretch index~($\Delta|\mathrm{Stretch\ Idx.}|$), team length~($\Delta|\mathrm{Team\ L.}|$), team width~($\Delta|\mathrm{Team\ W.}|$), Frobenius norm of pairwise distances~($\Delta|\mathrm{Frob.\ Norm}|$), synchronization order parameter~($\Delta|\mathrm{Sync.\ Ord.}|$), and team centroid displacement~($\Delta|\mathrm{Cent.\ Disp.}|$). Detailed definitions of these metrics are provided in Supplementary~\ref{appendix:metric}.}
    \label{fig:overview}
\end{figure}

As illustrated in Figure~\ref{fig:overview}a, these conditioning mechanisms unify diverse analytical applications, ranging from style simulation to tactical guidance, under a single, systematic trajectory forecasting paradigm.

\paragraph{Tactical event recognition.}
Beyond geometric accuracy, a comprehensive understanding of soccer tactics necessitates semantic interpretation. We evaluate GenTac's capacity for event recognition across two distinct settings:
(i) In \textbf{event grounding}, a tactical event~($\mathbf{e}$) is inferred directly from observed trajectories~($\mathbf{x}_{\text{h}}$) via:
\begin{equation}
    \mathbf{e} = \Phi_{\mathrm{event}}(\mathbf{x}_{\text{h}}),
\end{equation}
where $\Phi_{\mathrm{event}}$ maps multi-player trajectories to predefined tactical labels.
(ii) In \textbf{event forecasting}, GenTac first generates future trajectories from historical trajectories~($\mathbf{x}_{\text{h}}$) and context~($\mathbf{c}$), then infers the event outcome from the generated scenarios:
\begin{equation}
    \mathbf{e} = \Phi_{\mathrm{event}}\big(\Phi_\mathrm{traj}(\mathbf{x}_{\text{h}}, \mathbf{c})\big)
\end{equation}
Together, these settings establish a unified protocol for tactical analysis, explicitly linking the continuous spatiotemporal evolution of play to its discrete semantic interpretation.

\subsection{Benchmark Description}
\label{subsec:benchmark}

\paragraph{Benchmarks.}
We introduce \textbf{TacBench}, a unified benchmark for soccer tactical forecasting and analysis, constructed from publicly available full-match multi-player trajectory datasets with synchronized event annotations: \textit{SkillCorner}~\cite{skillcorner_opendata}~(Australian A-League), \textit{Sportec DFL}~\cite{bassek2025integrated}~(German Bundesliga and 2.\ Bundesliga), and \textit{Metrica Sports}~\cite{metrica_sports_sample_data}. All trajectory data are resampled to 25\ frame per second~(FPS), with each frame recording the on-pitch coordinates of all players and the ball. 
TacBench comprises two components corresponding to different tasks: a \textbf{trajectory-only benchmark} (TacBench-Trajectory) for trajectory forecasting and a \textbf{tactical event benchmark} (TacBench-Event) for tactical event recognition, as following shows. 
All evaluations are conducted on held-out test sets, and preprocessing and split details are provided in Supplementary~\ref{appendix:dataset}.

\vspace{-4pt}
\begin{itemize}
    \setlength \itemsep{4pt}
    \item \textbf{TacBench-Trajectory}: 2,838 15-second segments spanning diverse matches and leagues, each containing complete on-pitch coordinates for all players and the ball. This benchmark evaluates \textbf{unconditioned} and \textbf{opponent-conditioned} trajectory forecasting. Subsets annotated with team identity or league origin additionally support \textbf{team-} and \textbf{league-conditioned} forecasting.
    \item \textbf{TacBench-Event}: 423 trajectory segments with explicit tactical annotations, with durations defined by event boundaries. Segments are categorized into 5 major event types~(\textit{build-up}, \textit{transition}, \textit{interruption}, \textit{set piece}, and \textit{threat}) and 15 fine-grained subtypes~(Figure~\ref{fig:event}~ab). This benchmark supports \textbf{event grounding} and, for subsets associated with specific offensive or defensive objectives, \textbf{objective-conditioned} trajectory forecasting using the preceding 4\,s as historical context.
\end{itemize}

In addition to soccer, we incorporate multi-player trajectory benchmarks from other team sports to demonstrate the generalization of our method by retraining models on corresponding sports. These benchmarks are derived exclusively from the test splits of their respective source datasets and retain their original configurations, such as temporal resolutions~(as detailed in the {Supplementary}). 
Concretely, we include: 
(i) \textbf{Basketball:} 8,398 clips of 6 seconds at 5\,FPS sourced from \textit{NBA SportVU}~\cite{nba_sportvu_tracking}, collected from NBA games; 
(ii) \textbf{American football:} 301 clips of 8 seconds at 10\,FPS from the \textit{NFL Big Data Bowl 2021}~\cite{nfl_big_data_bowl_official}, comprising offensive plays of over 10 yards from regular-season matches; 
and 
(iii) \textbf{Ice hockey:} 307 clips of 10 seconds at 30\,FPS from the \textit{Big Data Cup 2026}~\cite{bigdatacup_2026_data_release}, spanning 4 matches.

\paragraph{Metrics.}
We evaluate trajectory forecasting and tactical recognition using a comprehensive suite of metrics covering trajectory fidelity, collective spatial structure, and semantic outcomes:
(i) \textbf{Trajectory accuracy.}
To quantify the geometric precision of multi-player forecasting and tactical simulation using the average displacement error~(ADE) and final displacement error~(FDE);
(ii) \textbf{Collective structure consistency.}
To assess the team organization and tactical coherence of forecasted trajectories, 
we utilize a set of established collective motion indicators, including the stretch index, surface area, team width and length, Frobenius norm, team centroid displacement, and the Kuramoto order parameter~\cite{Bartlett01082012,frencken2011oscillations,duarte2013competing};
(iii) \textbf{Soccer event accuracy.} We report top-k overall accuracy together with recall for trajectory-based tactical event grounding and forecasting. 
(iv) \textbf{Offense/defense performance.} For semantic-guided exploration, we evaluate the effectiveness of specific tactical guidance by 3 offensive metrics (off-ball expected threat, depth threat, width threat) describing the offensive threat inspired from previous works~\cite{singh2018xt} and 2 defensive metrics (defensive shape disruption and defensive dominant region~\cite{taki2000visualization}) models how well the defensing team performs, for example, its valid control region of pitch.
For all the evaluation settings involving trajectory forecasting, we sample $K$ prediction instances and analyze the distribution ({\em e.g.}, minimum, average) to further evaluate the performance of GenTac across these metrics. More detailed definitions and implementation are provided in the Supplementary~\ref{appendix:metric}.

\section{Results}

In this section, we systematically evaluate GenTac across the trajectory forecasting and tactical event recognition. 
Section~\ref{subsec:result-trajectory} focuses on multi-player trajectory forecasting under various conditional settings, examining geometric accuracy and structural consistency. 
Section~\ref{subsec:result-semantic} investigates tactical event recognition, including both event grounding from given trajectories and event forecasting from generated futures.

\subsection{GenTac demonstrates superior performance on multi-player trajectory forecasting}
\label{subsec:result-trajectory}

\paragraph{GenTac enables sampling multiple trajectory variants.}
\label{subsubsec:result-trajectory-causal}
Open-play soccer admits multiple tactically plausible continuations from the same history. 
GenTac therefore frames trajectory forecasting as sampling from a conditional future distribution rather than producing a single deterministic rollout. 
To generate long-horizon futures stably, GenTac uses a causal sliding window that decomposes the prediction horizon into short consecutive segments and samples them sequentially. 
For example, forecasting 1\,s into the future with a causal window of 0.2\,s requires five successive sampling steps, each conditioned on the previously generated segment. 
Repeating this process $K=20$ times from the same history and conditions yields diverse yet realistic collective evolutions. 
In the unconditioned setting, both teams are jointly predicted from shared history. 
Conditional variants introduce additional context, such as opponent futures, tactical objectives, or style cues, to guide specific future patterns while preserving collective tactical structure.

\paragraph{GenTac demonstrates stable long-horizon forecasting.}
We evaluate GenTac under \textbf{unconditioned} forecasting setting ($\mathbf{c}=\varnothing$), where both teams are jointly predicted from shared 4-second historical trajectories through a causal window. Under a 0.2s window, the minimum Average Displacement Error~(ADE) across 20 samples ranges from 0.62\,m to 4.55\,m when forecasting from 1s to 5s, while the corresponding Final Displacement Error~(FDE) grows from 1.22\,m to 10.80\,m, the mean errors exhibit a similar trend without diverging. An ablation on the window size further confirms that reducing the window improves geometric precision. For example, the minimum ADE and FDE at 5s decreases to 2.74\,m and 6.68\,m under a 0.04s window (see Figure~\ref{fig:overview}b). 
However, because a shorter window increases computational cost, we adopt 0.2\,s as a balanced trade-off in subsequent experiments. 
Beyond geometric accuracy, we assess collective structural consistency by reporting the absolute deviations of corresponding metrics on multiple future trajectories against those on the ground truth. 
Under a 0.2s causal window, for example, the deviation of stretch index ranges from 0.32 to 1.89, and the centroid displacement differentiates within 0.8--1.96\,m across horizons (see Figure~\ref{fig:overview}c). 
Together, these results indicate that unconditioned forecasting achieves stable performance in both geometric fidelity and collective structural coherence.

\paragraph{GenTac improves forecasting accuracy under opponent conditioning.}
We next evaluate \textbf{opponent-conditioned} forecasting ($\mathbf{c}=\mathbf{c}^{\mathrm{opp}}$), where future trajectories of one team are predicted given the motion of the opponent, enabling more reactive tactical evolution. As shown in Figure~\ref{fig:overview}b, conditioning on the opponent substantially improves geometric accuracy under the same 4s history and 0.2s causal window setting. 
At a 5s horizon, the minimum ADE and FDE decrease over 70\% to 1.30\,m and 2.89\,m, respectively, while mean errors follow the same trend across all horizons. Furthermore, collective structural deviations are significantly reduced. At 5s, the mean stretch index deviation decreases from 1.89 to 1.25, and centroid displacement deviation drops from 1.96\,m to 0.71\,m, with minimum deviations across sampled futures showing consistent improvements. 
These results demonstrate that incorporating opponent futures enhances both geometric precision and collective structural coherence, enabling more accurate and tactically grounded conditional forecasting.

\begin{figure}[p]
    \centering
    \includegraphics[width=\linewidth]{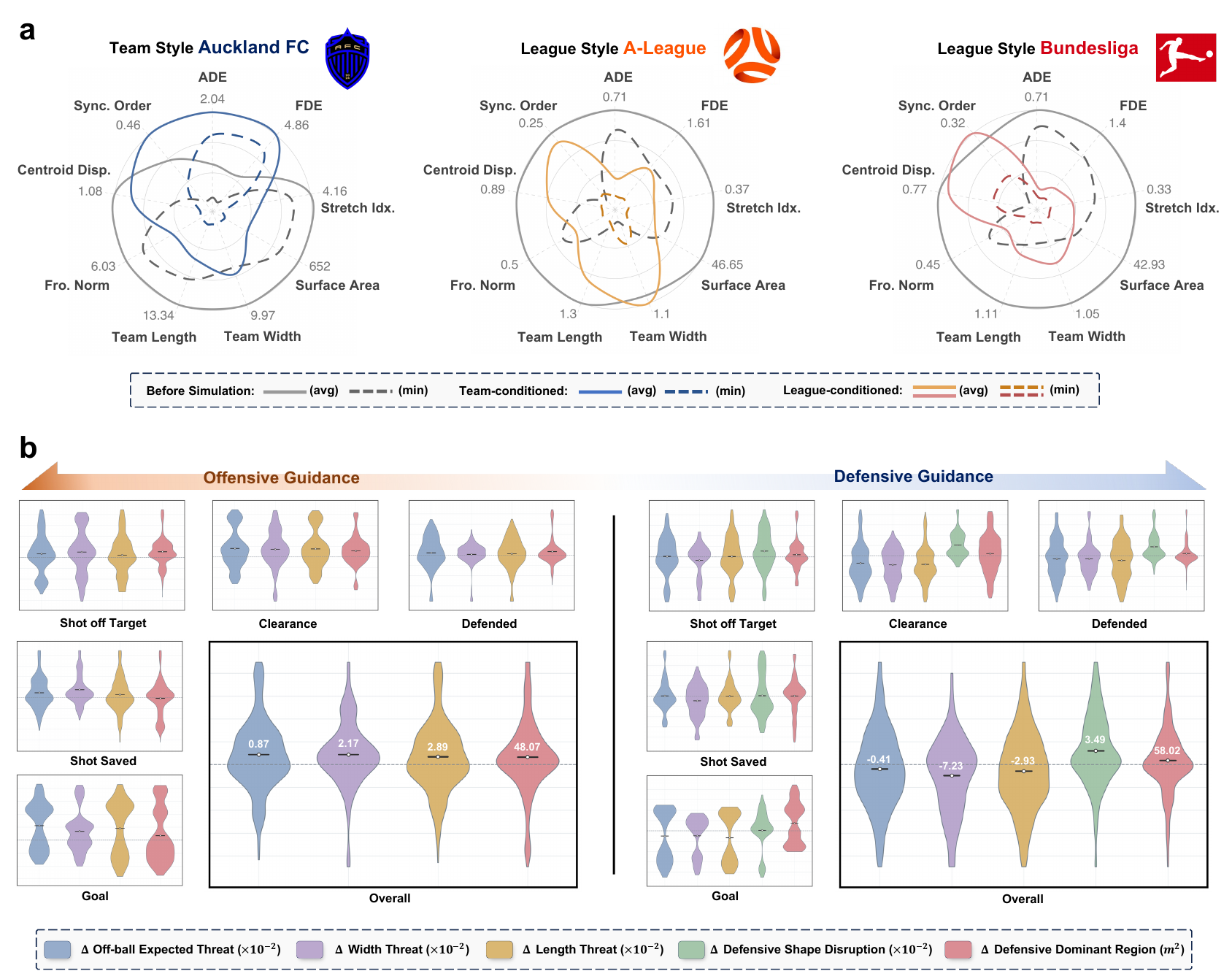}
    \vspace{6pt}
    \caption{\textbf{(a) Performance of team- and league-conditioned trajectory forecasting.} We evaluate the geometric accuracy and structural consistency of GenTac when generating collective trajectories conditioned on specific team or league identities. The results demonstrate the model's ability to simulate the distinct playing style of a specific team (Auckland FC) over a 5-second horizon, as well as the broader stylistic differences between competitions (the Australian A-League versus German leagues) over a 1-second horizon.
    \textbf{(b) Performance of objective-conditioned trajectory forecasting.} We illustrate the deviation in tactical metrics when guiding trajectory generation toward specific strategic objectives ({\em offense} versus {\em defense}). The evaluated metrics quantify both attacking threat and defensive solidity: {\em off-ball expected threat}, {\em width threat}, and {\em depth threat} measure attacking intent (where higher values indicate greater offensive threat), while {\em defensive shape disruption} and {\em defensive dominant region} characterize the structural integrity and spatial control of the defending team. Detailed definitions of these metrics are provided in Supplementary~\ref{appendix:metric}. }
    \label{fig:guidance}
\end{figure}

\paragraph{GenTac simulates team-specific tactical styles and responses.}
Building upon opponent-conditioned forecasting, we further examine whether GenTac can capture the tactical style of a specific team, denoted as \textbf{team-conditioned} forecasting~($\mathbf{c}=\mathbf{c}^{\mathrm{team}}$). 
Here, future trajectories are predicted given the opponent's motion, while additionally conditioning on a specific team identity to capture how that particular team reacts to its opponent. 
We fine-tune the opponent-conditioned forecasting model on Auckland FC under the same configuration~(4s history and 0.2s causal window), and forecast trajectories of Auckland FC as well. 
As shown in Figure~\ref{fig:guidance}a, though the geometric errors out of $K$(=20) future samples gets slightly increased after such style simulation~({\em e.g.}, average ADE increases from 1.65\,m to 2.04\,m at the 5s horizon). The accuracy across all collective structural metrics are substantially better. At 5s, the average deviation over 20 future samples decreases on the stretch index from 4.16 to 2.81, surface area from 652\,m$^2$ to 335\,m$^2$, and team width/length from 9.97/13.34\,m to 8.26/8.29\,m, with similar trends in the minimum deviation. Such consistent reductions indicate that team conditioning enables GenTac to better capture the movement patterns of a team responding to opponent actions.

\paragraph{GenTac captures league-specific tactical patterns in short-horizon forecasting.}
Soccer game styles vary across leagues due to differences in tactical culture and playing tempo. 
To evaluate whether GenTac reflects such league-level characteristics, we perform \textbf{league-conditioned} forecasting ($\mathbf{c}=\mathbf{c}^{\mathrm{league}}$) by fine-tuning the model of unconditioned forecasting on matches from the Australian A-League and the German leagues respectively under the same configuration (4s history and a 0.2s causal window). 
For each history segment, GenTac generates 20 future trajectory samples, and we evaluate their prediction errors.
As shown in Figure~\ref{fig:guidance}a, for both leagues, conditioning on league identity consistently improves short-horizon forecasting. 
For example, in the A-League, average ADE decreases 38.0\% at future 1s of forecasting and 14.9\% for 2s, while the trend remains the same in the German leagues, and the deviations of structural metrics exhibit comparable reductions at short futures as well.
As the forecast extends towards 5s, however, the advantage gradually diminishes, and the errors approach those of the previous baseline. 
Overall, these results indicate that league conditioning helps GenTac produce league-level collective motion tendencies, particularly in short-term trajectory evolution.

\paragraph{GenTac enables tactical guidance for a determined objective.}
Beyond team and leagues simulation, we further evaluate whether the generated trajectories can be guided by explicit tactical objectives, referred to as \textbf{objective-conditioned} forecasting ($\mathbf{c}=\mathbf{c}^{\mathrm{obj}}$).
We select segments of successful defense ({\em clearance, defended}) and threatening offense ({\em goal, shot saved, shot off target}) from TacBench-Event and fine-tune GenTac on them respectively. The resulting models are evaluated by generating trajectories for the defending or attacking side across all such segments in the test set. 
As shown in Figure~\ref{fig:guidance}b, under \textbf{defensive guidance}, the generated trajectories consistently suppress attacking threat. The off-ball expected threat decreases by about $0.41$, while depth threat and width threat decrease by about $7.23\times10^{-2}$ and $2.93\times10^{-2}$, respectively. Simultaneously, defensive disruption and dominant region control increase, particularly in {\em clearance} and {\em defended} scenarios, indicating more compact and coordinated defensive organization. Under \textbf{offensive guidance}, GenTac produces trajectories that expand attacking threat. Off-ball expected threat increases by about $0.87$, with depth and width threat are stably risen as well. These improvements are most evident in high-threat outcomes such as {\em goal} and {\em shot saved}, although they are accompanied by moderate concessions in pitch region control. Overall, these results show that GenTac can steer collective motion toward specified tactical objectives, producing coordinated team behaviors consistent with either offensive or defensive strategies.

\begin{figure}[t]
    \centering
    \includegraphics[width=\linewidth]{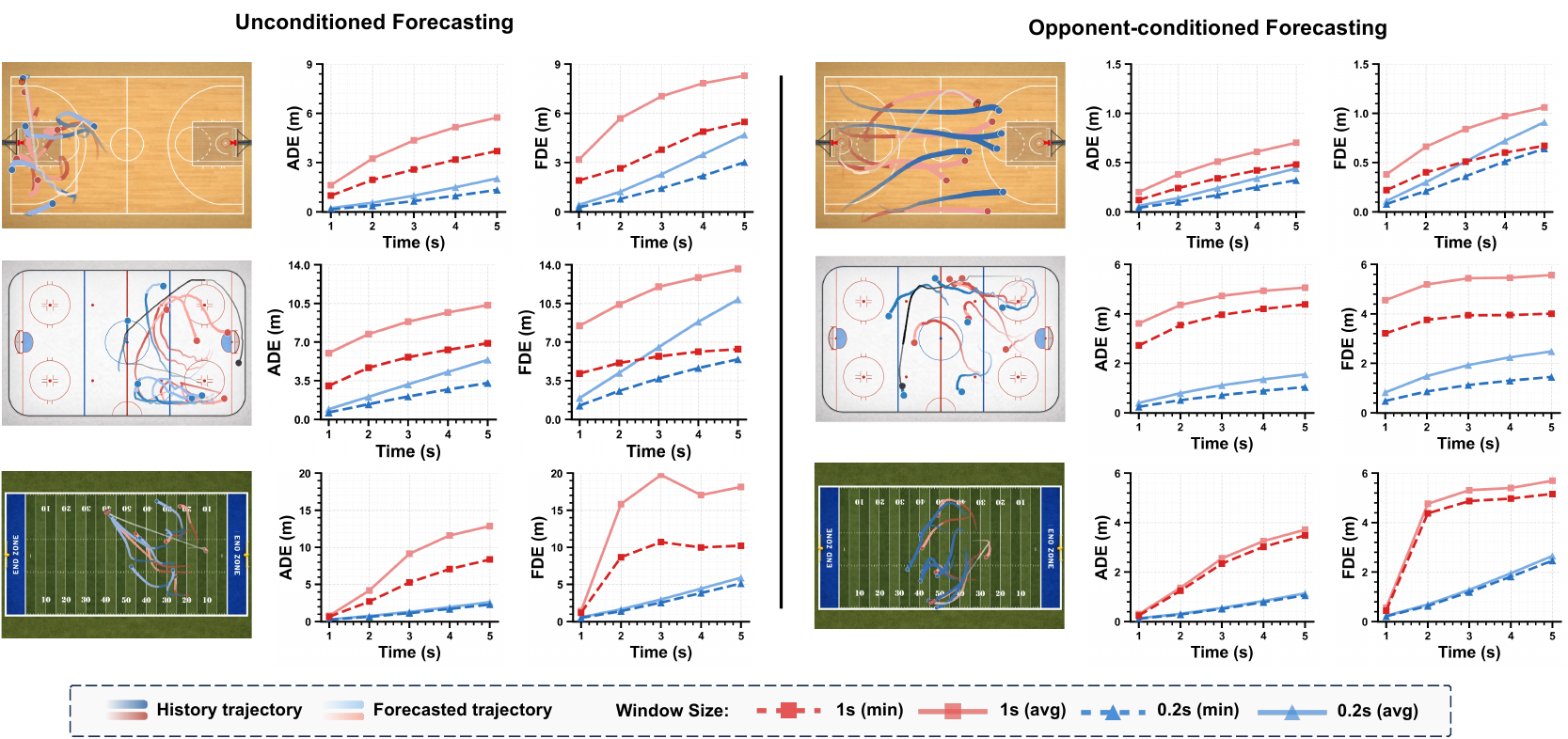}
    \vspace{6pt}
    \caption{
    \textbf{Geometric accuracy of multi-player trajectory forecasting in other team sports.} Demonstrating GenTac's ability to generalize across diverse multi-agent dynamics, we train and evaluate the framework on tracking data from basketball, American football, and ice hockey. We report the geometric accuracy of the forecasted trajectories under both unconditioned and opponent-conditioned settings, evaluated across varying historical observation windows (causal window sizes). 
    }
    \label{fig:generalization}
\end{figure}

\paragraph{GenTac demonstrates a strong capacity for trajectory forecasting in other team sports.}
Having validated diverse conditioning mechanisms in soccer, we also train GenTac on basketball, American football, and ice hockey to evaluate its performance across other team sports. We adopt two base forecasting settings: unconditioned ($\mathbf{c}=\varnothing$) and opponent-conditioned ($\mathbf{c}=\mathbf{c}^{\mathrm{opp}}$). Across all three sports, opponent-conditioned forecasting consistently yields lower geometrical error than the unconditioned setting, highlighting opponent futures as a universally transferable source of tactical context.
As shown in see Figure~\ref{fig:generalization}, at a 5\,s horizon of future prediction, minimum ADE remains at 0.32\,m in basketball, 1.06\,m in American football, and 1.04\,m in ice hockey under opponent conditioning. 
These results, together with qualitative examples in the same figure, suggest that GenTac successfully captures transferable principles of collective dynamics beyond soccer.

\begin{figure}[htbp!]
    \centering
    \includegraphics[width=\linewidth]{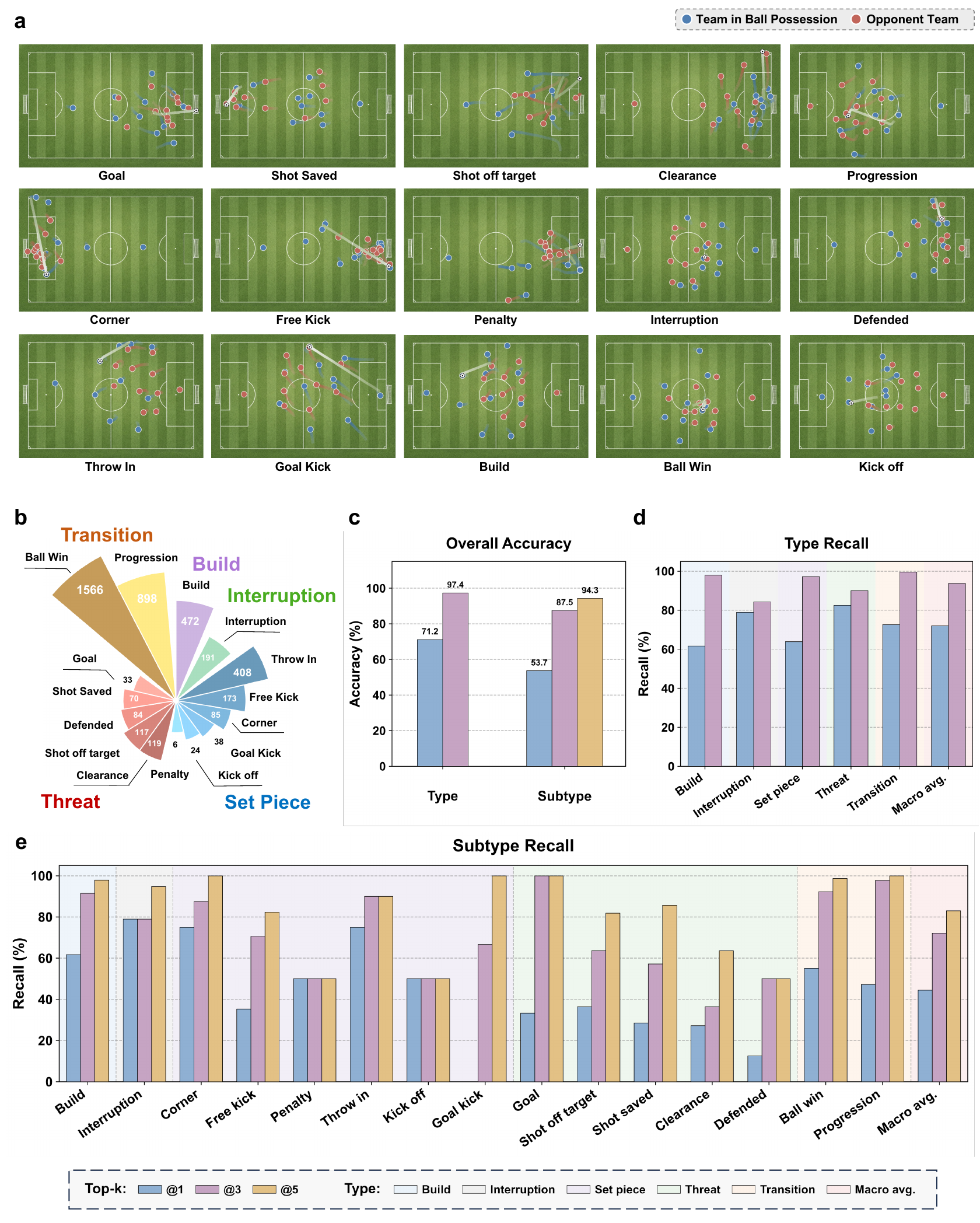}
    \vspace{6pt}
    \caption{\textbf{(a) Examples of tactical events with diverse semantics.} The TacBench-Event dataset encompasses 15 distinct event subtypes, each capturing unique tactical behaviors on the pitch. \textbf{(b) Distribution of event types and subtypes.} Numerical values indicate the total count of each event subtype, while colors correspond to the broader event categories (types). \textbf{(c) Overall accuracy.} We report the top-1, top-3, and top-5 accuracy metrics for both event type and subtype classification. \textbf{(d)-(e) Overall recall.} We report the top-1, top-3, and top-5 recall metrics for both event type and subtype classification.}
    \label{fig:event}
\end{figure}

\subsection{GenTac excels in tactical event recognition from trajectory}
\label{subsec:result-semantic}

Multi-player trajectories capture the spatial relations among players, providing an abstraction of soccer tactics. 
Beyond this geometric structure, trajectories also encode discriminative tactical semantics, enabling both \textit{event grounding and forecasting}. GenTac demonstrates this capability in two settings: 
(i) classifying tactical events from given trajectory segments, and (ii) inferring future event outcomes from generated trajectories. 

\paragraph{Tactical events are directly recoverable from given trajectories.}
Through trajectory evolution, GenTac reliably distinguishes tactical events at both type and subtype levels. 
For the five broad event types, event grounding achieves a top-1 accuracy of 71.2\% and a top-3 accuracy of 97.4\%. 
The macro-averaged Precision@1, Recall@1, and F1-score@1 are 71.3\%, 76.3\%, and 73.4\%, respectively, indicating balanced and stable type-level classification (Figure~\ref{fig:event}c,d). 
Representative categories such as {\em transition} (Recall@1 = 72.7\%, Recall@3 = 99.6\%) and {\em threat} (Recall@1 = 82.5\%, Recall@3 = 90.0\%) exhibit particularly robust separability.
At the finer subtype level (15 subtypes), top-1 accuracy reaches 53.7\%, rising to 87.5\% at top-3 and 94.3\% at top-5; the macro-averaged Recall@1, Recall@3, and Recall@5 are 44.4\%, 72.2\%, and 83.0\%, respectively~(Figure~\ref{fig:event}c,e). 
Collectively, these results demonstrate that GenTac's trajectory representations preserve not only geometric structure but also coherent and interpretable semantic organization.

\begin{figure}[p]
    \centering
    \includegraphics[width=1.08\linewidth]{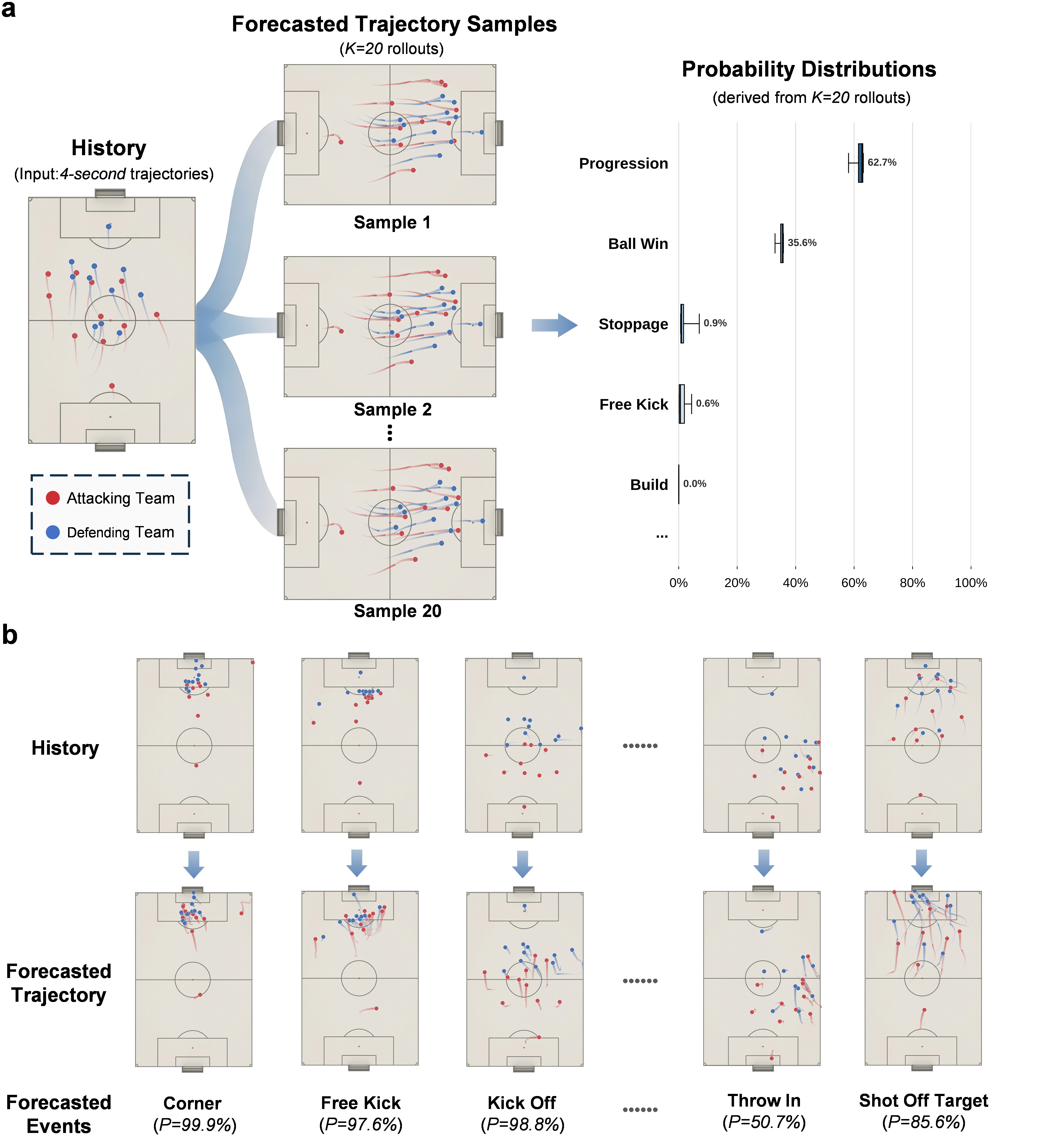}
    \vspace{4pt}
    \caption{\textbf{(a) Pipeline for tactical event forecasting.} Given a historical multi-player trajectory, GenTac encodes the sequence and samples its future evolution $K$(=20) times. Each generated future trajectory is then classified to predict a distribution over tactical events. The box plot illustrates the variance of these predictions across the $K$ independent rollouts, summarizing the top-5 predicted tactical events via their median probabilities, 10--90\% percentile intervals (shaded boxes), and min/max ranges (whiskers). \textbf{(b) Qualitative examples of trajectory and event forecasting.} Given historical observations, GenTac accurately forecasts both the future spatial movements of the players and the corresponding tactical events across a 15-class action space.}
    \label{fig:event_forecast}
\end{figure}

\paragraph{Tactical events can be forecasted from generated future trajectories.}
Previous approaches to soccer event forecasting have largely relied on statistical modeling of historical event sequences, offering limited insight into the tactical evolution that precedes future outcomes. By contrast, we leverage GenTac's ability to generate stable, tactically plausible multi-player trajectory futures. Specifically, we sample multiple future trajectory realizations from an observed history and classify the event associated with each generated trajectory, as illustrated in Figure~\ref{fig:event_forecast}. 
As established in the preceding trajectory forecasting experiments, these generated trajectories faithfully represent realistic tactical developments, thereby providing a grounded basis for event prediction. Consequently, different trajectory realizations correspond to different tactical outcomes, naturally inducing a probability distribution over possible future events while simultaneously revealing the spatial evolution that leads to each of them.

\section{Discussion}
In this work, we introduced GenTac, a generative framework that shifts the paradigm of soccer analytics from post-hoc descriptive analysis to dynamic tactical simulation. By jointly modeling multi-player trajectories and semantic events, this approach opens new avenues for understanding, anticipating, and evaluating complex team behaviors.

\paragraph{Reframing soccer tactics as a generative process.} 
Traditional tactical analyses have largely relied on event statistics, structural summaries, or expert interpretations of completed matches. 
While valuable for characterizing observed patterns, these approaches fundamentally fall short in modeling how tactics unfold dynamically over time. By jointly modeling multi-player trajectories and their associated tactical events, GenTac treats tactics as a dynamic, evolving process that can be generated, compared, and systematically explored. The primary contribution of this work, therefore, extends beyond achieving superior forecasting performance; it introduces a novel modeling paradigm in which tactical evolution becomes accessible to systematic, data-driven simulation.

\paragraph{Embracing unpredictability through distributional modeling.} 
In real-world matches, identical historical contexts can give rise to multiple tactically plausible continuations, driven by opponent responses, spatial organization, and collective team coordination. Deterministic predictions, while capable of capturing an average tendency, inherently fail to encompass the full spectrum of possible tactical developments. Generative modeling offers a more natural and principled alternative by representing future play as a distribution over plausible trajectories. This allows tactical evolution to be studied not as a singular forecasted path, but as a structured, probabilistic space of potential futures.

\paragraph{Elevating forecasting to scenario-based tactical analysis.} 
By conditioning on team and league styles, or on specific offensive and defensive objectives, GenTac simulates how tactical evolution adapts under varying constraints. Our guidance results demonstrate that generative models can effectively steer collective motion toward interpretable tactical goals while preserving coherent multi-player structures. This capability enables the systematic comparison of alternative tactical decisions and their potential consequences, positioning the generative framework not merely as a predictive tool, but as a powerful mechanism for probing how distinct objectives reshape the evolution of play.

\paragraph{Grounding tactical predictions with semantic event forecasting.} 
Tactical events do not arise in isolation; rather, they emerge from dynamic shifts in player spacing, collective structure, and local multi-player interactions. Mapping generated future trajectories to event probabilities establishes a crucial semantic layer for tactical recognition. Instead of predicting pure motion, the model infer the practical match-context meaning of a generated sequence—such as ball progression, clearances, or threat creation. Consequently, event forecasting is not merely an auxiliary task, but a core component that solidifies the interpretability of tactical predictions. It provides generative tactical analysis with a verifiable semantic foundation, advancing a more rigorous and scientific approach to sports analytics.

\section{Limitations}

The limitations of generative tactical modeling stem primarily from the open-ended and highly contextual nature of sports tactics. Real-world tactical decisions are influenced by latent factors not directly captured in standard tracking data, such as coaching intent, player cognition, on-pitch communication, and broader match context ({\em e.g.}, scoreline pressure or fatigue). Consequently, the present framework should be understood as a generative model of observable tactical behavior patterns rather than a complete reconstruction of the underlying cognitive decision-making process. Its outputs represent plausible branches within a tactical possibility space, rather than definitive, deterministic predictions of future play.

Furthermore, tactical structures vary substantially across teams, leagues, and different sporting environments. Although our cross-sport experiments suggest promising transferability, achieving universal generalization remains an open challenge. Additionally, the current study relies heavily on high-fidelity, structured trajectory data. Future work would benefit from incorporating richer, more accessible sources of information, such as trajectories reconstructed directly from broadcast video, pose estimation, and broader multimodal contexts. Such extensions will be vital for improving the robustness, accessibility, and explanatory power of generative tactical modeling in the wild.

\clearpage
\section{Methods}
This section details the architecture and mathematical formulation of the \textbf{GenTac} framework.

\subsection{Problem Formulation}
\label{subsec:problem_formulation}
Building upon the task definitions in Section~\ref{subsec:task}, we formalize the dual objectives of GenTac, 
denoted as $\Phi=\{\Phi_{\text{traj}},\Phi_{\text{event}}\}$: multi-player trajectory forecasting and tactical event recognition. 
We detail how these tasks are unified within our trajectory-grounded generative framework.

\subsubsection{Trajectory Forecasting}
\label{subsubsec:problem_formulation_trajectory}
GenTac formulates multi-player trajectory forecasting as learning a conditional probability distribution over future spatial coordinates, from which we can sample multiple plausible match evolutions. 

\paragraph{Forecasting as conditional generation.}
Rather than predicting a single deterministic outcome, the trajectory generation module $\Phi_{\text{traj}}$ models the stochastic evolution of the match. Given historical trajectory observations $\mathbf{x}_{\text{h}}$ and contextual conditioning $\mathbf{c}$ ({\em e.g.}, game state, team identity, strategic objectives), the predicted future trajectories $\mathbf{x}_{\text{f}}$ are sampled from the learned conditional distribution:
\begin{equation}
    \hat{\mathbf{x}}_{\text{f}} \sim \Phi_{\text{traj}} \big( \mathbf{x}_{\text{f}} \mid \mathbf{x}_{\text{h}}, \mathbf{c}\big),
\end{equation}

\paragraph{Two forecasting settings.}
We define trajectory forecasting under two distinct operational settings. 
Let $\mathbf{x}^i_{[t_1:t_2]}$ denote the trajectory sequence of team $i \in \{1, 2\}$ from time step $t_1$ to $t_2$, 
where $H$ is the length of the historical observation window and $T$ is the prediction horizon.

(i) \textbf{Joint forecasting for both teams:} The model predicts the future movements of all players on the pitch simultaneously. The historical context and future targets are defined as:
\begin{equation}
    \mathbf{x}_{\text{h}} = \{\mathbf{x}^1_{[0:H]}, \mathbf{x}^2_{[0:H]}\}, \quad \mathbf{x}_{\text{f}} = \{\mathbf{x}^1_{[H:T]}, \mathbf{x}^2_{[H:T]}\}.
\end{equation}
        
(ii) \textbf{Single-team forecasting (opponent-conditioned):} The model predicts the future trajectory of a target team $i$, explicitly conditioned on the known future trajectory of the opponent team $j$. This isolates a team's reactive spatial dynamics:
\begin{equation}
    \mathbf{x}_{\text{h}} = \{\mathbf{x}^i_{[0:H]}, \mathbf{x}^j_{[0:T]}\}, \quad \mathbf{x}_{\text{f}} = \{\mathbf{x}^i_{[H:T]}\}, \quad \text{s.t.} \quad i \neq j.
\end{equation}

\paragraph{Autoregressive sliding-window rollout.}
To maintain structural consistency over long-horizon forecasts, we employ an autoregressive sliding-window rollout strategy with a window size $w$. For joint forecasting, the model iteratively samples future segments of length $w$. The $k$-th future segment is sampled by conditioning on the concatenated history and all previously generated segments:
\begin{equation}
    (\hat{\mathbf{x}}^1_{[H+kw : H+(k+1)w]}, \hat{\mathbf{x}}^2_{[H+kw : H+(k+1)w]}) \sim \Phi_{\text{traj}}\big(\cdot \mid \mathbf{x}^1_{[0 : H+kw]}, \mathbf{x}^2_{[0 : H+kw]}, \mathbf{c}\big).
\end{equation}
For single-team forecasting, the opponent's ground-truth trajectory over the prediction window acts as an additional conditioning constraint:
\begin{equation}
    \hat{\mathbf{x}}^i_{[H+kw : H+(k+1)w]} \sim \Phi_{\text{traj}} \big(\cdot \mid \mathbf{x}^i_{[0 : H+kw]}, \mathbf{x}^j_{[0 : H+(k+1)w]}, \mathbf{c}\big).
\end{equation}

\subsubsection{Tactical Event Recognition}
\label{subsubsec:formulation_event}
Beyond continuous spatial forecasting, GenTac grounds collective player movements into discrete semantic actions via the event recognition module $\Phi_{\text{event}}$. Each tactical event $\mathbf{e}$ is structured hierarchically, comprising a coarse event type $y^{\text{type}} \in \mathcal{Y}^{\text{type}}$ and a fine-grained subtype $y^{\text{sub}} \in \mathcal{Y}^{\text{sub}}$. In the TacBench dataset, this label space spans 5 types and 15 nested subtypes.

\paragraph{Event grounding (historical).}
Given an observed historical trajectory segment $\mathbf{x}_{\text{h}}$, the module acts as a classifier to infer the corresponding tactical event:
\begin{equation}
    (\hat{y}^{\text{type}}, \hat{y}^{\text{sub}}) = \Phi_{\text{event}}(\mathbf{x}_{\text{h}}).
\end{equation}

\paragraph{Event forecasting (future).}
Crucially, GenTac anticipates future tactical outcomes by leveraging its generative capabilities. Instead of predicting events directly from the history, GenTac first samples future trajectories $\mathbf{x}_{\text{f}}$ via $\Phi_{\text{traj}}$, and subsequently classifies these generated rollouts:
\begin{equation}
    (\hat{y}^{\text{type}}, \hat{y}^{\text{sub}}) = \Phi_{\text{event}}(\mathbf{x}_{\text{f}}).
\end{equation}
This two-stage formulation ensures that tactical event forecasting is fundamentally \textit{trajectory-grounded}; the model explicitly simulates the spatial evolution of the play before interpreting its semantic tactical outcome.


\subsection{GenTac Architecture}
\label{subsec:architecture}
To model and forecast complex soccer tactics, GenTac first encodes the historical collective movements of both teams and the ball into a unified sequence of tokens. These tokens are processed by a shared spatiotemporal attention backbone to capture complex multi-agent interactions. Finally, the enriched representations are routed to task-specific decoders: a denoising decoder for trajectory forecasting~($\Psi_{\text{traj}}$) and a semantic decoder for tactical event recognition ($\Psi_{\text{event}}$).

\subsubsection{Multi-Agent Trajectory Encoding}

We first tokenize the raw continuous trajectories to form a unified spatiotemporal representation of the match. The coordinate tokens are then projected into a latent space and augmented with structural embeddings to retain spatial, temporal, and entity-level priors.

\paragraph{Trajectory tokenization.}

An observed trajectory segment of length $l = b - a$ for team $i \in \{1, 2\}$ is represented as $\mathbf{t}^{i}_{[a:b]} \in \mathbb{R}^{l \times N \times 2}$, where $N$ is the number of players per team and 2 represents the 2D pitch coordinate of bird eye view~(BEV). The ball trajectory is similarly represented as $\mathbf{B}_{[a:b]} \in \mathbb{R}^{l \times 1 \times 2}$. By concatenating the trajectories of both teams and the ball, we derive a dense multi-agent representation of shape $l \times (2N+1) \times 2$. We adjust this base representation depending on the downstream task:

For \textbf{trajectory forecasting}, the sequence is extended to a total length $L = l + w$, where $w$ is the future prediction horizon, resulting in $\mathbf{T}_{\mathrm{traj}} \in \mathbb{R}^{L \times (2N+1) \times 2}$. When forecasting both teams jointly, the first $l$ steps contain the observed historical trajectories, while the remaining $w$ steps are initialized with Gaussian noise, representing the corrupted future states to be refined by the diffusion process. When forecasting a single team conditioned on the opponent, noise is appended only to the target team's future steps, while the opponent's future steps are populated with their ground-truth coordinates.

For \textbf{tactical event recognition}, the input is constructed directly from the historical segment, $\mathbf{T}_{\mathrm{event}} \in \mathbb{R}^{l \times (2N+1) \times 2}$. To support batched processing, we enforce a fixed temporal length $L = l_{\max}$ by truncating longer sequences and zero-padding shorter ones. The resulting tensor is $\mathbf{T}_{\mathrm{event}} \in \mathbb{R}^{L \times (2N+1) \times 2}$, with padded positions explicitly masked during attention computation.

\paragraph{Token embedding.}
Let $\mathbf{T} \in \{\mathbf{T}_{\text{traj}}, \mathbf{T}_{\text{event}}\}$ denote the tokenized input. 
Each 2D coordinate token is mapped to a $d$-dimensional latent space via a learnable linear projection, 
producing $\mathbf{H} = \text{proj}(\mathbf{T}) \in \mathbb{R}^{L \times (2N+1) \times d}$. 

To inject crucial structural priors into this permutation-invariant representation, we add three distinct learnable embeddings:
(i) \textbf{temporal embedding} ($\mathbf{e}_{\text{time}} \in \mathbb{R}^{L \times d}$): Shared across all entities at a given time step to encode the chronological sequence.
(ii) \textbf{group embedding} ($\mathbf{e}_{\text{team1}}, \mathbf{e}_{\text{team2}}, \mathbf{e}_{\text{ball}} \in \mathbb{R}^{d}$): Distinguishes the two opposing teams and the ball.
(iii) \textbf{entity embedding} ($\mathbf{e}_{\text{id}} \in \mathbb{R}^{(2N+1) \times d}$): Assigns a unique identifier to each specific player slot and the ball, preserving individual agent continuity across time. 
The resulting representation \(\mathbf{H}\) comprehensively encodes spatial coordinates, temporal order, group affiliation, and individual identity.

\subsubsection{Spatiotemporal Attention Backbone}
Given the unified representation $\mathbf{H}$, we model the complex interactions among players and their temporal evolution using a stack of 
$M$ spatiotemporal attention layers. We adopt the factorized attention architecture from TimeSformer~\cite{bertasius2021TimeSformer}, adapting it from video pixel patches to continuous multi-agent coordinate tokens.

Each layer consists of a spatial attention module ($\mathcal{A}_s$) that captures instantaneous interactions between all players at a single time step, followed by a temporal attention module ($\mathcal{A}_t$) that models the dynamic evolution of each individual player across time. 
To handle invisible entities ({\em e.g.}, players missing due to red cards or tracking failures), we apply an attention mask by assigning a large negative value ($-\infty$) to their pre-softmax attention scores, preventing them from influencing visible entities.

Initializing $\mathbf{H}^{(0)} = \mathbf{H}$, the representation is updated at each layer $m \in \{1, \dots, M\}$ as follows:
\begin{equation}
    \mathbf{H}^{(m)} = \mathcal{A}_t\Big( \mathcal{A}_s\big(\mathbf{H}^{(m-1)}\big) \Big),
\end{equation}
where standard residual connections and layer normalization are applied within $\mathcal{A}_s$ and $\mathcal{A}_t$.
The final output $\mathbf{H}^{(M)} \in \mathbb{R}^{L \times (2N+1) \times d}$ serves as a rich, context-aware representation for the task-specific decoders.

\subsubsection{Task Decoders}

\paragraph{Trajectory forecasting decoder.}
For trajectory forecasting, GenTac employs a denoising decoder $\Psi_{\text{traj}}$ to iteratively recover clean future trajectories from the noise-corrupted representations. Given the contextualized output $\mathbf{H}^{(M)}$, the decoder extracts the features corresponding to the future horizon $w$ and projects them back into the 2D coordinate space. 
Formally, the predicted future trajectory $\hat{\mathbf{x}}_{\text{f}}$ is generated via:
\begin{equation}
    \hat{\mathbf{x}}_{\text{f}} \sim \Psi_{\text{traj}}\big(\mathbf{H}^{(M)}\big).
\end{equation}
In the joint forecasting setting, the decoder outputs $\hat{\mathbf{x}}_{\text{f}} = \{\hat{\mathbf{t}}^1_{[l:l+w]}, \hat{\mathbf{t}}^2_{[l:l+w]}\}$. In the single-team setting, it outputs $\hat{\mathbf{x}}_{\text{f}} = \{\hat{\mathbf{t}}^i_{[l:l+w]}\}$. By conditioning the denoising process on the historical context, the decoder effectively models the conditional distribution of future movements.

\paragraph{Tactical semantic recognition decoder.}
For tactical event recognition, we decode high-level semantic labels directly from the spatiotemporal representation. 
An aggregation operator ($\text{Agg}$), typically mean-pooling across both the temporal and entity dimensions, compresses $\mathbf{H}^{(M)}$ into a single global feature vector $\mathbf{h}_{\text{global}} \in \mathbb{R}^{d}$ that summarizes the entire match segment. 

This global representation is then passed through a semantic decoder $\Psi_{\text{event}}$, implemented as a Multi-Layer Perceptron (MLP) with two parallel classification heads, to predict the event type and subtype:

\begin{equation}
    (\hat{y}^{\text{type}}, \hat{y}^{\text{sub}}) = \Psi_{\text{event}}\big(\text{Agg}(\mathbf{H}^{(M)})\big).
\end{equation}

\subsection{Implementation Details}
\label{subsec:implementation}

In this section, we detail the implementation of the two primary tasks: trajectory forecasting~(Section~\ref{subsubsec:detail_trajectory}) and tactical event recognition~(Section~\ref{subsubsec:detail_event}). The exact hyperparameter configurations and experimental settings are provided in Section~\ref{subsubsec:detail_settings}.

\subsubsection{Trajectory Forecasting}
\label{subsubsec:detail_trajectory}

\paragraph{Causal-window formulation.}
As introduced in Section~\ref{subsubsec:result-trajectory-causal}, GenTac performs continuous trajectory forecasting using a sliding causal window of fixed length $w$. Given historical observations $\mathbf{x}_{\text{h}}$ and a conditioning signal $\mathbf{c}$, GenTac generates the target future segment $\mathbf{x}_{\text{f}}$ for the current window. Long-horizon forecasts are obtained by recursively appending the predicted future to the history and advancing the window. This autoregressive rollout allows the model to operate locally within a computationally tractable window while producing temporally extended, physically consistent futures.

\paragraph{Diffusion objective within a window.}
Within each causal window, we apply a diffusion model to the target future segment $\mathbf{x}_{\text{f}}$, 
while the historical observations $\mathbf{x}_{\text{h}}$ remain uncorrupted to serve as a clean conditional prior. 
All 2D pitch coordinates are linearly normalized to $[-1, 1]$. The forward diffusion process is defined over $S=100$ steps using a linear noise schedule. Specifically, for each diffusion step $s \in \{1, \dots, S\}$, we construct a noisy version of the future trajectory segment:
\begin{equation}
    \mathbf{x}_{\text{f}}^{(s)} = \sqrt{\bar{\alpha}_s}\,\mathbf{x}_{\text{f}} + \sqrt{1-\bar{\alpha}_s}\,\boldsymbol{\epsilon}, \qquad \boldsymbol{\epsilon} \sim \mathcal{N}(\mathbf{0}, \mathbf{I}),
\end{equation}
where $\bar{\alpha}_s = \prod_{r=1}^{s}(1-\beta_r)$, and $\{\beta_r\}_{r=1}^{S}$ follows the linear variance schedule. Here, $\mathbf{x}_{\text{f}}^{(s)}$ denotes the corrupted future trajectory at step $s$. 

The core objective of our conditional denoising network, $\boldsymbol{\epsilon}_\theta(\cdot)$, is to predict the injected noise $\boldsymbol{\epsilon}$ given the corrupted future, the clean history, and the time step:
\begin{equation}
    \hat{\boldsymbol{\epsilon}}^{(s)} = \boldsymbol{\epsilon}_\theta\bigl(\mathbf{x}_{\text{f}}^{(s)}, \mathbf{x}_{\text{h}}, \mathbf{c}, s\bigr).
\end{equation}
The model is optimized using the standard mean squared error (MSE) noise-prediction objective:
\begin{equation}
    \mathcal{L}_{\mathrm{traj}} = \mathbb{E}_{\mathbf{x}_{\text{f}}, \boldsymbol{\epsilon}, s} \left[ \left\| \boldsymbol{\epsilon} - \hat{\boldsymbol{\epsilon}}^{(s)} \right\|_2^2 \right].
\end{equation}
Through this objective, GenTac learns the conditional distribution of future trajectories given $\mathbf{x}_{\text{h}}$ and $\mathbf{c}$.

\paragraph{Inference and causal rollout.}
During inference, GenTac forecasts trajectories via reverse diffusion. For the current causal window, the future segment is initialized with pure Gaussian noise:
\begin{equation}
    \mathbf{x}_{\text{f}}^{(S)} \sim \mathcal{N}(\mathbf{0}, \mathbf{I}).
\end{equation}
Conditioned on $\mathbf{x}_{\text{h}}$ and $\mathbf{c}$, the model iteratively denoises the sequence from step $S$ down to $1$:
\begin{equation}
    \mathbf{x}_{\text{f}}^{(s-1)} = \operatorname{Denoise}\!\left( \mathbf{x}_{\text{f}}^{(s)}, \boldsymbol{\epsilon}_\theta\!\left(\mathbf{x}_{\text{f}}^{(s)}, \mathbf{x}_{\text{h}}, \mathbf{c}, s\right), s \right), \qquad s=S, \dots, 1,
\end{equation}
where $\operatorname{Denoise}(\cdot)$ represents a single reverse diffusion update step. The final predicted future for the window is $\hat{\mathbf{x}}_{\text{f}} = \mathbf{x}_{\text{f}}^{(0)}$. This generated segment $\hat{\mathbf{x}}_{\text{f}}$ is appended to the history, the window shifts forward, and the process repeats. Running this stochastic rollout $K$ times yields a diverse set of plausible multi-agent futures.

\paragraph{Target masking under different forecasting settings.}
As defined in Section~\ref{subsubsec:problem_formulation_trajectory}, we evaluate two primary forecasting settings. For \textbf{joint forecasting}, the future trajectories of both teams are masked and treated as the target $\mathbf{x}_{\text{f}}$. For \textbf{single-team forecasting}, only the target team's future is masked, while the opponent's ground-truth future trajectory within the window is retained as part of the conditional input. The diffusion objective remains identical; only the partition between the target $\mathbf{x}_{\text{f}}$ and the condition $\mathbf{x}_{\text{h}}$ changes.

\paragraph{Pretraining on base forecasting tasks.}
The conditioning vector $\mathbf{c}$ dictates the specific generative task. We first pretrain GenTac on the entire trajectory dataset using two base tasks: (i) \textbf{unconditioned forecasting} (jointly predicting both teams without explicit stylistic conditions) and (ii) \textbf{opponent-conditioned forecasting} (predicting one team given the other). This pretraining phase allows the model to learn fundamental multi-agent kinematics and causal rollout mechanics.

\paragraph{Fine-tuning for conditioned variants.}
Starting from the pretrained weights, we fine-tune GenTac to respond to explicit tactical conditions $\mathbf{c}$. The diffusion objective remains unchanged, but the dataset is partitioned or annotated to reflect specific tactical factors:
(i) \textbf{League-conditioned forecasting}: The unconditioned model is fine-tuned on data from specific leagues to capture distinct regional play styles.
(ii) \textbf{Team-conditioned forecasting}: The opponent-conditioned model is fine-tuned on specific teams to capture unique club-level tactics.
(iii) \textbf{Objective-conditioned forecasting}: The model is fine-tuned on trajectory segments annotated with specific tactical outcomes. Offensive patterns are learned from segments resulting in threats ({\em e.g.}, {\em goal}, {\em shot on target}), while defensive patterns are learned from successful stops ({\em e.g.}, {\em clearance}, {\em tackle}). This novel conditioning scheme allows GenTac to generate futures that actively pursue specified tactical goals.

\subsubsection{Tactical Event Grounding and Forecasting}
\label{subsubsec:detail_event}

\paragraph{Aggregation pooling.}
Following the spatiotemporal attention backbone, the semantic branch aggregates the tokenized representation $\mathbf{T}_{\mathrm{event}}$ into a single event-level feature vector. Let $\mathbf{H}^{(M)} = \{\mathbf{h}_{t,n}\}$ denote the output tokens from the final encoder layer, 
where $t$ indexes time and $n$ indexes entities. Because tactical events are often defined by a sparse subset of key players and moments, we introduce a lightweight attention pooling module, $\mathrm{Agg}(\cdot)$, to compress the spatiotemporal grid into a global representation $\mathbf{z}$. Each token is assigned a learned scalar importance score, normalized via softmax:
\begin{equation}
    a_{t,n} = \frac{\exp(\mathrm{MLP}(\mathbf{h}_{t,n}))}{\sum_{t',n'} \exp(\mathrm{MLP}(\mathbf{h}_{t',n'}))}, \qquad \mathbf{z} = \sum_{t,n} a_{t,n}\mathbf{h}_{t,n}.
\end{equation}

\paragraph{Event grounding.}
For event grounding, the semantic head maps the observed historical trajectories $\mathbf{x}_{\text{h}}$ directly to tactical labels. The label space is hierarchical, comprising 5 broad event types and 15 specific subtypes. Given the pooled representation $\mathbf{z}$, a primary classifier $f_{\mathrm{type}}$ predicts the event type $\hat{y}^{\mathrm{type}}$. This prediction routes the feature to the corresponding subtype classifier $f_{\mathrm{sub}}^{(\cdot)}$:
\begin{equation}
    \hat{y}^{\mathrm{type}} = \arg\max f_{\mathrm{type}}(\mathbf{z}), \quad \hat{y}^{\mathrm{sub}} = \arg\max f_{\mathrm{sub}}^{(\hat{y}^{\mathrm{type}})}(\mathbf{z}).
\end{equation}
This hierarchical design prevents incompatible subtypes from competing during optimization. During training, we compute the cross-entropy (CE) loss for both levels simultaneously:
\begin{equation}
    \mathcal{L}_{\mathrm{event}} = \mathcal{L}_{\mathrm{type}} + \lambda \mathcal{L}_{\mathrm{sub}},
\end{equation}
where $\mathcal{L}_{\mathrm{type}} = \mathrm{CE}\big(\hat{y}^{\mathrm{type}}, y^{\mathrm{type}}\big)$ and $\mathcal{L}_{\mathrm{sub}} = \mathrm{CE}\big(\hat{y}^{\mathrm{sub}}, y^{\mathrm{sub}}\big)$, with $\lambda$ balancing the two terms.

\paragraph{Event forecasting.}
For event forecasting, GenTac first generates the future trajectories using the diffusion decoder, and then applies the event recognition head to these generated futures:
\begin{equation}
    \hat{\mathbf{x}}_{\text{f}} = \Phi_{\mathrm{traj}}(\mathbf{x}_{\text{h}}, \mathbf{c}), \qquad (\hat{y}^{\mathrm{type}}, \hat{y}^{\mathrm{sub}}) = \Phi_{\mathrm{event}}(\hat{\mathbf{x}}_{\text{f}}).
\end{equation}
Unlike prior works that extrapolate labels directly from history, GenTac explicitly grounds future events in forecasted physical movements. Because the trajectory forecaster is probabilistic, sampling multiple futures from the same history naturally yields a distribution of potential tactical outcomes.

\subsubsection{Experiment Settings}
\label{subsubsec:detail_settings}

\paragraph{Shared configurations.}
All experiments in this work are conducted on a single NVIDIA A100 GPU for both training and inference. The trajectory forecasting and tactical event recognition tasks share the same spatiotemporal attention backbone. Specifically, the encoder consists of $M=4$ attention layers, each utilizing 8 attention heads, with a constant hidden dimension of $d=256$. For all soccer-related experiments, the number of modeled players per team is set to $N=11$, in accordance with standard soccer regulations. Detailed statistics of the trajectory datasets and their respective data splits are provided in Supplementary~\ref{appendix:dataset}.

\paragraph{Trajectory forecasting settings.}
For the trajectory forecasting task, the diffusion process is discretized into 100 steps. The model is optimized using the AdamW optimizer~\cite{adamW} with a peak learning rate of $1 \times 10^{-3}$ and a weight decay of $1 \times 10^{-4}$. The learning rate follows a cosine decay schedule with a linear warm-up ratio of $0.02$. Unless otherwise specified, models are trained for 60 epochs with a batch size of 200, and the checkpoint achieving the lowest validation loss is selected for evaluation. As detailed in Section~\ref{subsec:result-trajectory}, most forecasting experiments operate with a causal window of $0.2\,\mathrm{s}$ and a historical observation horizon of $4\,\mathrm{s}$. 
The denoising network ($\boldsymbol{\epsilon}_\theta$) employs a lightweight noise-prediction head that maps the 256-dimensional latent representation to a 2-dimensional noise residual via a normalization layer followed by a linear projection. For generalization experiments on other team sports, the number of players per team is adjusted according to the respective sport's rules: $N=5$ for basketball, $N=11$ for American football, and $N=6$ for ice hockey.

\paragraph{Tactical event grounding and forecasting settings.}
For both event grounding and event forecasting, each sample is represented as a trajectory sequence of fixed length $l_{\max}=250$ frames ($10\,\mathrm{s}$). Because event modeling is defined at the team level, each sample continuously tracks $N=11$ players. During training, data augmentation is limited to random horizontal and vertical spatial flipping, applied with a probability of $0.5$. The model is optimized using the standard Adam optimizer~\cite{diederik2014adam}. The initial learning rate is set to $5 \times 10^{-4}$ when training from scratch and $1 \times 10^{-4}$ for fine-tuning, with a constant weight decay of $1 \times 10^{-4}$. We use a batch size of 32 per GPU and train for up to 200 epochs. Gradients are clipped to a maximum $L_2$ norm of $1.0$, and early stopping is triggered if validation accuracy does not improve for 35 consecutive epochs. For the hierarchical classification objective, the balancing weight is set to $\lambda=1.0$. In the event forecasting setting, the predicted trajectory spanning the future 100 frames ($4\,\mathrm{s}$) serves as the input to the classification head.

\section{Data Availability}
The trajectory dataset and the TacBench benchmark utilized in this study are deposited at \url{https://github.com/jyrao/GenTac/tree/main/data} under a \textit{CC BY-NC-SA} license.

\section{Code Availability}
The complete source code for reproducing the experiments in this paper is publicly available at \url{https://github.com/jyrao/GenTac} under a \textit{CC BY-SA} license, and should be cited as~\cite{Our_code}.

\clearpage
\bibliographystyle{unsrt}
\bibliography{references} 

\clearpage

\section{Acknowledgments}
Weidi would like to acknowledge the funding from Scientific Research Innovation Capability Support Project for Young Faculty~(ZY-GXQNJSKYCXNLZCXM-I22).

\section{Author Contributions}

In this work, W.X. is the corresponding author, J.R., T.G, H.W, Y.W., and W.X. all make contributions to the conception or design of the work, and J.R., T.G. further perform acquisition, analysis, or interpretation of data for the work. In writing,  J.R. draft the work. T.G, H.W, Y.W., and W.X. review it critically for important intellectual content. All authors approve of the version to be published and agree to be accountable for all aspects of the work to ensure that questions related to the accuracy or integrity of any part of the work are appropriately investigated and resolved.

\section{Competing Interests}
The authors declare no competing interests.




\clearpage

\section{Supplementary}
\setcounter{table}{0}   
\setcounter{figure}{0}
\renewcommand{\tablename}{Supplementary Table}
\renewcommand{\figurename}{Supplementary Figure}

\subsection{More Details about Datasets}
\label{appendix:dataset}
To support the training and evaluation of GenTac under various conditioning settings, we construct a unified dataset and benchmark for multi-player trajectory forecasting and tactical event analysis. This benchmark enables the rigorous evaluation of trajectory forecasting under different conditioning mechanisms ({\em e.g.}, opponent, team, league, and objective), as well as tactical event grounding and forecasting. 

In Section~\ref{appendixsub:datasource}, we describe the sources of all trajectory data used in this work, covering both soccer and other team sports. In Section~\ref{appendixsub:datacuration}, we detail the data curation pipeline that converts the raw trajectory data into the standardized benchmark format used in our experiments. Finally, in Section~\ref{appendixsub:datastat}, we present the trajectory dataset split strategy and summarize the resulting statistics, while Section~\ref{appendixsub:event} details the tactical event definitions, data sources, and curation pipeline.

\subsubsection{Trajectory Data Source.} 
\label{appendixsub:datasource}
Our dataset integrates both publicly available sports tracking datasets and trajectories reconstructed from broadcast videos using state-of-the-art game state reconstruction pipelines. For training and evaluating soccer tactics forecasting, we collected the following two types of data:

\vspace{-4pt}
\begin{itemize}
    \setlength \itemsep{5pt}
    \item \textbf{Public soccer trajectory and event datasets:}  
        Full-match trajectories for all players and the ball, with synchronized event labels from professional leagues. This includes {\em Metrica Sports}~\cite{metrica_sports_sample_data} (3 matches), {\em SkillCorner}~\cite{skillcorner_opendata} from the Australian A-League (10 matches), and {\em Sportec DFL}~\cite{bassek2025integrated} from the German Bundesliga and 2.\ Bundesliga (7 matches).
    \item \textbf{Broadcast-derived data:}  
        Multi-player trajectories reconstructed from broadcast videos. This comprises \textit{SoccerNet-GSR}~\cite{SoccerNet-GSR} with 164 annotated 30-second clips providing trajectories only, and a curated subset generated using the \textit{SoccerFactory}~\cite{yang2025soccermaster} pipeline on broadcast videos from \textit{SoccerReplay-1988}~\cite{rao2025unisoccer}. This yields 363 variable-length clips with both trajectories and aligned event labels.
\end{itemize}

To evaluate the generalization capabilities of GenTac, we also incorporated publicly available multi-player trajectory datasets from other team sports:
\vspace{-4pt}
\begin{itemize}
    \setlength \itemsep{4pt}
    \item \textbf{Basketball:} {\em NBA SportVU}~\cite{nba_sportvu_tracking}, comprising 66,498 clips of 6-second duration.
    \item \textbf{American football:} {\em NFL Big Data Bowl 2021}~\cite{nfl_big_data_bowl_official}, consisting of offensive plays exceeding 10 yards from 256 games.
    \item \textbf{Ice hockey:} {\em Big Data Cup 2026}~\cite{bigdatacup_2026_data_release}, including 33 games segmented into three 20-minute periods.
\end{itemize}

\subsubsection{Trajectory Data Curation.} 
\label{appendixsub:datacuration}

A unified preprocessing pipeline is employed to harmonize heterogeneous raw trajectory data into a consistent structural paradigm.

\paragraph{Public soccer trajectory data.}
Raw trajectory data from heterogeneous public sources ({\em e.g.}, XML, JSON, CSV) are standardized through a unified preprocessing pipeline. To resolve inherent discrepancies in original sampling rates, temporal interpolation is applied to align all discrete time steps into a synchronized temporal sequence at a standardized frame rate of 25 frames per second (fps).

For compatibility with the diffusion models, the continuous trajectory is discretized into a frame-by-frame dictionary sequence. Each temporal frame is indexed by a sequential integer key, capturing the instantaneous 2D spatial coordinates of all tracked entities in meters. 
Players are partitioned into two opposing groups (\texttt{team0} and \texttt{team1}), preserving their original alphanumeric identifiers to maintain identity consistency across frames, while the ball is tracked as an independent entity. Missing tracking data for any entity at a given frame is uniformly imputed as a null coordinate pair.

Spatial coordinates are normalized to a standard $105 \times 68$ m pitch. The origin $(0, 0)$ is anchored at the center spot, bounding the longitudinal ($x$) and lateral ($y$) axes to the intervals $[-52.5, 52.5]$ and $[-34, 34]$, respectively. All positional values are rounded to two decimal places. A representative data structure is provided below:

\begin{lstlisting}[breaklines=true, basicstyle=\ttfamily\footnotesize]
{
  "13590": {"ball": [6.50, 4.20], "team0": {"Player1": [-0.74, -30.28], "Player10": [-0.06, 9.86], "Player11": [-42.91, 0.74], "Player2": [-12.45, -8.44], ...}, "team1": {"Player15": [-25.95, 10.47], "Player16": [18.10, -0.48], "Player17": [17.50, -7.31], ...}},
  "13591": {"ball": [null, null], "team0": {"Player1": [-0.73, -30.41], "Player10": [-0.37, 11.14], "Player11": [-42.88, 0.83], "Player2": [-12.54, -8.62], ...}, "team1": {"Player15": [-25.86, 10.47], "Player16": [18.11, -0.70], "Player17": [17.46, -7.53], ...}},
  ...
}
\end{lstlisting}

\paragraph{Broadcast-derived soccer trajectory data.}

To acquire broadcast-derived player trajectories, we extract the main camera center-view frames from the SoccerNet-v2 dataset~\cite{SoccerNetv2} and process them using the SoccerFactory pipeline~\cite{yang2025soccermaster}. The pipeline detects players, referees, and the ball from broadcast frames and projects their image locations onto pitch coordinates, producing frame-wise entity positions and track identities. To improve the temporal consistency of these projected trajectories, we apply the following post-refinement procedures. The quantitative performance of this pipeline is summarized in Supplementary Table~\ref{supptab:performance_gsr_pipeline}.

\vspace{-5pt}
\begin{itemize}
\setlength \itemsep{5pt}
    \item \textbf{Duplicate identity resolution:} Within each frame, if multiple detections from the same team share an identical jersey number, only one instance is retained. Conflicting detections are suppressed to avoid identity ambiguity.
    \item \textbf{Gap interpolation:} Short missing segments for tracked entities (players, goalkeepers, and referees) are filled via linear interpolation between the nearest valid observations.
    \item \textbf{Global anomaly detection:} Pairs of adjacent frames are examined for abnormal motion kinematics. If multiple tracked entities exhibit physically implausible position changes between frames, the corresponding frames are marked as potential anomaly segments.
    \item \textbf{Segment reconstruction:} Short abnormal segments are reconstructed by interpolating entity positions between the nearest valid frames, utilizing track identities when available to maintain trajectory continuity.
    \item \textbf{Temporal smoothing:} Finally, the reconstructed trajectories are smoothed using a bidirectional exponential moving average to suppress residual high-frequency jitter while preserving overall motion trends.
\end{itemize}

\begin{table}[h]
    \vspace{-3pt}
    \centering
    \footnotesize
    \resizebox{.7\linewidth}{!}{
        \begin{tabular}{l | c c c}
        \toprule
        \textbf{Method} & \textbf{GS-HOTA} & \textbf{GS-DetA} & \textbf{GS-AssA} \\
        \midrule
        Playbox \& MIXI & 58.1 & 41.3 & \textbf{81.6} \\
        Metrica-Sports & 58.2 & 44.4 & 76.2 \\
        KIST-GSR & 61.5 & 48.5 & 78.0 \\
        SoccerFactory~\cite{yang2025soccermaster} & \underline{64.1} & \textbf{51.5} & 79.9 \\
        \midrule
        \textbf{Ours} (SoccerFactory + post-refinement) & \textbf{64.2} & \underline{51.3} & \underline{80.5} \\
        \bottomrule
        \end{tabular}
    }
    \vspace{5pt}
    \caption{
        \textbf{Comparison on GSR Task according to SoccerNet 2025 Challenges Results~\cite{giancola2025soccernet2025challengesresults}.} 
        Compared to the previous sota model SoccerFactory~\cite{yang2025soccermaster}, our refinement keeps promoting the performance of GSR task on SoccerNet-GSR~\cite{SoccerNet-GSR} test set. The best and second-best results are \textbf{bolded} and \underline{underlined}.}
    \label{tab:performance_gsr_pipeline}
\end{table}

Following the application of the SoccerFactory pipeline~\cite{yang2025soccermaster} and the aforementioned post-refinement strategies, we extracted multi-player trajectories intended for downstream tactical reasoning tasks. To guarantee the positional reliability of these broadcast-derived trajectories, we adopt a manual annotation and verification protocol. Expert annotators systematically reviewed the dataset to identify and discard trajectory clips exhibiting physically implausible positional jumps.

As illustrated in Supplementary Fig.~\ref{fig:annotation_interface}, annotators utilized a custom graphical interface to efficiently filter out these anomalies. These positional discontinuities predominantly arise from camera calibration failures during rapid broadcast camera panning or zooming. Specifically, when field markings and keypoints are heavily occluded, the frame-to-frame homography estimation (transformation matrix) from the broadcast view to the 2D pitch coordinates becomes ill-conditioned. This instability induces unnatural coordinate shifts on the pitch plane, even when the underlying image-space multi-object tracking remains robust.

\begin{figure}[H]
    \centering
    \includegraphics[width=\linewidth]{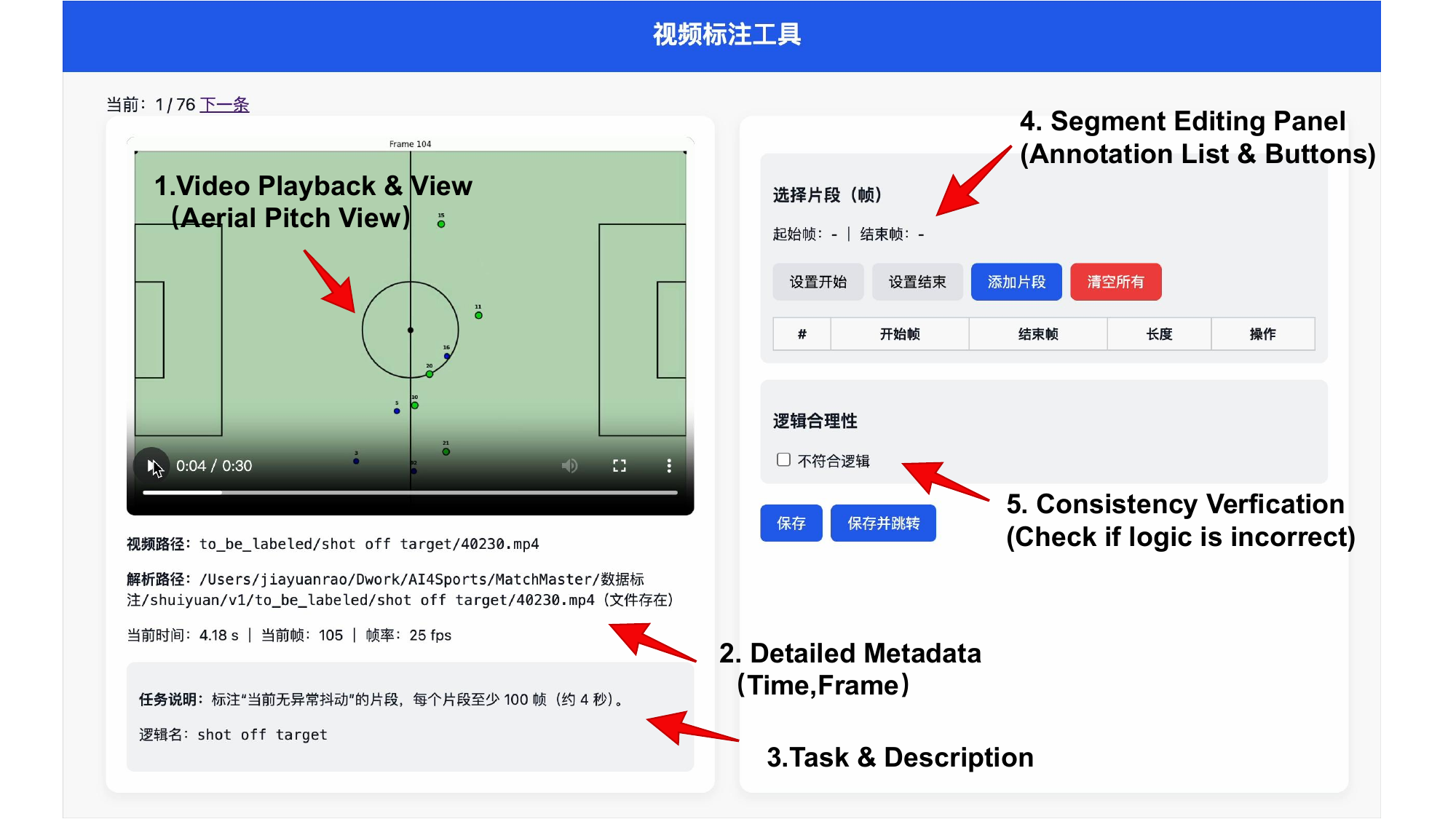}
    \vspace{6pt}
    \caption{\textbf{Manual data annotation interface.} The original Chinese layout is overlaid with red English labels to highlight core components.  Volunteers execute the annotation process through the following steps: \textbf{Step 1:} Watch tactical plays in the video playback view ``1"; \textbf{Step 2:} Identify target sequences based on the specific task and description ``3" \textbf{Step 3:} Utilize detailed metadata ``2" to pinpoint exact start and end frames within the segment editing panel ``4". To ensure dataset quality, a consistency check feature ``5" allows annotators to filter out anomalous or noisy data, such as abnormal positional jumps.}
    \label{fig:annotation_interface}
\end{figure}

\paragraph{Trajectory data of other team sports.}
To evaluate the cross-domain generalization of our generative modeling framework, we extended our data curation pipeline to three additional team sports: basketball, American football, and ice hockey. The raw tracking data from these domains were standardized into the identical frame-by-frame dictionary format utilized for the soccer dataset. This unified structure records the 2D spatial coordinates of the primary game object (ball or puck) and all active players from both teams over time. Consistent with the soccer pipeline, missing tracking observations are imputed as null coordinates.

To accommodate the domain-specific characteristics of each sport, we applied two primary structural adjustments. First, the expected number of tracked entities was adapted to reflect the respective official regulations ({\em e.g.}, 5-vs-5 for basketball, 11-vs-11 for American football, and 6-vs-6 for ice hockey). Second, the spatial coordinates were normalized to the standard playing surface dimensions of each sport. We preserved the native sampling rates of the source datasets to evaluate the model's robustness to varying temporal resolutions: 5 frames per second (fps) for basketball, 10 fps for American football, and 30 fps for ice hockey.
An example of the basketball data structure is shown below:
\begin{lstlisting}[breaklines=true, basicstyle=\ttfamily\footnotesize]
{
  "10420": {"ball": [6.50, 4.20], "team0": {"GS_Curry": [6.10, 3.80], "GS_Thompson": [4.50, -6.10], "GS_Green": [0.50, 2.10], "GS_Wiggins": [-3.20, 5.50], "GS_Looney": [-1.00, -1.50]}, "team1": {"LAL_James": [5.80, 3.50], "LAL_Davis": [-0.80, -1.20], "LAL_Reaves": [4.20, -5.80], "LAL_Russell": [-2.90, 5.20], "LAL_Hachimura": [0.20, 1.80]}},
  "10421": {"ball": [6.85, 2.50], "team0": {"GS_Curry": [6.25, 3.60], "GS_Thompson": [4.55, -6.05], "GS_Green": [0.65, 2.15], "GS_Wiggins": [-3.10, 5.45], "GS_Looney": [-0.95, -1.45]}, "team1": {"LAL_James": [6.00, 3.30], "LAL_Davis": [-0.75, -1.15], "LAL_Reaves": [4.25, -5.75], "LAL_Russell": [-2.85, 5.15], "LAL_Hachimura": [0.25, 1.85]}},
  ...
}
\end{lstlisting}

\subsubsection{Trajectory Dataset Split and Statistics.} 
\label{appendixsub:datastat}
The dataset is partitioned into training, validation, and test splits. For soccer, public data are distributed across all three splits, whereas broadcast-derived trajectories are restricted to training and validation. Consequently, the final test benchmark relies exclusively on public data.
Table \ref{tab:dataset_stats} summarizes the quantitative distribution across all curated sports datasets.

\begin{table}[htbp]
    \centering
    \scriptsize
    \setlength{\tabcolsep}{3.5pt}
    \renewcommand{\arraystretch}{1.08}
    \begin{threeparttable}
    \renewcommand{\tabularxcolumn}[1]{m{#1}}
    \begin{tabularx}{\textwidth}{
        >{\raggedright\arraybackslash}p{2.9cm}
        >{\raggedright\arraybackslash}p{4.2cm}
        >{\raggedright\arraybackslash}p{2.6cm}
        >{\centering\arraybackslash}p{0.9cm}
        >{\centering\arraybackslash}p{0.9cm}
        >{\centering\arraybackslash}p{0.9cm}
        >{\centering\arraybackslash}p{0.9cm}
        >{\centering\arraybackslash}p{1.0cm}
    }
        \toprule
        \textbf{Sport Category} & \textbf{Dataset} & \textbf{Source} & \textbf{Train} & \textbf{Valid} & \textbf{Test} & \textbf{Total} & \textbf{Unit} \\
        \midrule
        
        \multirow{5}{*}{Soccer} 
        & Metrica Sports~\cite{metrica_sports_sample_data} & Public Dataset & 1 & 0.5\tnote{*} & 0.5\tnote{*} & 2 & matches \\
        & SkillCorner Open Data~\cite{skillcorner_opendata} & Public Dataset & 8 & 1 & 1 & 10 & matches \\
        & DFL Bundesliga~\cite{bassek2025integrated} & Public Dataset & 5 & 1 & 1 & 7 & matches \\
        & SoccerNet-GSR~\cite{SoccerNet-GSR} & Broadcast-derived & 61 & 62 & -- & 123 & clips \\
        & SoccerFactory+refinement~\cite{yang2025soccermaster}$^\dagger$ & Broadcast-derived & -- & -- & -- & 353 & clips \\
        \midrule
        
        Basketball 
        & NBA SportVU~\cite{nba_sportvu_tracking} & Public Dataset & 42,054 & 16,046 & 8,398 & 66,498 & clips \\
        \midrule
        
        American Football 
        & NFL Big Data Bowl~\cite{nfl_big_data_bowl_official} & Public Dataset & 13,458 & 3,797 & 1,899 & 19,154 & clips \\
        \midrule
        
        Ice Hockey 
        & Big-Data-Cup~\cite{bigdatacup_2026_data_release} & Public Dataset & 23 & 6 & 4 & 33 & periods \\
        
        \bottomrule
    \end{tabularx}
    
    \begin{tablenotes}
        \footnotesize
        \item[*] The value 0.5 denotes one half of a soccer match.
        \item[$\dagger$] The dataset created by ourselves in this work.
    \end{tablenotes}
    \vspace{0.15cm}
    \caption{\textbf{Quantitative distribution of trajectory datasets.} 
    The table summarizes the allocation of data across training, validation, and test splits for all team sports. 
    As for the "source" column, "Public Dataset" represents the data from website that has already collected in the form of pitch coordinates, while "Broadcast-derived" represents the trajectory data generated from soccer broadcast video.
    The temporal scope of each dataset is defined in the "Unit" column. "Matches" indicates complete games, "clips" denotes shorter tracking sequences such as specific events or fixed-length segments, and "periods" refers to the standard 20-minute intervals specific to ice hockey.}
    \label{tab:dataset_stats}
    \end{threeparttable}
\end{table}

\subsubsection{Event Curation and Split.}
\label{appendixsub:event}
\paragraph{Data source and curation.}
The event dataset aggregates three sources: SkillCorner Open Data~\cite{skillcorner_opendata}, DFL Bundesliga~\cite{bassek2025integrated}, and  Metrica Sports~\cite{metrica_sports_sample_data}. To harmonize heterogeneous event data, a unified classification scheme consolidates 15 subtypes into five macro-categories: Build, Transition, Threat, Set Piece and Interruption.

For most events, we rely on the original labels provided by each dataset. However, the original data can be unclear for complex tactical phases, especially within the Build, Transition, and Threat categories. To ensure accuracy in these cases, we add our own manual labels to the automated process. We use the following mapping scheme to align the different datasets:

\begin{tcolorbox}[colback=gray!2, colframe=black!80, title=\textbf{Event Mapping Scheme}, fonttitle=\bfseries, fontupper=\scriptsize, arc=1mm, boxrule=0.5pt, left=2mm, right=2mm, top=2mm, bottom=2mm]

\textbf{1. Build} \\
\textit{Definition: Phases focused on controlled possession and structural advancement originating from the defensive zones.}
\begin{itemize}[leftmargin=*, nosep, itemsep=3pt]
    \item \textbf{Build}: The team progresses from the defensive third via patient, structured passing sequences while maintaining secure possession. \textbf{Criteria}: Filtered via \texttt{PASS} originating in the defensive third requiring forward progression $< 0.05$ and defensive centroid stability $\sigma < 0.1$ (Metrica, Bundesliga); successful \texttt{PASS} originating in the defensive third with high spatial separation ($\ge 3.0$) during build-up and create phases (Open Data).
\end{itemize}

\vspace{2.5mm}\textbf{2. Transition} \\
\textit{Definition: Phases characterized by dynamic shifts in possession and rapid vertical advancements across the pitch.}
\begin{itemize}[leftmargin=*, nosep, itemsep=3pt]
    \item \textbf{Ball Win}: Midfield interceptions and recoveries aimed at regaining possession from the opponent. \textbf{Criteria}: Midfield \texttt{CHALLENGE} (Metrica); \texttt{start\_type} $\in \{\texttt{pass\_interception}, \texttt{recovery}\}$ with separation $< 3.0$ indicating high pressure (Open Data); midfield \texttt{BallClaiming} (Bundesliga).
    \item \textbf{Progression}: Intentional forward advancements originating outside the defensive third ($x > 0.33$) to drive the attack. \textbf{Criteria}: Satisfying crossing midfield ($\zeta > 0.1$), deep progression ($\zeta > 0.2$), or penalty box entry; spatial filtration of \texttt{PASS}/\texttt{BALL LOST} (Metrica); spatial filtration of transition phases (Open Data); spatial filtration of \texttt{Play}/\texttt{Run} (Bundesliga).
\end{itemize}

\vspace{2.5mm}\textbf{3. Threat} \\
\textit{Definition: Final-stage actions involving direct scoring attempts, critical box defenses, or high-danger area interventions.}
\begin{itemize}[leftmargin=*, nosep, itemsep=3pt]
    \item \textbf{Goal}: Successful scoring events where the ball legally crosses the goal line. \textbf{Criteria}: \texttt{SHOT ON TARGET-GOAL} or \texttt{WOODWORK-GOAL} (Metrica); \texttt{game\_interruption\allowbreak\_before} $\in \{\texttt{goal\_for}, \allowbreak \texttt{goal\_against}\}$ (Open Data); \texttt{SuccessfulShot} (Bundesliga).
    \item \textbf{Shot Off Target}: Unsuccessful scoring attempts that miss the goal frame and are not saved by the goalkeeper. \textbf{Criteria}: \texttt{OFF TARGET}/\allowbreak\texttt{WOODWORK} (Metrica); \texttt{end\_type} == \texttt{shot} excluding actual goals (Open Data); \texttt{Shotwide}/\allowbreak\texttt{OtherShot}/\allowbreak\texttt{ShotWoodWork} (Bundesliga).
    \item \textbf{Shot Saved}: Scoring attempts directed on target but successfully intercepted by the goalkeeper. \textbf{Criteria}: \texttt{SAVED} (Metrica); non-goal target shots (Open Data); \texttt{SavedShot} (Bundesliga).
    \item \textbf{Clearance}: Defensive actions aimed primarily at safely clearing the ball from critical danger zones (e.g., the defensive third). \textbf{Criteria}: \texttt{CLEARANCE} or \texttt{PASS}/\texttt{BALL LOST+} in the defensive third (Metrica); \texttt{end\_type} == \texttt{clearance} (Open Data); verified via manual annotation.
    \item \textbf{Defense}: Successful defensive interventions actively disrupting attacks within the penalty area. \textbf{Criteria}: Spatial filtration of \texttt{BALL LOST} or \texttt{CHALLENGE} inside the penalty area ($x \ge 0.8$) (Metrica); disruption phases or blocked shots verified via manual annotation (Open Data, Bundesliga).
\end{itemize}

\vspace{2.5mm}\textbf{4. Set Piece} \\
\textit{Definition: Restarts of play from dead-ball situations governed by specific game rules.}
\begin{itemize}[leftmargin=*, nosep, itemsep=3pt]
    \item \textbf{Corner}: A restart of play awarded to the attacking team when the ball completely crosses the goal line after last touching a defending player. \textbf{Criteria}: \texttt{SET PIECE+CORNER KICK} (Metrica); \texttt{start\_type} $\in \{\texttt{corner\_reception}, \texttt{corner\_interception}\}$ (Open Data); \texttt{CornerKick} (Bundesliga).
    \item \textbf{Free Kick}: A restart of play following a foul or infringement by the opposing team. \textbf{Criteria}: \texttt{SET PIECE+FREE KICK} (Metrica); \texttt{start\_type} $\in \{\texttt{free\_kick\_reception}, \texttt{free\_kick\_interception}\}$ (Open Data); \texttt{Freekick} (Bundesliga).
    \item \textbf{Penalty}: A direct shot taken from the penalty mark, awarded for a foul committed inside the penalty area. \textbf{Criteria}: \texttt{SET PIECE+PENALTY} (Metrica); \texttt{game\_interruption\_before} $\in \{\texttt{penalty\_for}, \texttt{penalty\_against}\}$ (Open Data); \texttt{Penalty} (Bundesliga).
    \item \textbf{Throw-in}: A restart of play  when the ball completely crosses the touchline. \textbf{Criteria}: \texttt{THROW IN} (Metrica); \texttt{start\_type} == \texttt{throw\_in\_reception} (Open Data); \texttt{ThrowIn} (Bundesliga).
    \item \textbf{Kick-off}: The method of starting or restarting play at the beginning of a half or immediately after a goal is scored. \textbf{Criteria}: \texttt{KICK OFF} (Metrica); \texttt{game\_interruption\_before} $\in \{\texttt{goal\_for}, \texttt{goal\_against}\}$ (Open Data); \texttt{Kickoff} (Bundesliga).
    \item \textbf{Goal Kick}: A restart of play awarded to the defending team when the ball completely crosses the goal line after last touching an attacking player. \textbf{Criteria}: \texttt{GOAL KICK} (Metrica); \texttt{start\_type} $\in \{\texttt{goal\_kick\_reception}, \texttt{goal\_kick\_interception}\}$ (Open Data); \texttt{GoalKick} (Bundesliga).
\end{itemize}

\vspace{2.5mm}\textbf{5. Interruption} \\
\textit{Definition: Non-active play periods caused by rule infractions, ball out of bounds, or administrative stoppages.}
\begin{itemize}[leftmargin=*, nosep, itemsep=3pt]
    \item \textbf{Interruption}: Temporary game stoppages primarily due to rule infractions, substitutions, or the ball going out of bounds. \textbf{Criteria}: \texttt{CARD}/\texttt{FAULT RECEIVED} (Metrica); \texttt{end\_type} $\in \{\texttt{foul\_suffered}, \texttt{foul\_committed}\}$ (Open Data); \texttt{Foul}/\texttt{Substitution} (Bundesliga).
\end{itemize}
\end{tcolorbox}

Temporal extraction employs a Safety Center Window mechanism. Event boundaries are initially buffered by 1 seconds preceding and succeeding the action. To mitigate visual contamination between adjacent events, dynamic collision avoidance truncates overlapping windows: left boundaries are constrained to follow the preceding event's termination, and right boundaries precede the subsequent event's initiation. Compressed segments falling below minimum length thresholds (20 frames, 0.8s) receive random temporal padding, whereas prolonged sequences are capped at 25-50 frames. High-value events, including free kicks, goals and shot off targets, receive forced boundary extensions by an additional 1.0 seconds preceding the action and 0.6 seconds succeeding its termination to preserve tactical context.
\paragraph{Data split and statistics.}
To mitigate the uneven distribution of rare events across individual matches, extracted instances are aggregated globally into a unified pool before partitioning. This composite dataset is then allocated into training, validation, and test splits adhering to an approximate 7:2:1 ratio. The finalized event dataset comprises 4,284 samples , totalling 3.66 hours of effective duration, which are distributed into 3,010 (2.55 hours), 851 (0.73 hours), and 423 (0.37 hours) sequences for training, validation, and testing respectively. 
Supplementary Table \ref{tab:event_stats} details the quantitative distribution and average temporal characteristics across all evaluated subtypes.

\begin{table}[htbp]
    \centering
    \scriptsize
    \setlength{\tabcolsep}{3.5pt}
    \renewcommand{\arraystretch}{1.08}
    \begin{tabular*}{\textwidth}{@{\extracolsep{\fill}} l l c c c c c @{}}
        \toprule
        \textbf{Macro-category} & \textbf{Subtype} & \textbf{Total} & \textbf{Train} & \textbf{Valid} & \textbf{Test} & \textbf{Avg. Duration (s)} \\
        \midrule
        Build-up & Build & 472 & 331 & 94 & 47 & 3.04 \\
        \midrule
        \multirow{2}{*}{Transition} 
        & Ball Win & 1,566 & 1,097 & 313 & 156 & 2.09 \\
        & Progression & 898 & 630 & 179 & 89 & 3.20 \\
        \midrule
        \multirow{6}{*}{Threat} 
        & Goal & 33 & 24 & 6 & 3 & 6.32 \\
        & Shot Off Target & 117 & 83 & 23 & 11 & 5.94 \\
        & Shot Saved & 70 & 49 & 14 & 7 & 2.86 \\
        & Clearance & 119 & 85 & 23 & 11 & 2.85 \\
        & Defended & 84 & 60 & 16 & 8 & 1.95 \\
        \midrule
        \multirow{6}{*}{Set Piece} 
        & Corner & 85 & 60 & 17 & 8 & 5.86 \\
        & Free Kick & 173 & 122 & 34 & 17 & 5.61 \\
        & Penalty & 6 & 2 & 2 & 2 & 7.99 \\
        & Throw-in & 408 & 287 & 81 & 40 & 2.65 \\
        & Kick-off & 24 & 18 & 4 & 2 & 3.90 \\
        & Goal Kick & 38 & 28 & 7 & 3 & 3.05 \\
        \midrule
        Interruption & Stoppage & 191 & 134 & 38 & 19 & 6.17 \\
        \midrule
        \textbf{Total} &  & \textbf{4,284} & \textbf{3,010} & \textbf{851} & \textbf{423} & \textbf{3.08} \\
        \bottomrule
    \end{tabular*}
    \vspace{0.25cm} 
    \caption{\textbf{Quantitative distribution of event subtypes.} The table summarizes the allocation of the 4,284 extracted events across training, validation, and test splits (approximating a 7:2:1 ratio). To prevent severe class imbalance at the individual match level, partitioning is executed globally across the aggregated event pool.}
    \label{tab:event_stats}
\end{table}

\subsection{More Details about Metrics}
\label{appendix:metric}
In this paper, we evaluate our proposed model using a comprehensive suite of metrics designed to capture multi-faceted tactical behaviors and spatial dynamics. Specifically, in Section~\ref{appendixsub:geometrical}, we detail the distance-based metrics used to measure the geometric and spatial precision of the forecasted trajectories. In Section~\ref{appendixsub:structure}, we introduce a variety of metrics to assess team organizational coherence and spatial coordination. In Section~\ref{appendixsub:semantic}, we outline the classification metrics utilized for evaluating the accuracy of hierarchical semantic event predictions. Finally, in Section~\ref{appendixsub:performance}, we present advanced analytical metrics to quantify both dynamic offensive threats and defensive structural integrity.

\subsubsection{Geometrical Accuracy of Forecasted Trajectory.} 
\label{appendixsub:geometrical}
The average displacement error (ADE) and the final displacement error (FDE) are two metrics used to measure the spatial precision of the forecasted trajectories. 
\begin{itemize}
  \setlength \itemsep{5pt}
  \item \textbf{Average displacement error (ADE):}
  The ADE is calculated as the mean Euclidean distance between the predicted trajectory and the ground truth across all forecasted time steps:
  \[
    \text{ADE} = \frac{1}{T}\sum_{t=1}^T \| \hat{p}_t - p_t \|_2
  \]
  where $\hat{p}_t$ and $p_t$ denote the predicted and ground-truth positions at time step $t$, and $T$ denotes the number of steps. The metric ranges from $0$ to $125.1$\,m.

  \item \textbf{Final displacement error (FDE):}
  The FDE is calculated as the spatial deviation at the final time step $T$ of the prediction window to assess the model's ability to forecast trajectory endpoints:
  \[
    \text{FDE} = \| \hat{p}_{T} - p_{T} \|_2
  \]
  where $T$ denotes the final forecasted time step. The metric ranges from $0$ to $125.1$\,m.     
\end{itemize}

\subsubsection{Collective Structure Consistency}
\label{appendixsub:structure}
The following metrics are utilized to assess team organizational coherence and tactical coordination.
\begin{itemize}
    \setlength \itemsep{5pt}
    \item \textbf{Stretch index:}
     Given a team of $N = 11$ players, the team centroid $c$ can be obtained as $c = \frac{1}{N}\sum_{i=1}^N p_i$, where $p_i = (x_i, y_i)$ denotes the position of player $i$. The stretch index is calculated as the mean Euclidean distance from all players to the centroid:
    \[
      \text{Stretch index} = \frac{1}{N}\sum_{i=1}^N \| p_i - c \|_2
    \]
    which quantifies the spatial dispersion of the formation. The metric ranges from $0$ to $62.5$\,m.

    \item \textbf{Surface area:}
    The surface area captures the pitch region occupied by the team, calculated as the area of the convex hull enclosing all teammates. The metric ranges from $0$ to $7140$\,m$^2$.

    \item \textbf{Team width:}
    The team width characterizes the horizontal expansion of the team shape, calculated as the maximal transverse separation:
    \[
      \text{Team width} = \max_i(y_i) - \min_i(y_i)
    \]
    The metric ranges from $0$ to $68$\,m.

    \item \textbf{Team length:}
    The team length characterizes the vertical expansion of the team shape, calculated as the maximal longitudinal separation:
    \[
      \text{Team length} = \max_i(x_i) - \min_i(x_i)
    \]
    The metric ranges from $0$ to $105$\,m.

    \item \textbf{Frobenius norm:}
    The Frobenius norm provides a scalar measure of formation compactness, calculated as the square root of the sum of squared pairwise distances:
    \[
      \text{Frobenius norm} = \sqrt{\sum_{i=1}^N \sum_{j=1}^N \| p_i - p_j \|_2^2}
    \]

    \item \textbf{Team centroid displacement:}
    The team centroid displacement captures macro-level tactical shifts by evaluating the centroid's movement between consecutive frames $t-1$ and $t$:
    \[
      \text{Team centroid displacement} = \| c_t - c_{t-1} \|_2
    \]
    where practical values are limited by $\Delta t \times v_{\max}$ ($v_{\max} \approx 10$\,m/s).

    \item \textbf{Kuramoto order parameter:}
    The Kuramoto order parameter assesses directional synchronization in collective motion:
    \[
      R = \frac{1}{N} \left| \sum_{j=1}^N e^{i\theta_j} \right|
    \]
    where $\theta_j$ denotes the movement direction of player $j$ mapped onto the unit circle. It ranges from $0$ to $1$, where $1$ indicates perfectly coherent movement.
\end{itemize}
\begin{figure*}[htbp]
    \centering
    \begin{subfigure}{0.24\textwidth}
        \includegraphics[width=\linewidth]{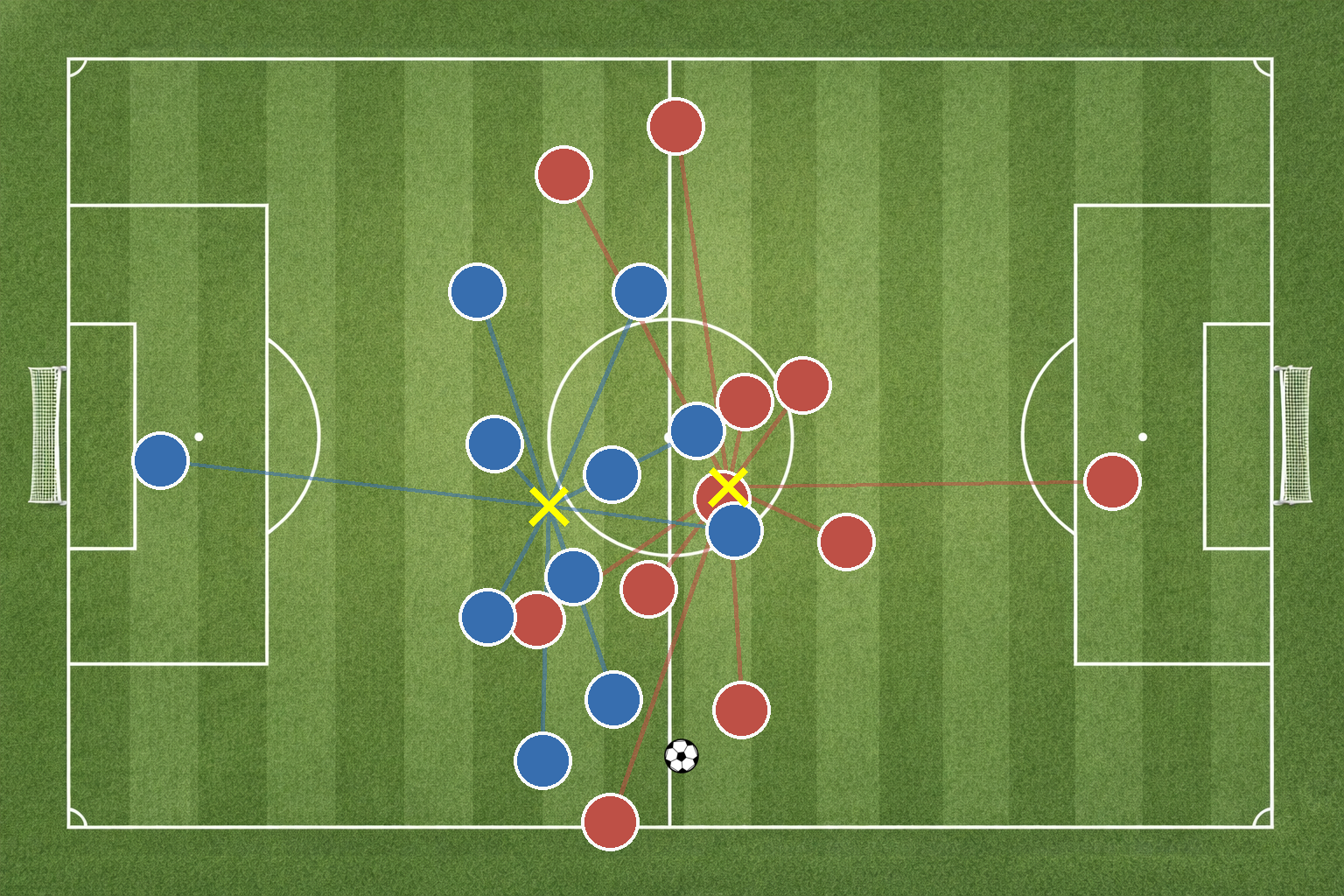}
        \caption{stretch index}
    \end{subfigure}\hfill
    \begin{subfigure}{0.24\textwidth}
        \includegraphics[width=\linewidth]{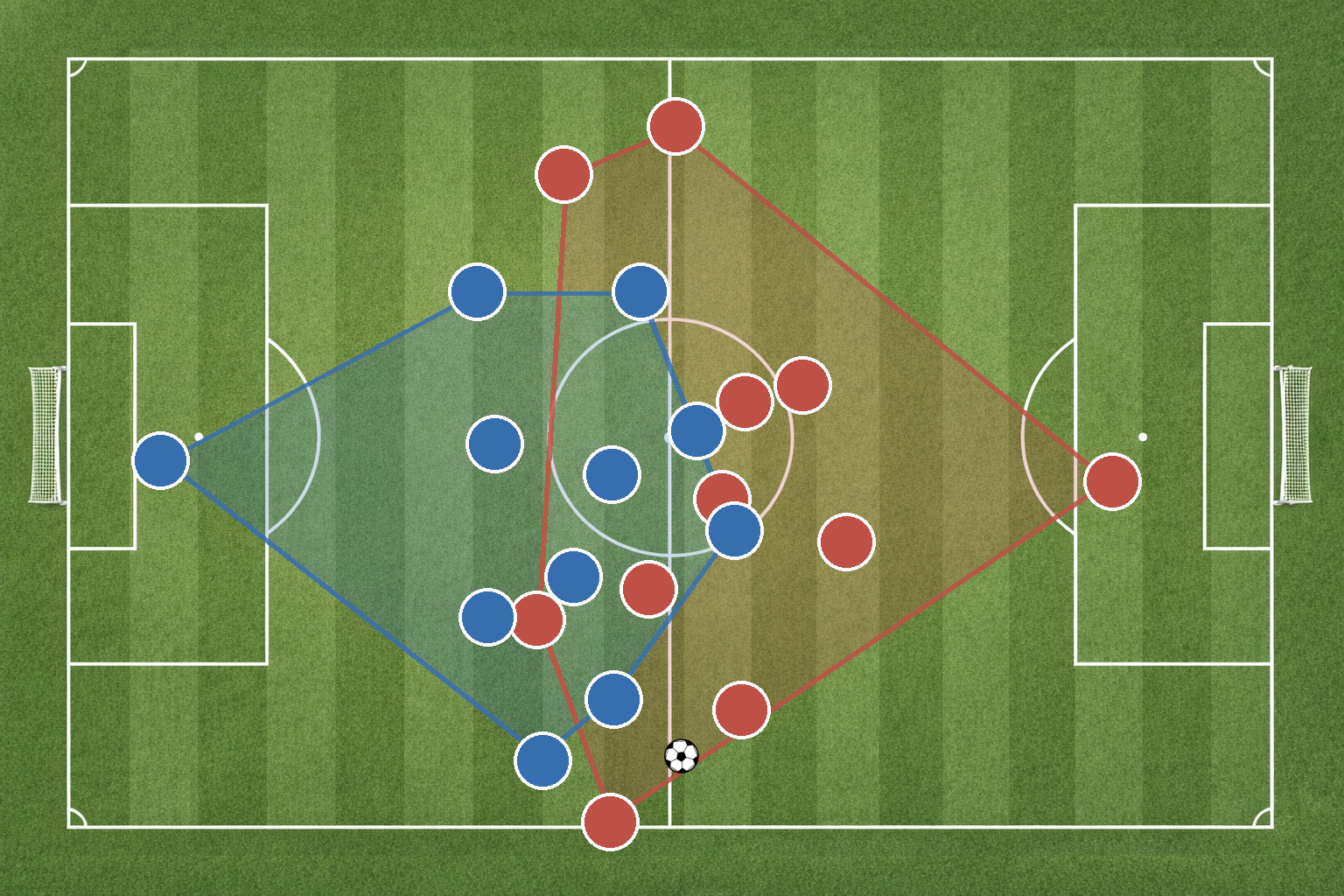}
        \caption{surface area}
    \end{subfigure}\hfill
    \begin{subfigure}{0.24\textwidth}
        \includegraphics[width=\linewidth]{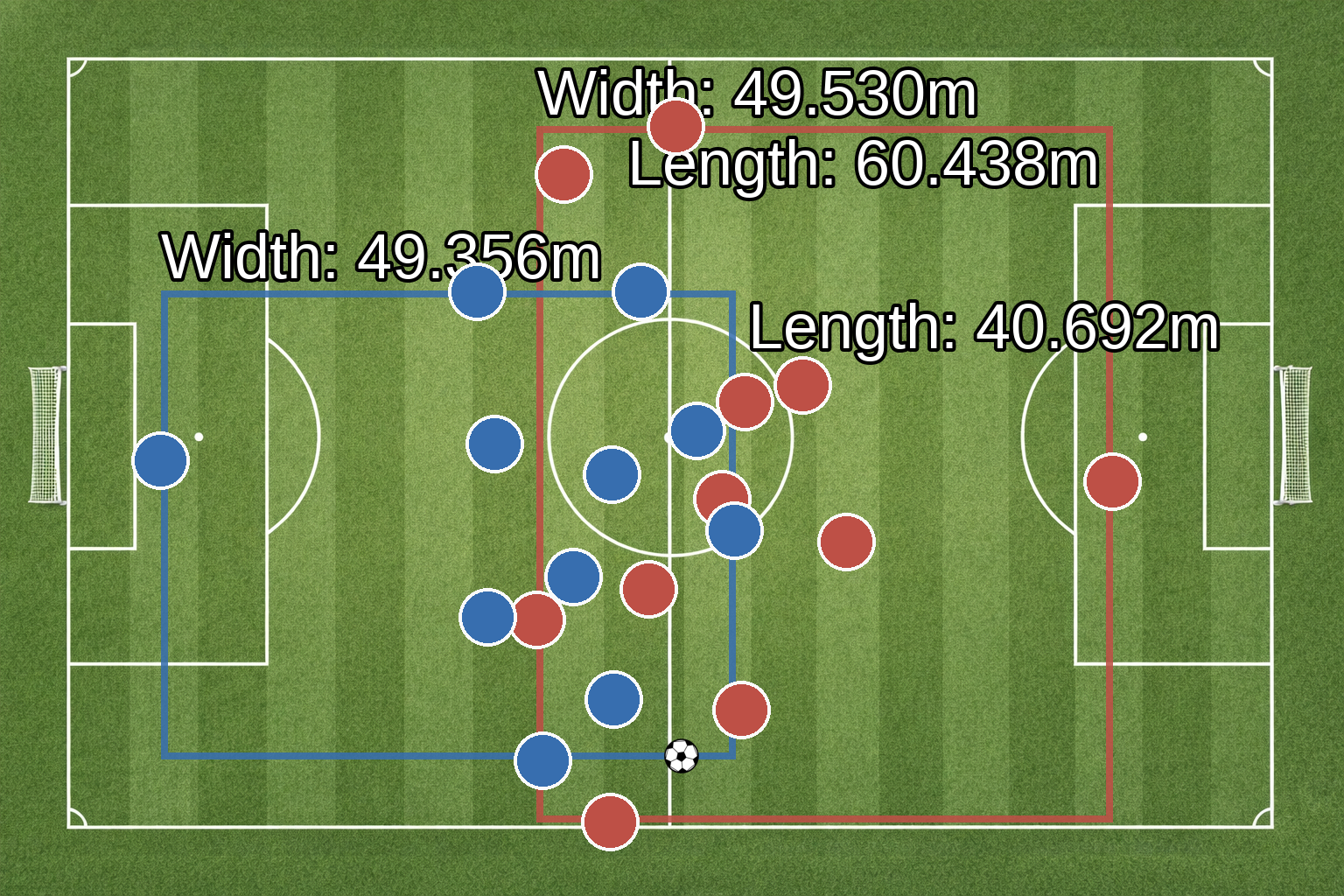}
        \caption{length \& width}
    \end{subfigure}\hfill
    \begin{subfigure}{0.24\textwidth}
        \includegraphics[width=\linewidth]{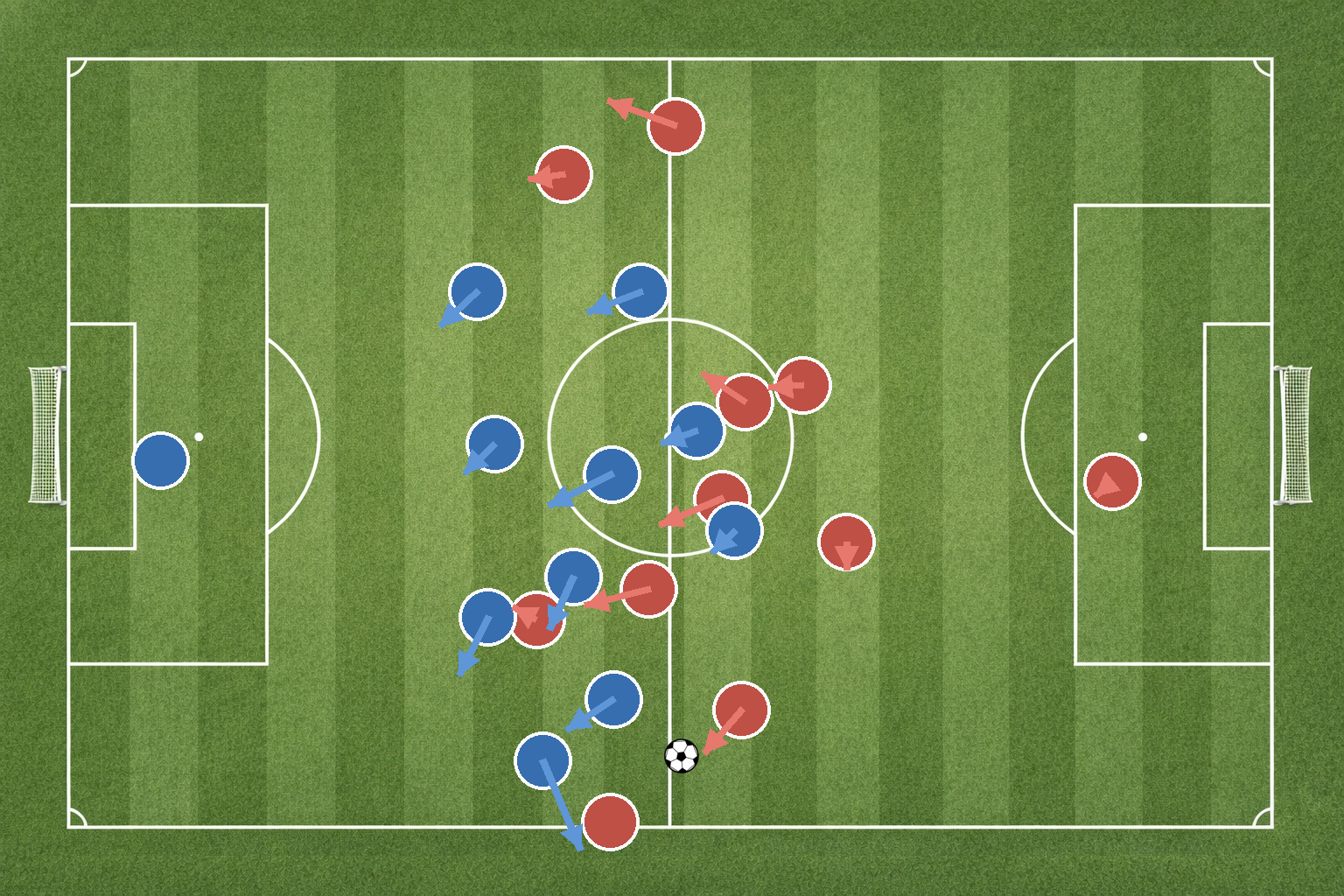}
        \caption{Kuramoto order}
    \end{subfigure}
    
    \caption{\textbf{Visualizations of Collective Structure Consistency.} Example from a progression phase (Red: attacking team, Blue: defending team). \textbf{(a) Stretch index} maps player distances to the team centroid (yellow cross), contrasting the defensive block's compactness with the attacking team's expanded shape. \textbf{(b) Surface area} uses convex hulls to illustrate spatial occupation, showing the attacking team utilizing greater pitch width and depth. \textbf{(c) Length \& width} quantifies these formations: the attacking team measures $49.53$\,m in width and $60.44$\,m in length, compared to the defending team's $49.36$\,m by $40.69$\,m. \textbf{(d) Kuramoto order} uses velocity arrows to visualize the Kuramoto order parameter, indicating the directional cohesion of both teams' movements.}
    \label{fig:metrics_structure}
\end{figure*}

\subsubsection{Semantic Event Recognition Accuracy}
\label{appendixsub:semantic}
We evaluate semantic event prediction using standard classification metrics, which are bounded between 0 and 1.

\begin{itemize}
    \setlength \itemsep{5pt}
    \item \textbf{Top-$k$ accuracy:} This metric measures the overall proportion of samples in which the ground-truth event label is present within the model's top-k predicted classes. It provides a general assessment of the model's predictive capacity.

    \item \textbf{Per-class recall@\(k\):} Because overall accuracy can be disproportionately skewed by majority classes in highly imbalanced sports datasets ({\em e.g.}, frequent passes versus rare goals), we utilize per-class recall to ensure a rigorous and fair evaluation of both common and rare tactical events. To comprehensively capture the hierarchical taxonomy of tactical actions, we conduct a fine-grained analysis at two levels of granularity: (i) At the broader {\em type} level ({\em e.g.}, Build, Threat), we report the individual recall for each specific category alongside their unweighted macro-average. These are evaluated for the top-1 and top-3 predictions (recall@1 and recall@3);
    (ii) At the more granular {\em subtype} level ({\em e.g.}, Goal, Shot-saved), precise action prediction becomes significantly more challenging. Consequently, we report the individual recall for each specific subtype and their macro-average across the top-1, top-3, and top-5 predictions (recall@1, 3, and 5).
\end{itemize}

\subsubsection{Offense/Defense Performance}
\label{appendixsub:performance}

To evaluate team-level tactical efficacy, we introduce a suite of advanced analytical metrics comprising three offensive indicators (off-ball expected threat, depth threat, and width threat) and two defensive indicators (defensive shape disruption and defensive dominant region).

Central to several of these metrics is the concept of expected possession value (EPV), which quantifies the latent offensive threat associated with specific spatial regions on the pitch. We utilize a pre-calculated 2D EPV grid sourced from the \textit{Friends-of-Tracking-Data} (FoTD) open-source repository~\cite{fotd_laurieontracking}. This spatial grid maps the historical probability of a possession ultimately resulting in a goal based on the current coordinates of the ball or player.

\begin{itemize}
    \setlength \itemsep{5pt}
    \item \textbf{Off-ball expected threat (OBET):}
    The OBET evaluates the attacking team's spatial advantage. We calculate it as the ratio of the EPV in areas controlled by the attacking team to the total pitch EPV:
    $$
    \text{OBET} = \frac{\sum_{k \in \mathcal{A}} \text{EPV}_k}{\sum_{k \in \mathcal{A} \cup \mathcal{D}} \text{EPV}_k}
    $$
    where $\mathcal{A}$ and $\mathcal{D}$ denote grid points controlled by the attacking and defending teams, respectively. The metric ranges from 0 to 1.

    \item \textbf{Depth threat:}
    The depth threat measures the offensive penetration along the vertical axis. The pitch is divided into $Z=32$ vertical zones, and the metric is calculated as the weighted sum of the attacking control ratio in each zone, weighted by the zone's EPV proportion:
    \[
      \text{Depth threat} = \sum_{z=1}^Z \left( \frac{N_{z}^{\text{atk}}}{N_z} \right) \left( \frac{\text{EPV}_z}{\text{EPV}_{\text{total}}} \right)
    \]
    where $N_z$ denotes the total number of grid points in zone $z$, and $N_{z}^{\text{atk}}$ denotes the points controlled by the attacking team. The metric ranges from $0$ to $1$.

    \item \textbf{Width threat:}
    The width threat measures horizontal offensive dispersion along the transverse axis, utilizing the identical weighted sum formulation as Depth Threat, but segmenting the pitch into horizontal zones. The metric ranges from $0$ to $1$.

    \item \textbf{Defensive shape disruption:}
    The defensive shape disruption quantifies the structural instability imposed on the opposing team, calculated as the normalized change in the defending team's surface area between consecutive tactical states $t-1$ and $t+1$:
    \[
      \text{Disruption} = \text{Clip}\left( \frac{\text{Area}_{t+1} - \text{Area}_{t-1}}{\text{Pitch Area}} \times 100, \,-1.0, \,1.0 \right)
    \]
    The metric captures forced expansion or compression of the defensive block. We scale the normalized change by a factor of $100$ to map typical frame-to-frame area variations to a more interpretable and symmetric boundary, yielding a final metric bounded between $-1$ and $1$.
    \item \textbf{Defensive dominant region:}
    The defensive dominant region \cite{taki2000visualization} evaluates the spatial dominance of the defending team. It represents the total pitch area where defensive players arrive earlier than attacking players, adjusting distance matrices using respective velocity vectors~\cite{taki2000visualization}. The metric ranges from $0$ to $7140$m$^2$.
\end{itemize}
\begin{figure*}[htbp]
    \centering
    \begin{subfigure}{0.24\textwidth}
        \includegraphics[width=\linewidth]{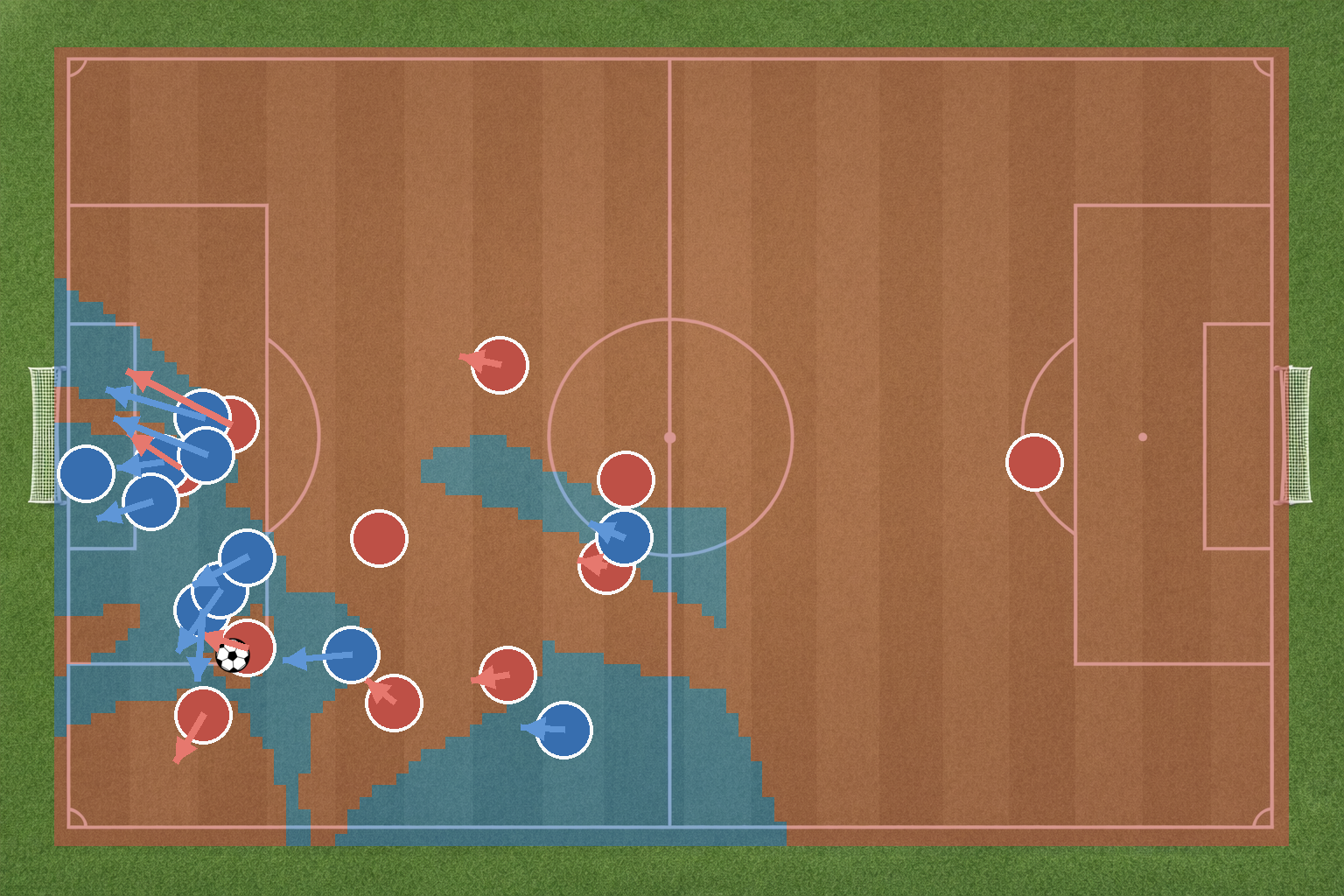}
        \caption{Off-ball expected threat}
    \end{subfigure}\hfill
    \begin{subfigure}{0.24\textwidth}
        \includegraphics[width=\linewidth]{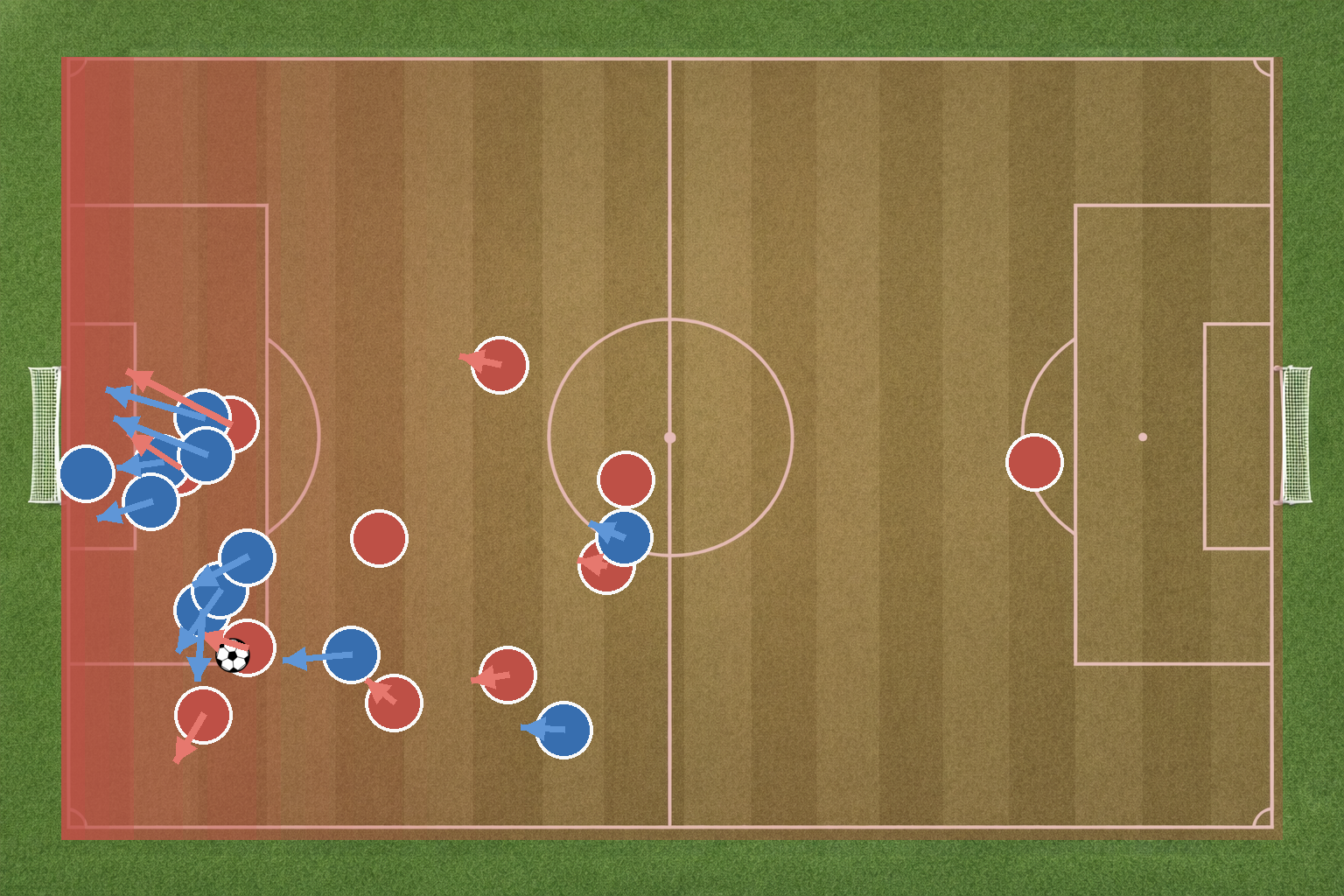}
        \caption{Depth threat}
    \end{subfigure}\hfill
    \begin{subfigure}{0.24\textwidth}
        \includegraphics[width=\linewidth]{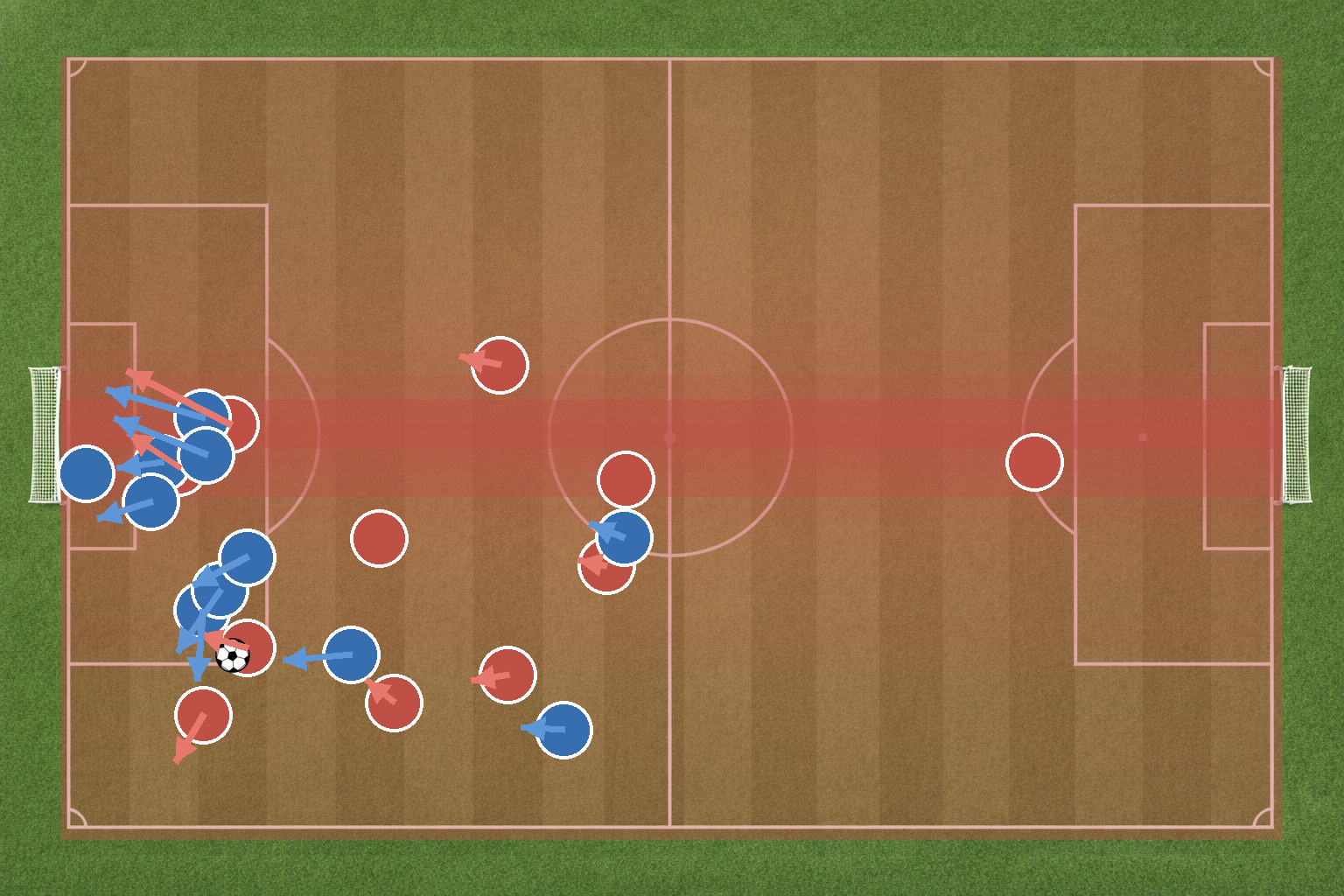}
        \caption{Width threat}
    \end{subfigure}\hfill
    \begin{subfigure}{0.24\textwidth}
        \includegraphics[width=\linewidth]{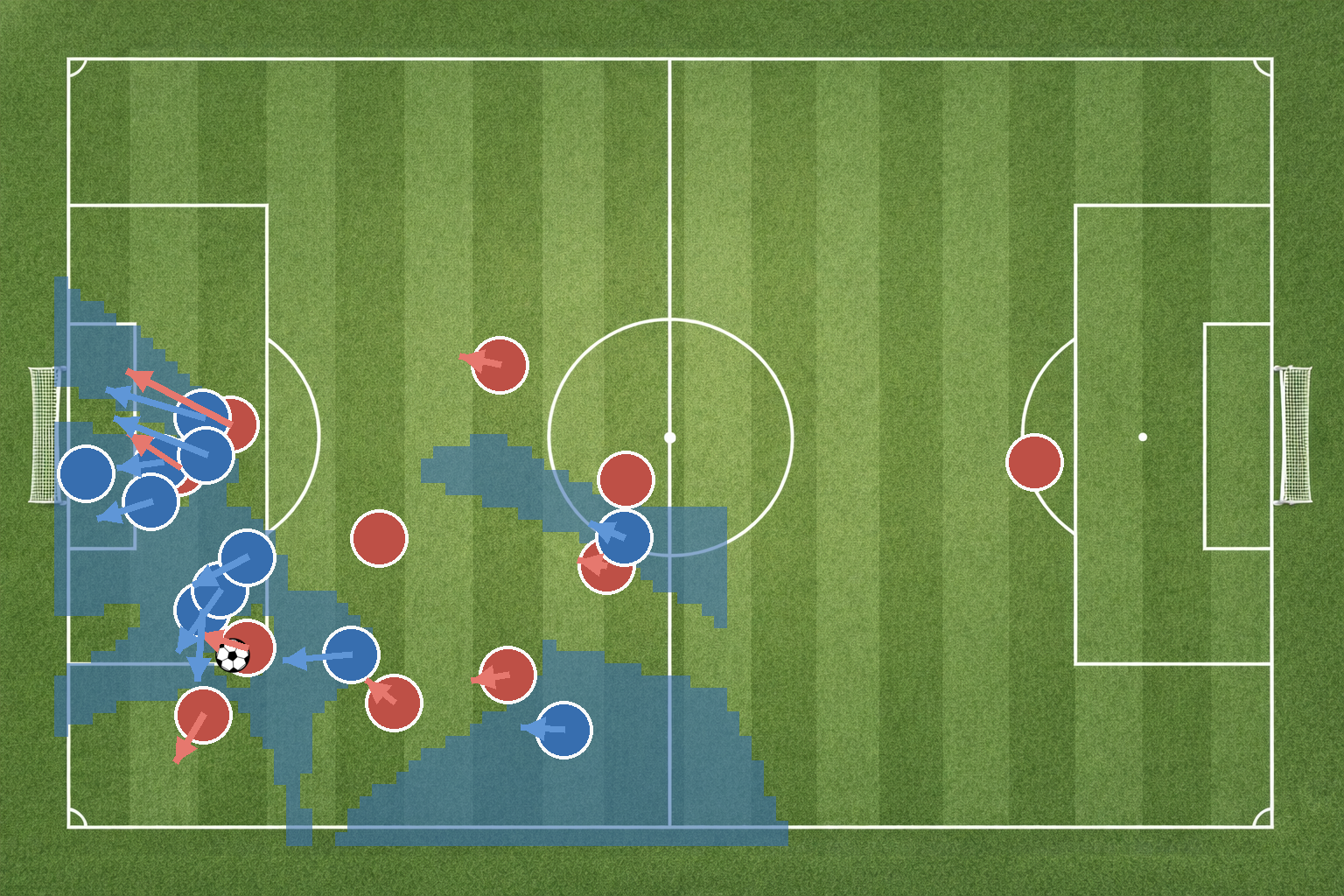}
        \caption{Defensive dominant region}
    \end{subfigure}
    
    \caption{\textbf{Visualizations of Offense and Defense Performance.} Example from a deep attacking phase (Red: attacking team, Blue: defending team). \textbf{(a) Off-ball expected threat} maps dynamic spatial control, showing the attacking team dominating part of the penalty arc while the defending team is highly compressed. \textbf{(b) Depth threat} evaluates vertical penetration by segmenting the pitch into vertical strips; the intense red bands near the goal line indicate successful offensive progression into the final third, highlighting where possession yields the highest threat coming. \textbf{(c) Width threat} assesses lateral distribution via horizontal strips; the prominent red band along the central axis demonstrates that despite the presence of wide players, the core offensive threat is concentrated in the middle, forcing the defense to prioritize central passing lanes. \textbf{(d) Defensive dominant region} isolates the defending team's spatial dominance, demonstrating their constrained control restricted primarily to the penalty box.}
    \label{fig:metrics_performance}
\end{figure*}

\clearpage

\subsection{Experiment Results}
This section provides detailed quantitative results supporting the analyses presented in the main text. 
We report experimental evaluations of trajectory forecasting under various conditioning settings, its generalization to other team sports, and the performance of tactical event grounding and forecasting.

\subsubsection{Performance of Unconditioned and Opponent-conditioned Trajectory Forecasting}

\paragraph{Geometrical Accuracy.} 
We report the geometric accuracy of GenTac's soccer trajectory forecasting under different causal window lengths and opponent-conditioning settings in Supplementary Tables~\ref{tab:soccer_history_window_condition} and \ref{tab:soccer_history_window_condition_avg}. Across all experiments in these tables, the historical observation horizon is fixed at 4\,s, while the prediction horizons range from 1\,s to 5\,s. 
To account for the stochastic nature of the generative model, metrics are computed over $K=20$ generated trajectory samples per instance, reporting both the minimum (best-of-$K$) and average error values.

As detailed in tables, for a given causal window size, incorporating the future trajectories of opposing players consistently improves the geometric accuracy of the forecasts across all prediction horizons (1\,s to 5\,s). This advantage is reflected in both the minimum and average error metrics. The improvement becomes particularly pronounced at longer horizons, where the performance gap between opponent-conditioned and unconditioned forecasting progressively widens. 

Regarding the effect of the causal window size, reducing the window from 1.0\,s to 0.2\,s yields a substantial improvement in geometric accuracy across all prediction horizons. This trend is consistent across both minimum and average errors, with the benefit becoming increasingly pronounced at longer horizons. This suggests that shorter local rollout steps effectively mitigate the compounding accumulation of forecasting errors over time. Conversely, further reducing the window from 0.2\,s to 0.04\,s provides only marginal additional gains. While this finer temporal granularity slightly enhances trajectory precision, the improvement is negligible compared to the notable gains achieved when transitioning from 1.0\,s to 0.2\,s. Furthermore, because covering an identical future horizon with a 0.04\,s window requires five times as many autoregressive inference steps as a 0.2\,s window, the computational inference cost increases substantially. This presents a clear trade-off between marginal accuracy gains and computational efficiency.

\begin{table}[h]
    \centering
    \footnotesize
    \setlength{\tabcolsep}{4pt}
    \begin{adjustbox}{width=\linewidth}
    \begin{tabular}{cc|cc|cc|cc|cc|cc}
    \toprule[1pt]
    \multicolumn{2}{c|}{\textbf{Setting}}
    & \multicolumn{2}{c|}{\textbf{1s}}
    & \multicolumn{2}{c|}{\textbf{2s}}
    & \multicolumn{2}{c|}{\textbf{3s}}
    & \multicolumn{2}{c|}{\textbf{4s}}
    & \multicolumn{2}{c}{\textbf{5s}} \\
    \cmidrule(lr){1-2}\cmidrule(lr){3-4}\cmidrule(lr){5-6}\cmidrule(lr){7-8}\cmidrule(lr){9-10}\cmidrule(lr){11-12}
    \textbf{Win.(s)} & \textbf{OC} 
    & minADE$_K$ & minFDE$_K$
    & minADE$_K$ & minFDE$_K$
    & minADE$_K$ & minFDE$_K$
    & minADE$_K$ & minFDE$_K$
    & minADE$_K$ & minFDE$_K$ \\
    \midrule
    0.04 & \ding{51} & 0.09 & 0.23 & 0.28 & 0.71 & 0.53 & 1.32 & 0.82 & 2.01 & 1.14 & 2.74 \\
    0.04 & \ding{55} & 0.23 & 0.55 & 0.67 & 1.69 & 1.27 & 3.16 & 1.97 & 4.85 & 2.74 & 6.68 \\
    0.2  & \ding{51} & 0.12 & 0.30 & 0.36 & 0.85 & 0.65 & 1.52 & 0.97 & 2.21 & 1.30 & 2.89 \\
    0.2  & \ding{55} & 0.62 & 1.22 & 1.34 & 2.87 & 2.24 & 5.11 & 3.32 & 7.82 & 4.55 & 10.80 \\
    1    & \ding{51} & 0.37 & 0.86 & 1.12 & 1.98 & 1.87 & 3.08 & 2.37 & 3.99 & 3.01 & 5.87 \\
    1    & \ding{55} & 0.95 & 1.78 & 1.84 & 3.44 & 3.59 & 6.70 & 5.25 & 9.62 & 6.03 & 10.93\\
    \bottomrule[1pt]
    \end{tabular}
    \end{adjustbox}
    \vspace{6pt}
    \caption{
    \textbf{Geometrical accuracy of GenTac trajectory forecasting (minimum errors) under varying window lengths and opponent conditioning.}
    minADE$_K$ and minFDE$_K$ denote the minimum ADE and FDE across $K=20$ generated trajectory samples per instance.
    }
    \label{tab:soccer_history_window_condition}
\end{table}

\begin{table}[h]
    \centering
    \footnotesize
    \setlength{\tabcolsep}{4pt}
    \begin{adjustbox}{width=\linewidth}
    \begin{tabular}{cc|cc|cc|cc|cc|cc}
    \toprule[1pt]
    \multicolumn{2}{c|}{\textbf{Setting}}
    & \multicolumn{2}{c|}{\textbf{1s}}
    & \multicolumn{2}{c|}{\textbf{2s}}
    & \multicolumn{2}{c|}{\textbf{3s}}
    & \multicolumn{2}{c|}{\textbf{4s}}
    & \multicolumn{2}{c}{\textbf{5s}} \\
    \cmidrule(lr){1-2}\cmidrule(lr){3-4}\cmidrule(lr){5-6}\cmidrule(lr){7-8}\cmidrule(lr){9-10}\cmidrule(lr){11-12}
    \textbf{Win.(s)} & \textbf{OC} 
    & avgADE$_K$ & avgFDE$_K$
    & avgADE$_K$ & avgFDE$_K$
    & avgADE$_K$ & avgFDE$_K$
    & avgADE$_K$ & avgFDE$_K$
    & avgADE$_K$ & avgFDE$_K$ \\
    \midrule
    0.04 & \ding{51} & 0.10 & 0.26 & 0.32 & 0.81 & 0.60 & 1.51 & 0.93 & 2.31 & 1.29 & 3.16 \\
    0.04 & \ding{55} & 0.26 & 0.62 & 0.75 & 1.87 & 1.39 & 3.49 & 2.15 & 5.34 & 3.00 & 7.38 \\
    0.2  & \ding{51} & 0.15 & 0.37 & 0.43 & 1.05 & 0.77 & 1.84 & 1.15 & 2.67 & 1.54 & 3.47 \\
    0.2  & \ding{55} & 0.70 & 1.39 & 1.55 & 3.45 & 2.64 & 6.20 & 3.94 & 9.41 & 5.39 & 12.81 \\
    1    & \ding{51} & 0.50 & 1.13 & 1.46 & 2.33 & 2.05 & 3.65 & 2.64 & 4.90 & 4.05 & 7.37 \\
    1    & \ding{55} & 1.28 & 2.45 & 2.45 & 4.66 & 4.68 & 8.97 & 6.80 & 12.90 & 7.81 & 14.67 \\
    \bottomrule[1pt]
    \end{tabular}
    \end{adjustbox}
    \vspace{6pt}
    \caption{
    \textbf{Geometrical accuracy of GenTac trajectory forecasting (average errors) under varying window lengths and opponent conditioning.}
    avgADE$_K$ and avgFDE$_K$ denote the average ADE and FDE across $K=20$ generated trajectory samples per instance.
    }
    \label{tab:soccer_history_window_condition_avg}
\end{table}

\newpage

\paragraph{Collective Structural Consistency.} 
We further evaluate the collective structural consistency of predicted trajectories by measuring the difference of high-level tactical metrics between forecasted trajectories and ground truth. 
For each prediction horizon (1--5\,s), we compute the absolute value of errors among metrics, reported as $\Delta(|\cdot|)$. 
Errors are summarized using two aggregation schemes out of $K$(=20) generated trajectory samples : (i) Average error, which averages errors compared with ground truth over all generated trajectory samples, and (ii) Minimum error, which selects the minimum errors. 

Supplementary Table~\ref{tab:tactical_metrics_uncond} and Supplementary Table~\ref{tab:tactical_metrics_oc} report the results for unconditioned and opponent-conditioned trajectory forecasting, respectively.
The evaluated metrics characterize the absolute value of the changes in collective team structure metrics on forecasted trajectories compared with ground truth, including stretch index ($\Delta$|SI|), surface area ($\Delta$|SA|), team width ($\Delta$|TW|), team length ($\Delta$|TL|), Frobenius norm of pairwise distances ($\Delta$|FN|), team centroid displacement ($\Delta$|CD|), and synchronization order parameter ($\Delta$|SO|). For these collective structural metrics, their abbreviated forms will be consistently used in all subsequent tables. 
All the results below are demonstrated under 0.2 second causal window and as shown in Supplementary Table~\ref{tab:tactical_metrics_uncond} and Supplementary Table~\ref{tab:tactical_metrics_oc}, similar trend could be found in collective structural consistency as in geometrical accuracy.


\begin{table}[h]
    \centering
    \scriptsize
    \setlength{\tabcolsep}{4pt}
    \renewcommand{\arraystretch}{1.05}
    
    \begin{adjustbox}{width=\linewidth}
    \begin{tabular}{c|ccccccc}
    \toprule[1pt]
    
    \multicolumn{1}{c|}{\textbf{Metrics}}
    & $\Delta$|SI|$\downarrow$ (m)
& $\Delta$|SA|$\downarrow$ (m$^2$)
& $\Delta$|TW|$\downarrow$ (m)
& $\Delta$|TL|$\downarrow$ (m)
& $\Delta$|FN|$\downarrow$ (m)
& $\Delta$|CD|$\downarrow$ (m)
& $\Delta$|SO|$\downarrow$ (-) \\
    
    \cmidrule(lr){1-1}\cmidrule(lr){2-8}
    
    \multicolumn{1}{c|}{\textbf{Time}}
    & \multicolumn{7}{c}{\textbf{Average Error}} \\
    \midrule
    
    \textbf{1s} & 0.32 & 40.67 & 1.07 & 1.09 & 0.44 & 0.80 & 0.28 \\
    \textbf{2s} & 0.64 & 82.81 & 2.15 & 2.18 & 0.89 & 1.41 & 0.35 \\
    \textbf{3s} & 1.00 & 130.33 & 3.32 & 3.38 & 1.38 & 1.79 & 0.37 \\
    \textbf{4s} & 1.42 & 186.64 & 4.56 & 4.88 & 1.97 & 1.94 & 0.37 \\
    \textbf{5s} & 1.89 & 255.55 & 5.87 & 6.94 & 2.66 & 1.96 & 0.37 \\
    
    \midrule
    
    \multicolumn{1}{c|}{\textbf{Time}}
    & \multicolumn{7}{c}{\textbf{Minimum Error}} \\
    \midrule
    
    \textbf{1s} & 0.19 & 21.24 & 0.60 & 0.52 & 0.26 & 0.29 & 0.07 \\
    \textbf{2s} & 0.40 & 46.81 & 1.32 & 1.15 & 0.55 & 0.71 & 0.13 \\
    \textbf{3s} & 0.62 & 74.92 & 2.08 & 1.81 & 0.86 & 1.01 & 0.15 \\
    \textbf{4s} & 0.87 & 105.61 & 2.90 & 2.53 & 1.22 & 1.13 & 0.16 \\
    \textbf{5s} & 1.17 & 142.05 & 3.75 & 3.52 & 1.65 & 1.14 & 0.16 \\
    
    \bottomrule[1pt]
    \end{tabular}
    \end{adjustbox}
    
    \vspace{6pt}
    \caption{
    \textbf{Errors of collective structural consistency induced by trajectories from unconditioned forecasting.}
    }
    \label{tab:tactical_metrics_uncond}
\end{table}


\begin{table}[h]
    \centering
    \scriptsize
    \setlength{\tabcolsep}{4pt}
    \renewcommand{\arraystretch}{1.05}
    
    \begin{adjustbox}{width=\linewidth}
    \begin{tabular}{c|ccccccc}
    \toprule[1pt]
    
    \multicolumn{1}{c|}{\textbf{Metrics}}
    & $\Delta$|SI|$\downarrow$ (m)
& $\Delta$|SA|$\downarrow$ (m$^2$)
& $\Delta$|TW|$\downarrow$ (m)
& $\Delta$|TL|$\downarrow$ (m)
& $\Delta$|FN|$\downarrow$ (m)
& $\Delta$|CD|$\downarrow$ (m)
& $\Delta$|SO|$\downarrow$ (-) \\
    
    \cmidrule(lr){1-1}\cmidrule(lr){2-8}
    
    \multicolumn{1}{c|}{\textbf{Time}}
    & \multicolumn{7}{c}{\textbf{Average Error}} \\
    \midrule
    
    \textbf{1s} & 0.12 & 16.82 & 0.43 & 0.48 & 0.17 & 0.30 & 0.16 \\
    \textbf{2s} & 0.35 & 48.41 & 1.20 & 1.37 & 0.49 & 0.47 & 0.22 \\
    \textbf{3s} & 0.64 & 86.90 & 2.18 & 2.39 & 0.89 & 0.58 & 0.26 \\
    \textbf{4s} & 0.95 & 126.40 & 3.21 & 3.49 & 1.32 & 0.66 & 0.28 \\
    \textbf{5s} & 1.25 & 164.75 & 4.21 & 4.53 & 1.74 & 0.71 & 0.29 \\
    
    \midrule
    
    \multicolumn{1}{c|}{\textbf{Time}}
    & \multicolumn{7}{c}{\textbf{Minimum Error}} \\
    \midrule
    
    \textbf{1s} & 0.08 & 11.25 & 0.26 & 0.33 & 0.12 & 0.18 & 0.08 \\
    \textbf{2s} & 0.25 & 33.37 & 0.75 & 0.95 & 0.35 & 0.28 & 0.11 \\
    \textbf{3s} & 0.47 & 60.29 & 1.38 & 1.67 & 0.65 & 0.34 & 0.12 \\
    \textbf{4s} & 0.70 & 87.55 & 2.03 & 2.43 & 0.96 & 0.37 & 0.13 \\
    \textbf{5s} & 0.91 & 112.64 & 2.61 & 3.14 & 1.27 & 0.40 & 0.13 \\
    
    \bottomrule[1pt]
    \end{tabular}
    \end{adjustbox}
    
    \vspace{6pt}
    \caption{
    \textbf{Errors of collective structural consistency induced by trajectories from opponent-conditioned forecasting.}
    }
    \label{tab:tactical_metrics_oc}
\end{table}

\clearpage
\subsubsection{Performance of Team-conditioned Trajectory Forecasting}
We further evaluate the impact of team-conditioned trajectory forecasting by measuring the induced errors in collective tactical metrics for Auckland FC. 
For each prediction horizon (1--5\,s), we compare the original (opponent-conditioned) forecasting model with the team-conditioned setting (fine-tuned on matches of Auckland FC). 
Errors are summarized using average and minimum errors among K$(=20)$ generated trajectory samples as well.
Results are reported in Supplementary Table~\ref{tab:tactical_mean} and Supplementary Table~\ref{tab:tactical_min}.

\begin{table}[htbp]
\centering
\footnotesize
\setlength{\tabcolsep}{3pt}
\renewcommand{\arraystretch}{1.05}

\begin{adjustbox}{width=\linewidth}
\begin{tabular}{c|ccccccccc}
\toprule[1pt]

\multicolumn{1}{c|}{\textbf{Metrics}}
& ADE$\downarrow$ (m)
& FDE$\downarrow$ (m)
& $\Delta$|SI|$\downarrow$ (m)
& $\Delta$|SA|$\downarrow$ (m$^2$)
& $\Delta$|TW|$\downarrow$ (m)
& $\Delta$|TL|$\downarrow$ (m)
& $\Delta$|FN|$\downarrow$ (m)
& $\Delta$|CD|$\downarrow$ (m)
& $\Delta$|SO|$\downarrow$ (-) \\

\cmidrule(lr){1-1}\cmidrule(lr){2-10}

\multicolumn{1}{c|}{\textbf{Time}}
& \multicolumn{9}{c}{\textbf{Original (Average Error)}} \\
\midrule

\textbf{1s} & 0.15 & 0.38 & 0.26 & 34.73 & 0.83 & 0.73 & 0.36 & 0.48 & 0.22 \\
\textbf{2s} & 0.45 & 1.11 & 0.81 & 110.38 & 2.42 & 2.31 & 1.14 & 0.71 & 0.29 \\
\textbf{3s} & 0.82 & 1.98 & 1.63 & 231.16 & 4.65 & 4.83 & 2.32 & 0.85 & 0.32 \\
\textbf{4s} & 1.23 & 2.89 & 2.75 & 409.88 & 7.19 & 8.48 & 3.95 & 0.96 & 0.35 \\
\textbf{5s} & 1.65 & 3.76 & 4.16 & 652.25 & 9.97 & 13.34 & 6.03 & 1.08 & 0.38 \\

\midrule\midrule

\multicolumn{1}{c|}{\textbf{Time}}
& \multicolumn{9}{c}{\textbf{Conditioned on Team (Average Error)}} \\
\midrule
\textbf{1s} & 0.18 & 0.43 & 0.19 & 25.70 & 0.93 & 0.71 & 0.28 & 0.59 & 0.27 \\
\textbf{2s} & 0.52 & 1.31 & 0.57 & 73.44 & 2.32 & 2.12 & 0.82 & 0.86 & 0.37 \\
\textbf{3s} & 0.97 & 2.42 & 1.11 & 141.44 & 4.09 & 3.91 & 1.59 & 0.95 & 0.42 \\
\textbf{4s} & 1.49 & 3.62 & 1.82 & 228.39 & 6.12 & 5.89 & 2.57 & 0.96 & 0.44 \\
\textbf{5s} & 2.04 & 4.86 & 2.81 & 335.55 & 8.26 & 8.29 & 3.92 & 0.97 & 0.46 \\

\bottomrule[1pt]
\end{tabular}
\end{adjustbox}

\vspace{6pt}
\caption{
\textbf{Average errors of trajectory forecasting metrics for Auckland FC with and without team conditioning.}
}
\label{tab:tactical_mean}
\end{table}

\begin{table}[htbp]
\centering
\footnotesize
\setlength{\tabcolsep}{3pt}
\renewcommand{\arraystretch}{1.05}

\begin{adjustbox}{width=\linewidth}
\begin{tabular}{c|ccccccccc}
\toprule[1pt]

\multicolumn{1}{c|}{\textbf{Metrics}}
& ADE$\downarrow$ (m)
& FDE$\downarrow$ (m)
& $\Delta$|SI|$\downarrow$ (m)
& $\Delta$|SA|$\downarrow$ (m$^2$)
& $\Delta$|TW|$\downarrow$ (m)
& $\Delta$|TL|$\downarrow$ (m)
& $\Delta$|FN|$\downarrow$ (m)
& $\Delta$|CD|$\downarrow$ (m)
& $\Delta$|SO|$\downarrow$ (-) \\

\cmidrule(lr){1-1}\cmidrule(lr){2-10}

\multicolumn{1}{c|}{\textbf{Time}}
& \multicolumn{9}{c}{\textbf{Original (Minimum Error)}} \\
\midrule

\textbf{1s} & 0.12 & 0.31 & 0.20 & 24.60 & 0.55 & 0.47 & 0.28 & 0.29 & 0.12 \\
\textbf{2s} & 0.37 & 0.91 & 0.66 & 83.76 & 1.75 & 1.61 & 0.93 & 0.50 & 0.17 \\
\textbf{3s} & 0.69 & 1.63 & 1.37 & 183.53 & 3.51 & 3.57 & 1.95 & 0.61 & 0.21 \\
\textbf{4s} & 1.03 & 2.39 & 2.35 & 333.71 & 5.60 & 6.51 & 3.39 & 0.67 & 0.23 \\
\textbf{5s} & 1.40 & 3.12 & 3.61 & 539.36 & 7.92 & 10.51 & 5.26 & 0.71 & 0.25 \\

\midrule\midrule

\multicolumn{1}{c|}{\textbf{Time}}
& \multicolumn{9}{c}{\textbf{Conditioned on Team (Minimum Error)}} \\
\midrule

\textbf{1s} & 0.15 & 0.38 & 0.10 & 12.34 & 0.50 & 0.30 & 0.15 & 0.29 & 0.12 \\
\textbf{2s} & 0.46 & 1.18 & 0.33 & 39.32 & 1.36 & 1.05 & 0.49 & 0.50 & 0.22 \\
\textbf{3s} & 0.88 & 2.20 & 0.67 & 80.08 & 2.60 & 2.00 & 1.00 & 0.57 & 0.27 \\
\textbf{4s} & 1.36 & 3.31 & 1.15 & 131.27 & 4.10 & 3.05 & 1.67 & 0.57 & 0.28 \\
\textbf{5s} & 1.87 & 4.46 & 1.86 & 193.52 & 5.52 & 4.44 & 2.65 & 0.54 & 0.31 \\

\bottomrule[1pt]
\end{tabular}
\end{adjustbox}

\vspace{6pt}
\caption{
\textbf{Minimum errors of trajectory forecasting metrics for Auckland FC with and without team conditioning.}
}
\label{tab:tactical_min}
\label{tab:tactical_best_k}
\end{table}

As shown in Supplementary Table~\ref{tab:tactical_mean} and Supplementary Table~\ref{tab:tactical_min}., 
all results here are obtained with a fixed causal window of 0.2\,s. Compared with the original opponent-conditioned model, team conditioning slightly increases ADE and FDE across all horizons, suggesting that enforcing team-specific style does not necessarily yield the geometrically closest future. In contrast, it consistently reduces errors in collective tactical metrics, especially at longer horizons. For example, at 5\,s, the average errors of surface area, team width, and team length decrease from 652.25 to 335.55, from 9.97 to 8.26, and from 13.34 to 8.29, respectively. This indicates that team conditioning helps generate futures more consistent with the characteristic collective organization of Auckland FC.

\clearpage
\subsubsection{Performance of League-conditioned Trajectory Forecasting}

To evaluate the performance of league-conditioned trajectory forecasting, we simulate league-specific playing styles using two representative football competitions: the Australian Football League (AFL) and the German Bundesliga. 
For each league, we compare the original (unconditioned) trajectory forecasting setting with the league-conditioned setting (finetuned on matches from specific league) and measure the resulting errors in trajectory-induced tactical metrics across prediction horizons from 1\,s to 5\,s, where errors are summarized using average and minimun errors K$(=20)$ generated trajectory samples.

Results for AFL conditioning are reported in Supplementary Table~\ref{tab:afl_tactical_avg} and Supplementary Table~\ref{tab:afl_tactical_min},
while for Bundesliga conditioning are presented in Supplementary Table~\ref{tab:bundesliga_tactical_avg} and Supplementary Table~\ref{tab:bundesliga_tactical_min}.

\begin{table}[h]
\centering
\footnotesize
\setlength{\tabcolsep}{3pt}
\renewcommand{\arraystretch}{1.05}

\begin{adjustbox}{width=\linewidth}
\begin{tabular}{c|ccccccccc}
\toprule[1pt]

\multicolumn{1}{c|}{\textbf{Metrics}}
& ADE$\downarrow$ (m)
& FDE$\downarrow$ (m)
& $\Delta$|SI|$\downarrow$ (m)
& $\Delta$|SA|$\downarrow$ (m$^2$)
& $\Delta$|TW|$\downarrow$ (m)
& $\Delta$|TL|$\downarrow$ (m)
& $\Delta$|FN|$\downarrow$ (m)
& $\Delta$|CD|$\downarrow$ (m)
& $\Delta$|SO|$\downarrow$ (-) \\

\cmidrule(lr){1-1}\cmidrule(lr){2-10}

\multicolumn{1}{c|}{\textbf{Time}}
& \multicolumn{9}{c}{\textbf{Original (Average Error)}} \\
\midrule

\textbf{1s} & 0.71 & 1.44 & 0.37 & 46.65 & 1.06 & 1.30 & 0.50 & 0.89 & 0.25 \\
\textbf{2s} & 1.61 & 3.61 & 0.73 & 91.07 & 2.15 & 2.48 & 1.01 & 1.42 & 0.32 \\
\textbf{3s} & 2.76 & 6.46 & 1.10 & 134.33 & 3.41 & 3.61 & 1.51 & 1.66 & 0.34 \\
\textbf{4s} & 4.10 & 9.69 & 1.52 & 183.07 & 4.80 & 4.99 & 2.08 & 1.74 & 0.34 \\
\textbf{5s} & 5.57 & 13.09 & 2.00 & 242.65 & 6.22 & 7.06 & 2.76 & 1.77 & 0.35 \\

\midrule\midrule

\multicolumn{1}{c|}{\textbf{Time}}
& \multicolumn{9}{c}{\textbf{Conditioned on AFL (Average Error)}} \\
\midrule

\textbf{1s} & 0.44 & 1.11 & 0.21 & 29.21 & 1.10 & 1.16 & 0.31 & 0.68 & 0.23 \\
\textbf{2s} & 1.37 & 3.57 & 0.70 & 82.81 & 3.74 & 3.10 & 0.97 & 1.25 & 0.31 \\
\textbf{3s} & 2.67 & 6.95 & 1.38 & 145.42 & 7.60 & 4.96 & 1.91 & 1.74 & 0.35 \\
\textbf{4s} & 4.24 & 10.88 & 2.29 & 225.61 & 12.08 & 6.34 & 3.21 & 2.16 & 0.37 \\
\textbf{5s} & 6.00 & 15.08 & 3.46 & 334.85 & 16.56 & 7.18 & 4.92 & 2.50 & 0.38 \\

\bottomrule[1pt]
\end{tabular}
\end{adjustbox}

\vspace{6pt}
\caption{
\textbf{Average errors of trajectory forecasting metrics for AFL with and without league conditioning.}
}
\label{tab:afl_tactical_avg}
\end{table}

\begin{table}[h]
\centering
\footnotesize
\setlength{\tabcolsep}{3pt}
\renewcommand{\arraystretch}{1.05}

\begin{adjustbox}{width=\linewidth}
\begin{tabular}{c|ccccccccc}

\toprule[1pt]

\multicolumn{1}{c|}{\textbf{Metrics}}
& ADE$\downarrow$ (m)
& FDE$\downarrow$ (m)
& $\Delta$|SI|$\downarrow$ (m)
& $\Delta$|SA|$\downarrow$ (m$^2$)
& $\Delta$|TW|$\downarrow$ (m)
& $\Delta$|TL|$\downarrow$ (m)
& $\Delta$|FN|$\downarrow$ (m)
& $\Delta$|CD|$\downarrow$ (m)
& $\Delta$|SO|$\downarrow$ (-) \\

\cmidrule(lr){1-1}\cmidrule(lr){2-10}

\multicolumn{1}{c|}{\textbf{Time}}
& \multicolumn{9}{c}{\textbf{Original (Minimum Error)}} \\
\midrule

\textbf{1s} & 0.63 & 1.25 & 0.26 & 31.18 & 0.68 & 0.82 & 0.36 & 0.45 & 0.10 \\
\textbf{2s} & 1.39 & 3.03 & 0.55 & 66.26 & 1.49 & 1.72 & 0.76 & 0.83 & 0.16 \\
\textbf{3s} & 2.36 & 5.42 & 0.84 & 99.30 & 2.43 & 2.55 & 1.16 & 1.04 & 0.18 \\
\textbf{4s} & 3.50 & 8.21 & 1.16 & 133.57 & 3.51 & 3.44 & 1.59 & 1.10 & 0.19 \\
\textbf{5s} & 4.78 & 11.21 & 1.53 & 173.06 & 4.59 & 4.71 & 2.08 & 1.12 & 0.20 \\

\midrule\midrule

\multicolumn{1}{c|}{\textbf{Time}}
& \multicolumn{9}{c}{\textbf{Conditioned on AFL (Minimum Error)}} \\
\midrule

\textbf{1s} & 0.37 & 0.94 & 0.14 & 18.20 & 0.79 & 0.78 & 0.20 & 0.31 & 0.11 \\
\textbf{2s} & 1.17 & 3.02 & 0.49 & 53.76 & 2.81 & 2.23 & 0.69 & 0.57 & 0.14 \\
\textbf{3s} & 2.28 & 5.83 & 0.97 & 92.64 & 5.77 & 3.59 & 1.36 & 0.79 & 0.16 \\
\textbf{4s} & 3.60 & 8.98 & 1.63 & 143.05 & 9.21 & 4.42 & 2.31 & 0.95 & 0.16 \\
\textbf{5s} & 5.05 & 12.26 & 2.50 & 212.07 & 12.70 & 4.75 & 3.61 & 1.03 & 0.17 \\

\bottomrule[1pt]
\end{tabular}
\end{adjustbox}

\vspace{6pt}
\caption{
\textbf{Minimum errors of trajectory forecasting metrics for AFL with and without league conditioning.}
}
\label{tab:afl_tactical_min}
\end{table}

\newpage
League-conditioned forecasting shows a consistent short-horizon advantage for both AFL and Bundesliga. 
In AFL, league conditioning improves the average ADE/FDE at 1\,s from 0.71/1.44~m to 0.44/1.11~m, while in Bundesliga the corresponding values decrease from 0.71/1.40~m to 0.40/0.99~m. Similar trends are observed in the minimum errors. 
In addition to geometrical accuracy, several collective tactical metrics, including stretch index, surface area, Frobenius norm, and centroid displacement, are also reduced at short horizons after league conditioning. 
However, this advantage is not maintained uniformly as the prediction horizon extends. 
In particular, for AFL, structural errors such as team width, Frobenius norm, and surface area become noticeably larger at 4\,s and 5\,s, while Bundesliga remains relatively more stable but still shows a weaker gain than at short horizons. 
Overall, these results suggest that league conditioning mainly acts as a short-term tactical prior, helping the model better reproduce league-level playing tendencies in the early stages of trajectory rollout.

\vspace{6pt}

\begin{table}[htbp]
\centering
\footnotesize
\setlength{\tabcolsep}{3pt}
\renewcommand{\arraystretch}{1.05}

\begin{adjustbox}{width=\linewidth}
\begin{tabular}{c|ccccccccc}
\toprule[1pt]

\multicolumn{1}{c|}{\textbf{Metrics}}
& ADE$\downarrow$ (m)
& FDE$\downarrow$ (m)
& $\Delta$|SI|$\downarrow$ (m)
& $\Delta$|SA|$\downarrow$ (m$^2$)
& $\Delta$|TW|$\downarrow$ (m)
& $\Delta$|TL|$\downarrow$ (m)
& $\Delta$|FN|$\downarrow$ (m)
& $\Delta$|CD|$\downarrow$ (m)
& $\Delta$|SO|$\downarrow$ (-) \\

\cmidrule(lr){1-1}\cmidrule(lr){2-10}

\multicolumn{1}{c|}{\textbf{Time}}
& \multicolumn{9}{c}{\textbf{Original (Average Error)}} \\
\midrule

\textbf{1s} & 0.71 & 1.40 & 0.33 & 42.93 & 1.05 & 1.11 & 0.45 & 0.77 & 0.29 \\
\textbf{2s} & 1.54 & 3.39 & 0.65 & 85.39 & 2.03 & 2.19 & 0.88 & 1.42 & 0.36 \\
\textbf{3s} & 2.61 & 6.06 & 0.98 & 131.61 & 3.06 & 3.37 & 1.35 & 1.88 & 0.39 \\
\textbf{4s} & 3.87 & 9.18 & 1.33 & 181.99 & 4.15 & 4.63 & 1.86 & 2.10 & 0.39 \\
\textbf{5s} & 5.28 & 12.51 & 1.70 & 234.92 & 5.30 & 5.96 & 2.40 & 2.16 & 0.40 \\

\midrule\midrule

\multicolumn{1}{c|}{\textbf{Time}}
& \multicolumn{9}{c}{\textbf{Conditioned on Bundesliga (Average Error)}} \\
\midrule

\textbf{1s} & 0.40 & 0.99 & 0.19 & 26.67 & 0.82 & 0.76 & 0.27 & 0.72 & 0.32 \\
\textbf{2s} & 1.22 & 3.17 & 0.56 & 74.35 & 2.42 & 1.86 & 0.80 & 1.11 & 0.39 \\
\textbf{3s} & 2.36 & 6.18 & 1.07 & 133.94 & 4.56 & 3.00 & 1.50 & 1.50 & 0.41 \\
\textbf{4s} & 3.77 & 9.76 & 1.67 & 198.30 & 7.05 & 4.07 & 2.33 & 1.86 & 0.42 \\
\textbf{5s} & 5.37 & 13.67 & 2.34 & 263.32 & 9.74 & 5.03 & 3.27 & 2.08 & 0.42 \\

\bottomrule[1pt]
\end{tabular}
\end{adjustbox}

\vspace{6pt}
\caption{
\textbf{Average errors of trajectory forecasting tactical metrics for the Bundesliga with and without league conditioning.}
}
\label{tab:bundesliga_tactical_avg}
\end{table}

\begin{table}[htbp]
\centering
\footnotesize
\setlength{\tabcolsep}{3pt}
\renewcommand{\arraystretch}{1.05}

\begin{adjustbox}{width=\linewidth}
\begin{tabular}{c|ccccccccc}
\toprule[1pt]

\multicolumn{1}{c|}{\textbf{Metrics}}
& ADE$\downarrow$ (m)
& FDE$\downarrow$ (m)
& $\Delta$|SI|$\downarrow$ (m)
& $\Delta$|SA|$\downarrow$ (m$^2$)
& $\Delta$|TW|$\downarrow$ (m)
& $\Delta$|TL|$\downarrow$ (m)
& $\Delta$|FN|$\downarrow$ (m)
& $\Delta$|CD|$\downarrow$ (m)
& $\Delta$|SO|$\downarrow$ (-) \\

\cmidrule(lr){1-1}\cmidrule(lr){2-10}

\multicolumn{1}{c|}{\textbf{Time}}
& \multicolumn{9}{c}{\textbf{Original (Minimum Error)}} \\
\midrule

\textbf{1s} & 0.64 & 1.24 & 0.24 & 28.16 & 0.70 & 0.69 & 0.32 & 0.34 & 0.12 \\
\textbf{2s} & 1.35 & 2.84 & 0.48 & 59.22 & 1.43 & 1.45 & 0.66 & 0.80 & 0.19 \\
\textbf{3s} & 2.22 & 4.98 & 0.73 & 92.91 & 2.18 & 2.27 & 1.01 & 1.17 & 0.22 \\
\textbf{4s} & 3.26 & 7.59 & 1.00 & 129.34 & 2.95 & 3.14 & 1.40 & 1.36 & 0.23 \\
\textbf{5s} & 4.44 & 10.46 & 1.28 & 166.06 & 3.76 & 4.01 & 1.80 & 1.40 & 0.23 \\

\midrule\midrule

\multicolumn{1}{c|}{\textbf{Time}}
& \multicolumn{9}{c}{\textbf{Conditioned on Bundesliga (Minimum Error)}} \\
\midrule

\textbf{1s} & 0.36 & 0.91 & 0.14 & 18.85 & 0.59 & 0.53 & 0.20 & 0.49 & 0.19 \\
\textbf{2s} & 1.12 & 2.90 & 0.45 & 56.44 & 1.88 & 1.40 & 0.64 & 0.79 & 0.25 \\
\textbf{3s} & 2.17 & 5.64 & 0.88 & 104.65 & 3.67 & 2.34 & 1.23 & 1.08 & 0.27 \\
\textbf{4s} & 3.46 & 8.90 & 1.39 & 157.05 & 5.73 & 3.23 & 1.95 & 1.38 & 0.28 \\
\textbf{5s} & 4.93 & 12.48 & 1.97 & 209.15 & 7.98 & 3.99 & 2.76 & 1.57 & 0.29 \\

\bottomrule[1pt]
\end{tabular}
\end{adjustbox}

\vspace{6pt}
\caption{
\textbf{Minimum errors of trajectory forecasting tactical metrics for the Bundesliga with and without league conditioning.}
}
\label{tab:bundesliga_tactical_min}
\end{table}

\clearpage
\subsubsection{Performance of Objective-conditioned Trajectory Forecasting}
To evaluate the effect of objective-conditioned trajectory forecasting, we analyze how tactical metrics change after applying offensive or defensive guidance during trajectory generation. 
For each tactical scenario ({\em e.g.}, clearance, defended, goal, shot off target, and shot saved), 
we measure the resulting variations in high-level tactical metrics induced by the generated trajectories.

The evaluated metrics include off-ball expected threat ($\Delta$xThreat), depth threat ($\Delta$Dpt. Threat), width threat ($\Delta$Wid. Threat), defensive shape disruption ($\Delta$Def. Dis.), and defensive dominant region ($\Delta$Def. Dom. Reg.). 
Depth threat, width threat, and defensive shape disruption are scaled by $10^{-2}$ for readability.
Results under offensive guidance and defensive guidance are reported in Supplementary Table~\ref{tab:offensive_guidance} and Supplementary Table~\ref{tab:defensive_guidance}, respectively. 
For offensive guidance, larger increases in offensive metrics indicate more effective attacking behaviors, while reductions in defensive stability metrics reflect stronger pressure on the defense. 
Conversely, for defensive guidance, improvements correspond to reduced offensive threat and enhanced defensive structural stability.

\begin{table}[htbp]
\centering
\footnotesize
\setlength{\tabcolsep}{3pt}
\renewcommand{\arraystretch}{1.05}

\begin{adjustbox}{width=\linewidth}
\begin{tabular}{l|ccc|cc}
\toprule[1pt]

\multicolumn{1}{c|}{\multirow{2}{*}{\textbf{Scenario}}} &
\multicolumn{3}{c|}{\textbf{Offensive Side}} &
\multicolumn{2}{c}{\textbf{Defensive Side}} \\

\cmidrule(lr){2-4} \cmidrule(lr){5-6}

\multicolumn{1}{c|}{} &
$\Delta$xThreat$\uparrow$ &
$\Delta$Dpt. Threat$(\times 10^{-2})$$\uparrow$ &
$\Delta$Wid. Threat$(\times 10^{-2})$$\uparrow$ &
$\Delta$Def. Dis.$(\times 10^{-2})$$\downarrow$ &
$\Delta$Def. Dom. Reg.$(\text{m}^2)$$\downarrow$ \\

\midrule[1pt]
Clearance & 1.58 & 4.95 & 5.05 & -- & 101.24 \\
Defended & 0.30 & 0.58 & 0.53 & -- & 40.81 \\
Goal & 1.15 & 2.63 & 2.98 & -- & 45.05 \\
Shot off target & 0.65 & 1.27 & 2.22 & -- & 64.55 \\
Shot saved & 0.70 & 1.43 & 3.67 & -- & -11.32 \\

\cmidrule[1pt]{1-6}
\textbf{Overall} & \textbf{0.87} & \textbf{2.17} & \textbf{2.89} & \textbf{--} & \textbf{48.07} \\
\bottomrule[1pt]
\end{tabular}
\end{adjustbox}

\vspace{6pt}

\caption{
\textbf{Changes in tactical metrics after offensive guidance.}
Positive values indicate increased offensive threat or expanded attacking structures. The defensive shape disruption column is not available here since we control the defensive team in the same trajectory as condition to forecast and guide the evolution of offensive team. As a result, the deviation of defensive shape disruption is always 0.
}
\label{tab:offensive_guidance}
\end{table}

\begin{table}[htbp]
\centering
\footnotesize
\setlength{\tabcolsep}{3pt}
\renewcommand{\arraystretch}{1.05}

\begin{adjustbox}{width=\linewidth}
\begin{tabular}{l|ccc|cc}
\toprule[1pt]

\multicolumn{1}{c|}{\multirow{2}{*}{\textbf{Scenario}}} &
\multicolumn{3}{c|}{\textbf{Offensive Side}} &
\multicolumn{2}{c}{\textbf{Defensive Side}} \\

\cmidrule(lr){2-4} \cmidrule(lr){5-6}

\multicolumn{1}{c|}{} &
$\Delta$xThreat$\downarrow$ &
$\Delta$Dpt. Threat$(\times 10^{-2})$$\downarrow$ &
$\Delta$Wid. Threat$(\times 10^{-2})$$\downarrow$ &
$\Delta$Def. Dis.$(\times 10^{-2})$$\uparrow$ &
$\Delta$Def. Dom. Reg.$(\text{m}^2)$$\uparrow$ \\

\midrule[1pt]
Clearance & -1.10 & -4.58 & -5.56 & 8.90 & 27.51 \\
Defended & -0.45 & -2.28 & -2.00 & 5.86 & 32.07 \\
Goal & -0.70 & -3.14 & -3.52 & 0.43 & 171.55 \\
Shot off target & 0.08 & 0.16 & -3.25 & 1.96 & 51.20 \\
Shot saved & 0.12 & 0.16 & -3.90 & -0.33 & -7.77 \\

\cmidrule[1pt]{1-6}
\textbf{Overall} & \textbf{-0.41} & \textbf{-7.23} & \textbf{-2.93} & \textbf{3.49} & \textbf{58.02} \\
\bottomrule[1pt]
\end{tabular}
\end{adjustbox}

\vspace{6pt}
\caption{
\textbf{Changes in tactical metrics after defensive guidance.}
Improvements correspond to reduced offensive threat and enhanced defensive structural stability.
}
\label{tab:defensive_guidance}
\end{table}

As shown in Supplementary Table~\ref{tab:offensive_guidance} and Supplementary Table~\ref{tab:defensive_guidance}, objective-conditioned forecasting is consistent with the intended tactical guidance. 
Under offensive guidance, the generated trajectories increase off-ball expected threat, depth threat, and width threat overall by 0.87, 2.17, and 2.89, indicating more aggressive attacking behavior. 
In contrast, defensive guidance reduces these offensive metrics by 0.41, 7.23, and 2.93, respectively, while increasing defensive shape disruption and defensive dominant region by 3.49 and 58.02. 
This suggests that objective conditioning can effectively guide trajectory generation toward meaningful offensive or defensive tactical tendencies.

\clearpage
\subsubsection{Performance of Generalization of Trajectory Forecasting of GenTac on Other Team Sports}

We further evaluate the generalization ability of GenTac trajectory forecasting across multiple team sports, including basketball, American football, and ice hockey, as reported in Supplementary Table~\ref{tab:table_multisport_history_window_condition} and Supplementary Table~\ref{tab:table_multisport_history_window_condition_avg}. 
Compared with Supplementary Table~\ref{tab:soccer_history_window_condition} and Supplementary Table~\ref{tab:soccer_history_window_condition_avg}, the experimental protocol remains identical except that an additional parameter \textbf{H(s)} is introduced to denote the history horizon length used for trajectory forecasting.
The frame rate of each sport is indicated in parentheses. 

For all cross-sport trajectory forecasting experiments, we evaluate GenTac under the same two causal window settings, i.e., 0.2\,s and 1\,s, while preserving the original frame rate of each dataset, namely 5 FPS for basketball, 10 FPS for American football, and 30 FPS for ice hockey. This setting allows us to assess the generalization ability of GenTac across different team sports under consistent forecasting protocols while respecting the native temporal resolution of each benchmark. 
A different history length is additionally considered for basketball due to the intrinsic structure of its dataset. Specifically, the basketball benchmark consists of 6-second trajectory clips. Therefore, when fixing the history horizon to 4\,s, the remaining future available for evaluation is limited to only 2\,s. To enable longer-horizon forecasting analysis from 1\,s to 5\,s, we further reduce the history horizon to 1\,s for basketball, thereby reserving a sufficiently long future segment for evaluation.


\begin{table}[htbp]
\centering
\footnotesize
\setlength{\tabcolsep}{4pt}
\begin{adjustbox}{width=\linewidth}
\begin{tabular}{ccc|cc|cc|cc|cc|cc}
\toprule[1pt]
\multicolumn{3}{c|}{\textbf{Time}}
& \multicolumn{2}{c|}{\textbf{1s}}
& \multicolumn{2}{c|}{\textbf{2s}}
& \multicolumn{2}{c|}{\textbf{3s}}
& \multicolumn{2}{c|}{\textbf{4s}}
& \multicolumn{2}{c}{\textbf{5s}} \\
\cmidrule(lr){1-3}\cmidrule(lr){4-5}\cmidrule(lr){6-7}\cmidrule(lr){8-9}\cmidrule(lr){10-11}\cmidrule(lr){12-13}
\textbf{H(s)} & \textbf{W(s)} & \textbf{OC}
& minADE$_K$ & minFDE$_K$
& minADE$_K$ & minFDE$_K$
& minADE$_K$ & minFDE$_K$
& minADE$_K$ & minFDE$_K$
& minADE$_K$ & minFDE$_K$ \\
\midrule


\multicolumn{13}{c}{\textbf{Basketball (5FPS)}} \\
\midrule
4 & 0.2 & \ding{51} & 0.04 & 0.08 & 0.07 & 0.19 & -- & -- & -- & -- & -- & -- \\
4 & 0.2 & \ding{55} & 0.23 & 0.38 & 0.46 & 0.86 & -- & -- & -- & -- & -- & -- \\
4 & 1   & \ding{51} & 0.05 & 0.10 & 0.12 & 0.23 & -- & -- & -- & -- & -- & -- \\
4 & 1   & \ding{55} & 0.21 & 0.40 & 0.50 & 1.02 & -- & -- & -- & -- & -- & -- \\
1 & 0.2 & \ding{51} & 0.04 & 0.08 & 0.1 & 0.21 & 0.17 & 0.36 & 0.25 & 0.51 & 0.32 & 0.64 \\
1 & 0.2 & \ding{55} & 0.15 & 0.28 & 0.37 & 0.78 & 0.65 & 1.43 & 0.97 & 2.19 & 1.33 & 3.01 \\
1 & 1   & \ding{51} & 0.12 & 0.22 & 0.24 & 0.40 & 0.34 & 0.51 & 0.42 & 0.60 & 0.48 & 0.67 \\
1 & 1   & \ding{55} & 0.98 & 1.91 & 1.94 & 2.64 & 2.57 & 3.78 & 3.17 & 4.89 & 3.70 & 5.47 \\
\midrule

\multicolumn{13}{c}{\textbf{American Football (10FPS)}} \\
\midrule
4 & 0.2 & \ding{51} & 0.11 & 0.21 & 0.28 & 0.63 & 0.51 & 1.19 & 0.78 & 1.81 & 1.06 & 2.46 \\
4 & 0.2 & \ding{55} & 0.24 & 0.48 & 0.63 & 1.39 & 1.13 & 2.55 & 1.69 & 3.84 & 2.29 & 5.12 \\
4 & 1   & \ding{51} & 0.24 & 0.44 & 1.24 & 4.37 & 2.34 & 4.87 & 3.02 & 4.97 & 3.48 & 5.16 \\
4 & 1   & \ding{55} & 0.64 & 1.17 & 2.68 & 8.67 & 5.26 & 10.70 & 7.06 & 9.98 & 8.35 & 10.22 \\
\midrule

\multicolumn{13}{c}{\textbf{Ice Hockey (30FPS)}} \\
\midrule
4 & 0.2 & \ding{51} & 0.24 & 0.48 & 0.51 & 0.86 & 0.71 & 1.12 & 0.89 & 1.30 & 1.04 & 1.45 \\
4 & 0.2 & \ding{55} & 0.62 & 1.24 & 1.37 & 2.57 & 2.07 & 3.70 & 2.71 & 4.65 & 3.29 & 5.44 \\
4 & 1   & \ding{51} & 2.72 & 3.21 & 3.54 & 3.76 & 3.96 & 3.94 & 4.20 & 3.95 & 4.38 & 4.01 \\
4 & 1   & \ding{55} & 3.03 & 4.15 & 4.67 & 5.10 & 5.63 & 5.70 & 6.30 & 6.13 & 6.89 & 6.35 \\
\bottomrule[1pt]
\end{tabular}
\end{adjustbox}
\vspace{6pt}
\caption{
\textbf{GenTac trajectory forecasting performance (minimum error) across multiple team sports under varying window lengths and opponent conditioning.}
}
\label{tab:table_multisport_history_window_condition}
\end{table}

\begin{table}[htbp]
\centering
\footnotesize
\setlength{\tabcolsep}{4pt}
\begin{adjustbox}{width=\linewidth}
\begin{tabular}{ccc|cc|cc|cc|cc|cc}
\toprule[1pt]
\multicolumn{3}{c|}{\textbf{Time}}
& \multicolumn{2}{c|}{\textbf{1s}}
& \multicolumn{2}{c|}{\textbf{2s}}
& \multicolumn{2}{c|}{\textbf{3s}}
& \multicolumn{2}{c|}{\textbf{4s}}
& \multicolumn{2}{c}{\textbf{5s}} \\
\cmidrule(lr){1-3}\cmidrule(lr){4-5}\cmidrule(lr){6-7}\cmidrule(lr){8-9}\cmidrule(lr){10-11}\cmidrule(lr){12-13}
\textbf{H(s)} & \textbf{W(s)} & \textbf{OC}
& avgADE$_K$ & avgFDE$_K$
& avgADE$_K$ & avgFDE$_K$
& avgADE$_K$ & avgFDE$_K$
& avgADE$_K$ & avgFDE$_K$
& avgADE$_K$ & avgFDE$_K$ \\
\midrule


\multicolumn{13}{c}{\textbf{Basketball (5FPS)}} \\
\midrule
4 & 0.2 & \ding{51} & 0.06 & 0.12 & 0.14 & 0.30 & -- & -- & -- & -- & -- & -- \\
4 & 0.2 & \ding{55} & 0.33 & 0.55 & 0.66 & 1.31 & -- & -- & -- & -- & -- & -- \\
4 & 1   & \ding{51} & 0.08 & 0.16 & 0.19 & 0.38 & -- & -- & -- & -- & -- & -- \\
4 & 1   & \ding{55} & 0.35 & 0.70 & 0.87 & 1.90 & -- & -- & -- & -- & -- & -- \\
1 & 0.2 & \ding{51} & 0.06 & 0.11 & 0.14 & 0.3 & 0.24 & 0.51 & 0.34 & 0.72 & 0.44 & 0.91 \\
1 & 0.2 & \ding{55} & 0.22 & 0.42 & 0.55 & 1.22 & 0.98 & 2.29 & 1.49 & 3.49 & 2.03 & 4.69 \\
1 & 1   & \ding{51} & 0.20 & 0.38 & 0.38 & 0.66 & 0.51 & 0.84 & 0.61 & 0.97 & 0.70 & 1.06 \\
1 & 1   & \ding{55} & 1.62 & 3.18 & 3.24 & 5.69 & 4.35 & 7.04 & 5.15 & 7.83 & 5.74 & 8.29 \\
\midrule

\multicolumn{13}{c}{\textbf{American Football (10FPS)}} \\
\midrule
4 & 0.2 & \ding{51} & 0.13 & 0.23 & 0.31 & 0.69 & 0.55 & 1.28 & 0.83 & 1.95 & 1.14 & 2.65 \\
4 & 0.2 & \ding{55} & 0.28 & 0.57 & 0.74 & 1.65 & 1.30 & 2.97 & 1.94 & 4.43 & 2.61 & 5.91 \\
4 & 1   & \ding{51} & 0.30 & 0.56 & 1.36 & 4.76 & 2.55 & 5.31 & 3.25 & 5.40 & 3.71 & 5.69 \\
4 & 1   & \ding{55} & 0.77 & 1.41 & 4.18 & 15.83 & 9.13 & 19.76 & 11.6 & 17.04 & 12.85 & 18.14 \\
\midrule

\multicolumn{13}{c}{\textbf{Ice Hockey (30FPS)}} \\
\midrule
4 & 0.2 & \ding{51} & 0.40 & 0.82 & 0.79 & 1.48 & 1.11 & 1.93 & 1.35 & 2.24 & 1.55 & 2.48 \\
4 & 0.2 & \ding{55} & 0.92 & 1.92 & 2.03 & 4.21 & 3.17 & 6.54 & 4.29 & 8.84 & 5.39 & 10.86 \\
4 & 1   & \ding{51} & 3.61 & 4.55 & 4.36 & 5.19 & 4.73 & 5.44 & 4.93 & 5.46 & 5.06 & 5.57 \\
4 & 1   & \ding{55} & 6.00 & 8.47 & 7.72 & 10.41 & 8.85 & 12.00 & 9.68 & 12.85 & 10.35 & 13.63 \\
\bottomrule[1pt]
\end{tabular}
\end{adjustbox}
\vspace{6pt}
\caption{
\textbf{GenTac trajectory forecasting performance (average error) across multiple team sports under varying window lengths and opponent conditioning.}
}
\label{tab:table_multisport_history_window_condition_avg}
\end{table}

\clearpage
\subsubsection{Performance of Tactical Event Grounding and Forecasting}
We further evaluate the capability of GenTac to ground and forecast tactical events directly from predicted trajectories. 
The evaluation follows the event taxonomy introduced in our benchmark, including both high-level event types and fine-grained subtypes. 
Supplementary Table~\ref{tab:traj_cls_type} reports the trajectory-based classification performance for 5 event \textbf{types}, while Supplementary Table~\ref{tab:traj_cls_subtype} presents the results for more detailed 15 \textbf{subtypes}. 
Performance is reported using top-$k$ recall for $k \in \{1,3\}$ for event types and $k \in \{1,3,5\}$ for event subtypes, together with macro-average recall and overall accuracy.

An observation from our classification analysis is the intrinsic ambiguity in 2D trajectory representations. Since our dataset captures the $(x, y)$ coordinates of players and the ball, certain events exhibit near-identical physical patterns despite having distinct tactical outcomes. For instance, a``shot off target'' that travels over the crossbar may be indistinguishable from a ``Goal'' in the 2D plane.
\begin{table}[h]
\centering
\small
\setlength{\tabcolsep}{20pt}
\renewcommand{\arraystretch}{1.15}

\begin{adjustbox}{max width=0.38\linewidth}
\begin{tabular}{l|cc}
\toprule[1.2pt]
\textbf{Top-$k$} & \textbf{@1} & \textbf{@3} \\
\cmidrule[0.5pt]{1-3}
\textbf{Type} & \multicolumn{2}{c}{\textbf{Recall}} \\
\midrule

Build         & 61.70 & 97.87 \\
Interruption  & 78.95 & 84.21 \\
Set piece     & 63.89 & 97.22 \\
Threat        & 82.50 & 90.00 \\
Transition    & 72.65 & 99.59 \\
\cmidrule[1pt]{1-3}

\textbf{Macro avg.} & 71.94 & 93.78 \\
\cmidrule[1pt]{1-3}

\textbf{Accuracy} & 71.16 & 97.40 \\

\bottomrule[1.2pt]
\end{tabular}
\end{adjustbox}
\vspace{10pt}

\caption{\textbf{Trajectory-based event classification (Type).}
Per-class Performance Recall is reported for top-$k$ predictions ($k\in\{1,3\}$), together with Macro average and Overall Performance accuracy.}
\label{tab:traj_cls_type}
\end{table}

\begin{table}[h]
\centering
\scriptsize
\setlength{\tabcolsep}{20pt}
\renewcommand{\arraystretch}{1.15}

\begin{adjustbox}{max width=0.8\linewidth}
\begin{tabular}{l|ccc}
\toprule[1.2pt]
\textbf{Top-$k$} & \textbf{@1} & \textbf{@3} & \textbf{@5} \\
\cmidrule[0.5pt]{1-4}
\textbf{Subtype} & \multicolumn{3}{c}{\textbf{Recall}} \\
\midrule

Ball win           & 55.13 & 92.31 & 98.72 \\
Clearance          & 27.27 & 36.36 & 63.64 \\
Defended           & 12.50 & 50.00 & 50.00 \\
Goal               & 33.33 & 100.00 & 100.00 \\
Goal kick          & 0.00  & 66.67 & 100.00 \\
Progression        & 47.19 & 97.75 & 100.00 \\
Shot off target    & 36.36 & 63.64 & 81.82 \\
Shot saved         & 28.57 & 57.14 & 85.71 \\
Build              & 61.70 & 91.49 & 97.87 \\
Corner             & 75.00 & 87.50 & 100.00 \\
Free kick          & 35.29 & 70.59 & 82.35 \\
Kick off           & 50.00 & 50.00 & 50.00 \\
Penalty            & 50.00 & 50.00 & 50.00 \\
Stoppage           & 78.95 & 78.95 & 94.74 \\
Throw in           & 75.00 & 90.00 & 90.00 \\
\cmidrule[1pt]{1-4}

\textbf{Macro avg.} & 44.42 & 72.16 & 82.99 \\
\cmidrule[1pt]{1-4}

\textbf{Accuracy} & 53.66 & 87.47 & 94.33 \\

\bottomrule[1.2pt]
\end{tabular}
\end{adjustbox}
\vspace{10pt}

\caption{\textbf{Trajectory-based event classification (Subtype).}
Per-class Performance Recall is reported for top-$k$ predictions ($k\in\{1,3,5\}$), together with Macro average and Overall Performance accuracy.}
\label{tab:traj_cls_subtype}
\end{table}

\end{document}